\documentclass[letterpaper, 10 pt, conference]{ieeeconf}
\IEEEoverridecommandlockouts                              

\usepackage{cite}
\usepackage{amsmath}
\usepackage{graphicx}
\usepackage{float}
\usepackage{subfigure}
\usepackage{overpic}
\usepackage{wrapfig}
\usepackage{balance} 
\usepackage{ulem}
\usepackage{color}
\usepackage[colorlinks,linkcolor=red]{hyperref} 
\usepackage{multirow}
\usepackage{mathtools}
\usepackage{booktabs} 
\usepackage{newtxmath}
\usepackage[linesnumbered, ruled]{algorithm2e}
\graphicspath{{figures/}}

\title{\LARGE \bf
  GS-SDF: LiDAR-Augmented Gaussian Splatting and Neural SDF\\for Geometrically Consistent Rendering and Reconstruction
}

\author{
  Jianheng Liu, Yunfei Wan, Bowen Wang, Chunran Zheng, Jiarong Lin, and Fu Zhang
  \thanks{J. Liu, Y. W, B. W, C. Z, J. L, and F. Zhang are with the Department of Mechanical Engineering, University of Hong Kong. Email: jianheng@connect.hku.hk, fuzhang@hku.hk}
}%

\begin{document}

\maketitle
\thispagestyle{empty}
\pagestyle{empty}



\begin{abstract}

  Digital twins are fundamental to the development of autonomous driving and embodied artificial intelligence.
  However, achieving high-granularity surface reconstruction and high-fidelity rendering remains a challenge.
  Gaussian splatting offers efficient photorealistic rendering but struggles with geometric inconsistencies due to fragmented primitives and sparse observational data in robotics applications.
  Existing regularization methods, which rely on render-derived constraints, often fail in complex environments.
  Moreover, effectively integrating sparse LiDAR data with Gaussian splatting remains challenging.
  We propose a unified LiDAR-visual system that synergizes Gaussian splatting with a neural signed distance field.
  The accurate LiDAR point clouds enable a trained neural signed distance field to offer a manifold geometry field.
  This motivates us to offer an SDF-based Gaussian initialization for physically grounded primitive placement and a comprehensive geometric regularization for geometrically consistent rendering and reconstruction.
  Experiments demonstrate superior reconstruction accuracy and rendering quality across diverse trajectories.
  To benefit the community, the codes are released at \url{https://github.com/hku-mars/GS-SDF}.

\end{abstract}

\section{Introduction}

\begin{figure}
  \centering
  \setlength{\subfigcapskip}{-3pt} 
  \subfigure[Data Collection Trajectory and Input Point Clouds]{
    \includegraphics[width=0.46\textwidth]{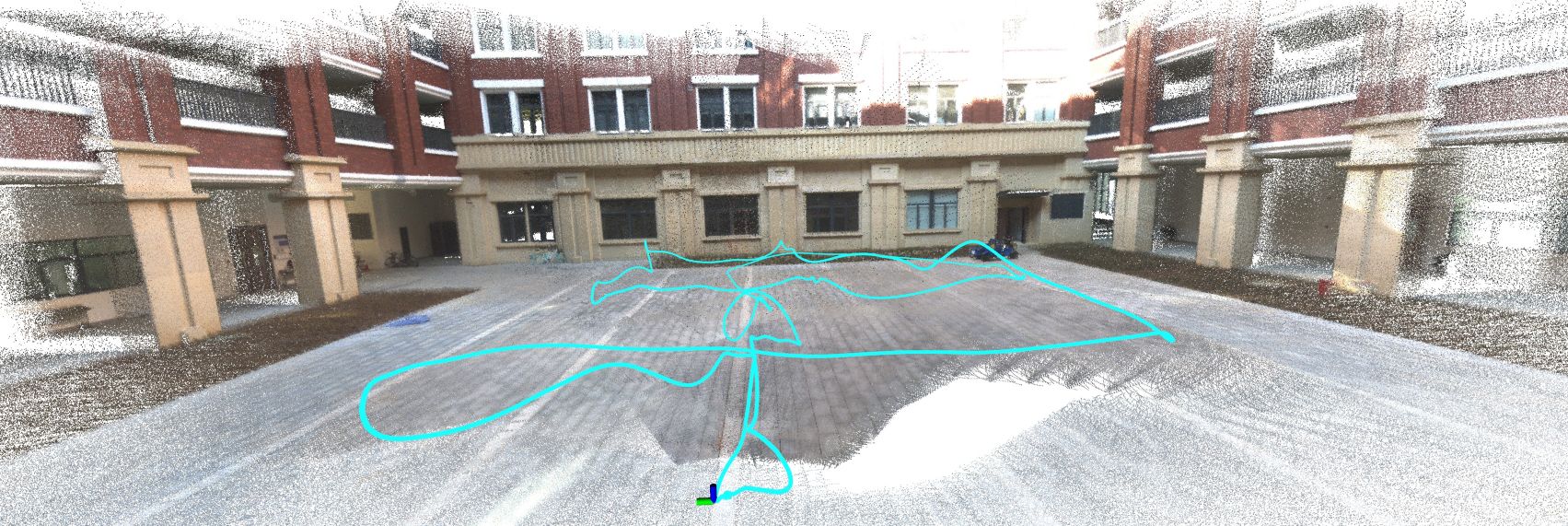}
  }
  \vspace{-6pt}

  \subfigure[Surface Reconstruction]{
    \includegraphics[width=0.46\textwidth]{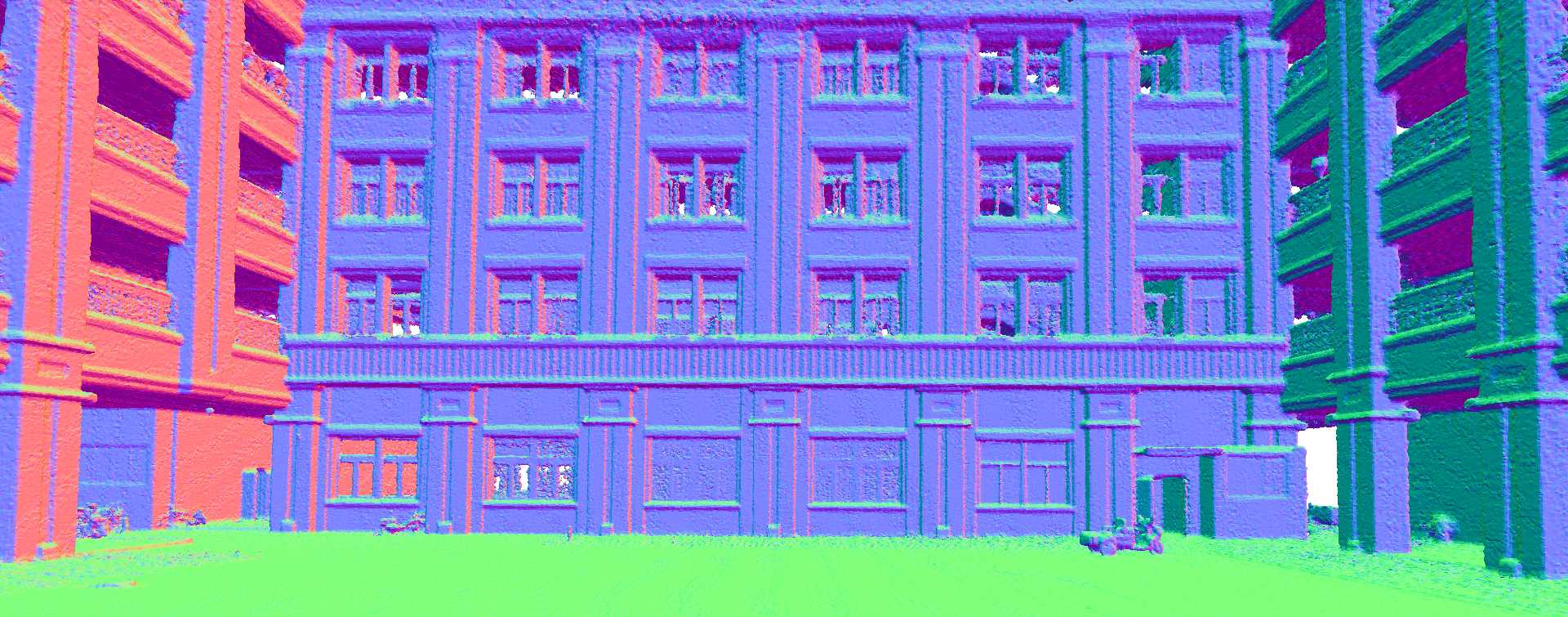}
  }
  \vspace{-6pt}

  \subfigure[Geometrically Consistent Gaussian Splats]{
    \includegraphics[width=0.46\textwidth]{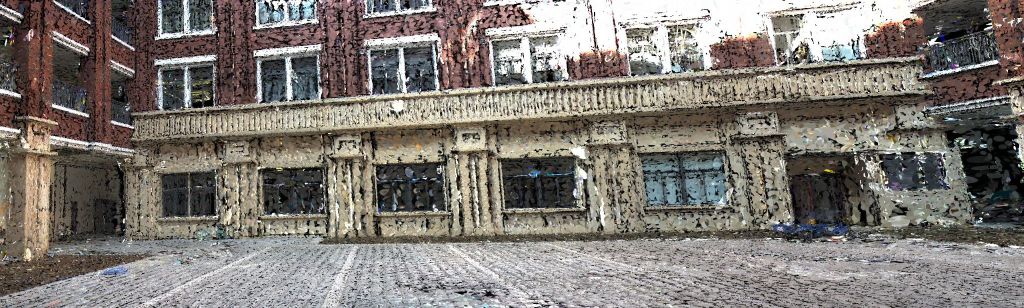}
  }
  \vspace{-6pt}

  \subfigure[Novel View Synthesis]{
    \centering
    \begin{minipage}[b]{0.15\textwidth}
      \includegraphics[width=\linewidth]{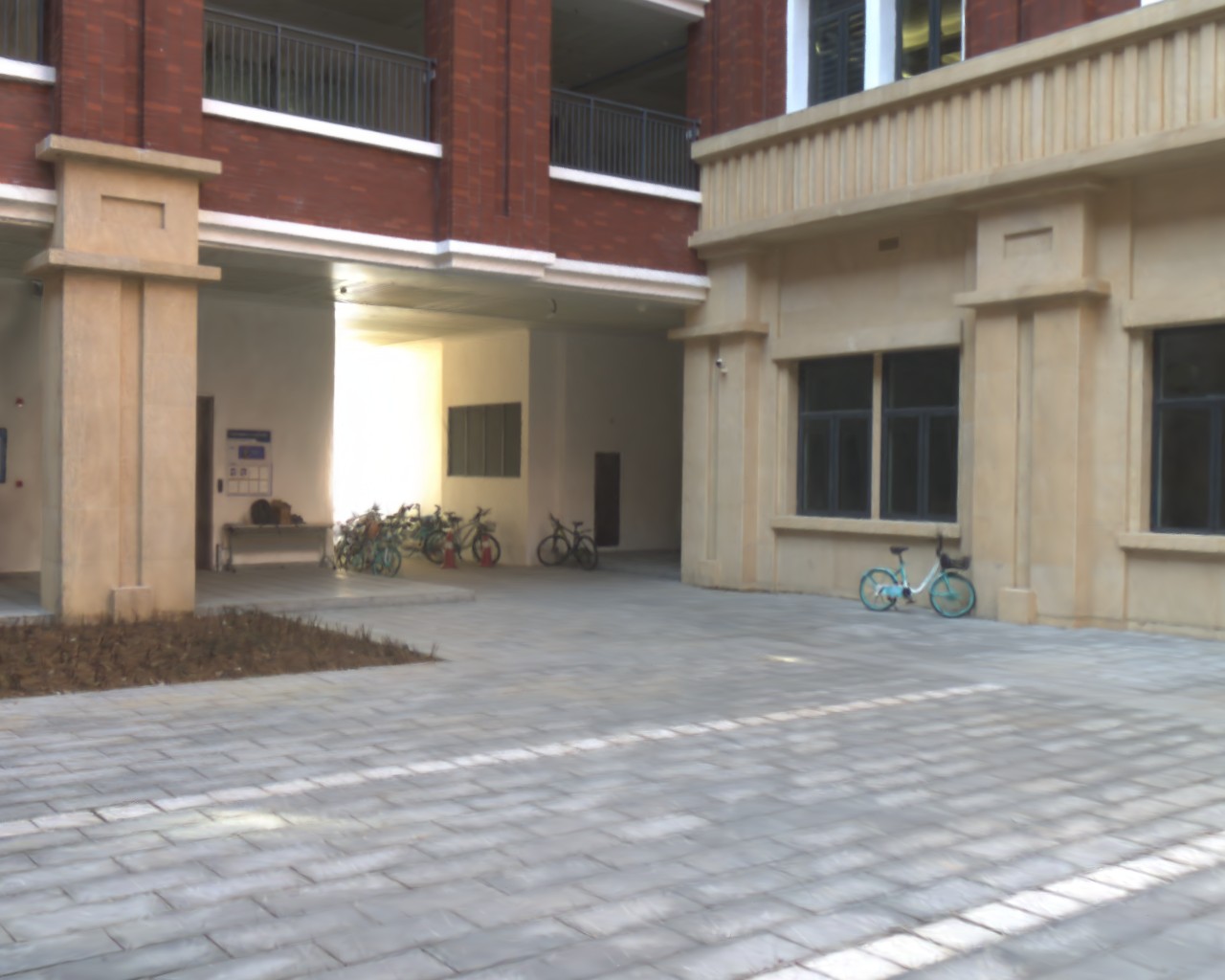}
    \end{minipage}
    \hfill
    \begin{minipage}[b]{0.15\textwidth}
      \includegraphics[width=\linewidth]{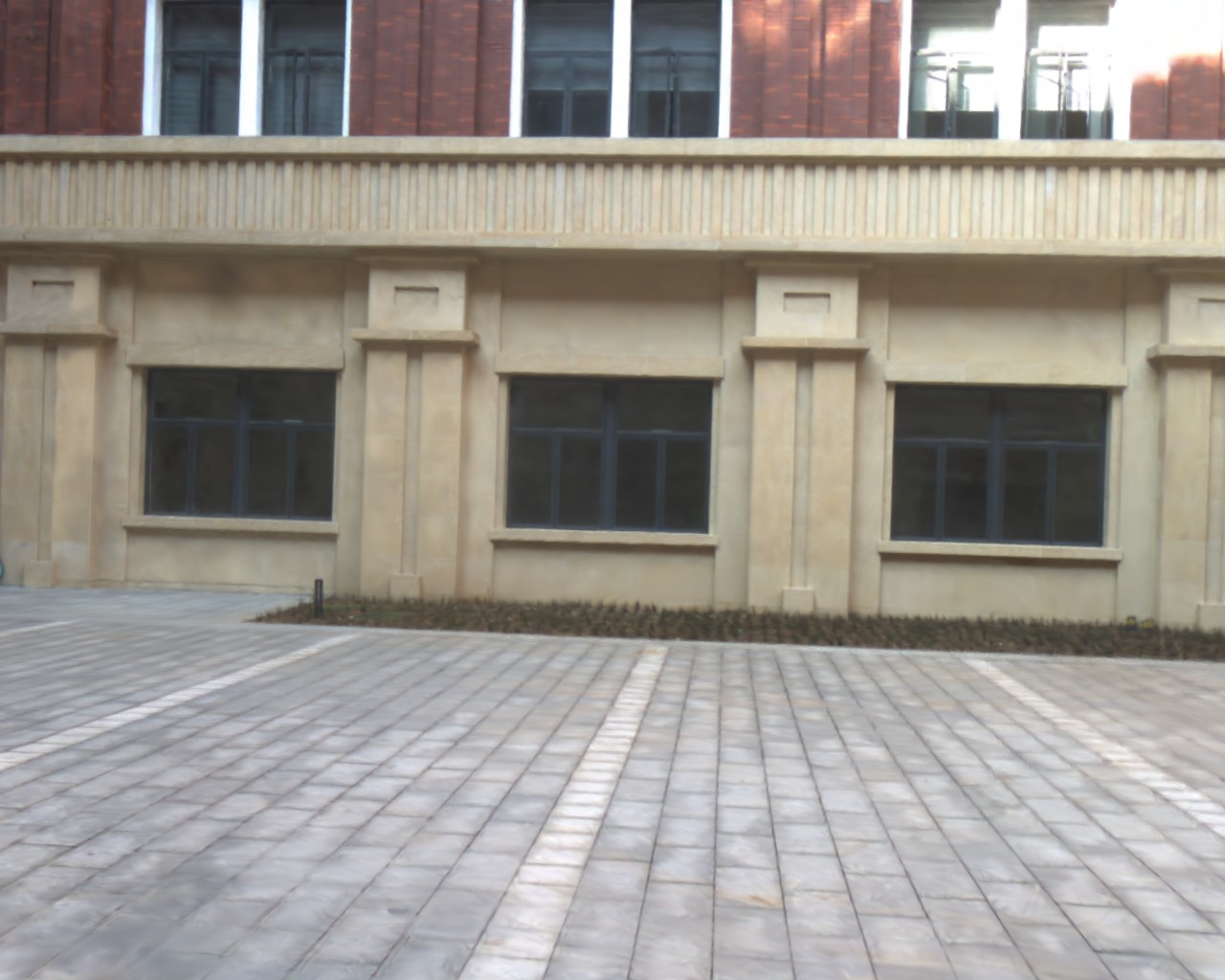}
    \end{minipage}
    \hfill
    \begin{minipage}[b]{0.15\textwidth}
      \includegraphics[width=\linewidth]{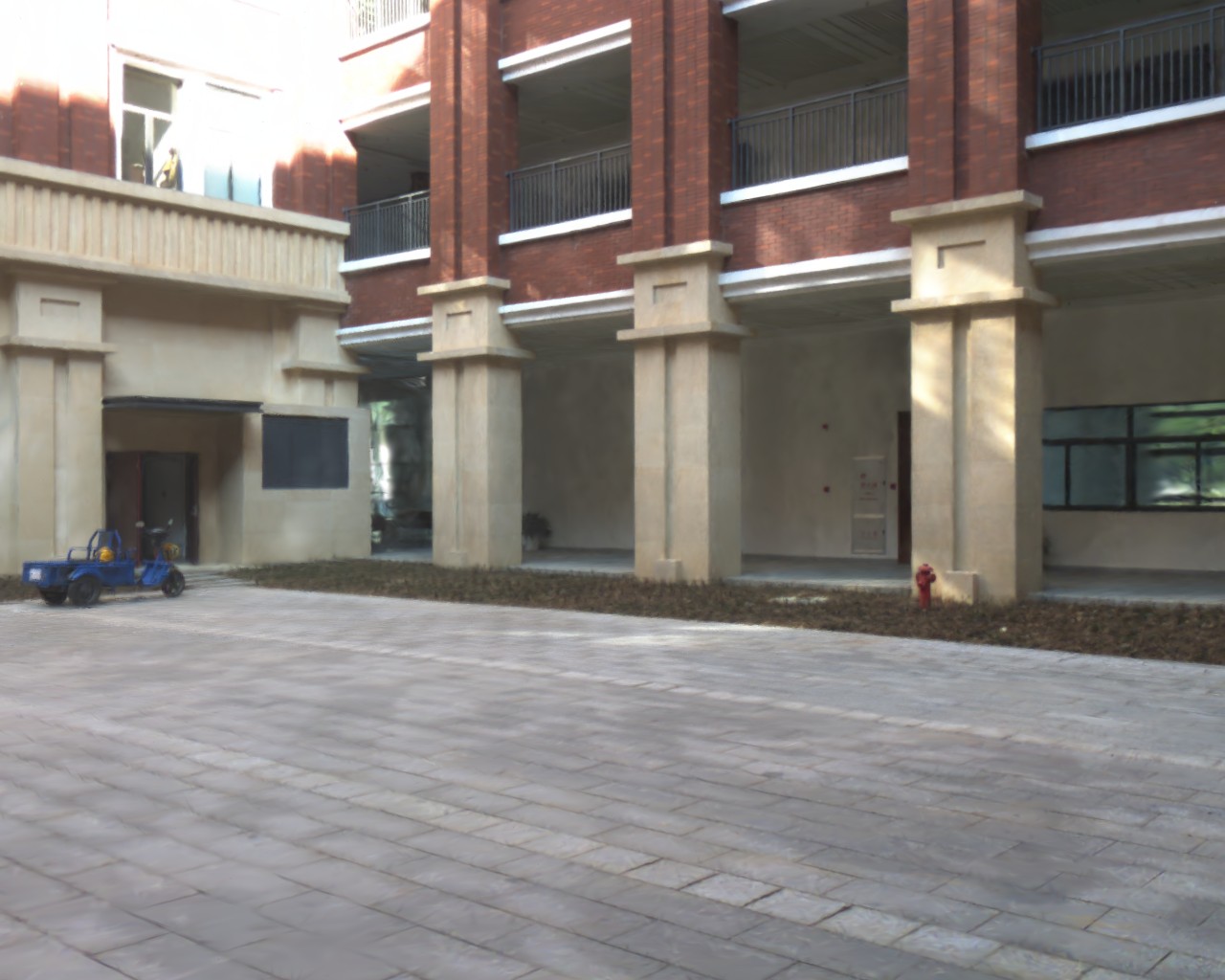}
    \end{minipage}
  }
  \vspace{-6pt}

  \caption{We show the input raw colorized point clouds (FAST-LIVO2 SYSU Scene) collected by a real-world LiDAR-visual sensor system.
  We demonstrate the proposed system, \textbf{GS-SDF}, with the surface reconstruction results (color indicates the normal direction), Gaussian splats' geometric distribution, and novel view synthesis results.}
  \vspace{-14pt}
  \label{fig:fast_livo_sysu}
\end{figure}

3D surface reconstructions and photorealistic renderings are essential for a wide range of applications, like augmented reality\cite{wang2022survey}, and embodied artificial intelligence\cite{wang2024embodiedscan}.
The increasing accessibility of cameras and cost-effective LiDAR sensors on robots has enabled the collection of rich multimodal data.
LiDAR-visual SLAM technologies\cite{lvisam2021shan,lin2022r,zheng2024fast} are widely applied in 3D reconstruction tasks.
However, these methods typically produce only discrete colorized raw point clouds, which lack complete surface structure and rendering ability for view-dependent novel view synthesis.
In practical digital-twin applications \cite{straub2019replica}, high-quality watertight surface reconstructions and photorealistic rendering are critical.

Neural Radiance Fields (NeRFs) \cite{mildenhall2021nerf} have gained attention for their ability to generate high-fidelity renderings.
However, their computational intensity and inefficiency in real-time applications limit their practicality.
3D Gaussian Splatting (3DGS) \cite{kerbl20233d} has emerged as a compelling alternative to NeRFs for efficient, high-quality rendering.
Unlike NeRF, which requires laborious volume rendering, 3DGS employs explicit ellipsoidal Gaussian primitives for efficient rasterization.
However, the fragmented structure of these primitives compromises structural continuity and often misaligns with underlying geometry.
This misalignment hinders geometrically consistent rendering, particularly in robotics applications where free-view trajectories yield insufficient observations.
Furthermore, under image-only supervision with imperfect camera poses, 3DGS frequently overfits to photometric cues, producing floating artifacts.
It motivates us to introduce geometric regularization into the 3DGS framework.

Geometric inconsistencies in 3DGS manifest as rendering distortions.
Current regularization strategies focus on rendering-depth-derived constraints, such as rendering normal consistency from 2D Gaussian splatting (2DGS) \cite{huang20242d} or multi-view alignment from PGSR\cite{chen2024pgsr}.
They focus on object-centric scenes with abundant multi-view regularization and limited generalization to complex environments.
Depth cameras and LiDARs providing direct structural priors inspire us to incorporate these sensors into the 3DGS framework for geometric regularization.
And depth cameras with limited precision are confined to indoor applications\cite{GSSLAM2024}.
While LiDAR provides sparse point clouds, it remains challenging to integrate with NeRF\cite{liu2024neural} or 3DGS\cite{jiang2024li} frameworks designed for dense, image-scale geometry.

The combination of LiDAR and neural signed distance fields (NSDF) has shown promise in high-granularity surface reconstructions\cite{liu2023towards}.
The NSDF, providing a manifold geometry field, motivates us to explore a more comprehensive geometric regularization for Gaussian splats and the importance of the Gaussian initialization to the results.
Surface reconstruction from Gaussian splats is complicated by their discontinuous nature.
The truncated Signed Distance Field (TSDF) fusion \cite{newcombe2011kinectfusion} is commonly applied for 3DGS-based surface reconstructions through depth images rendered from Gaussian splatting, which demonstrates unsatisfactory inaccuracies in surface reconstruction for generalized scenes.

To address the above limitations, we propose the combination of 2D Gaussian splatting and neural signed distance field for LiDAR-visual systems.
We aim to reconstruct both the appearance and surface of generalized scenes using posed images and low-cost LiDAR data from any casual trajectories.
We emphasize the important role of the Gaussian initialization in returning a good structure of the Gaussian splats.
We explore the combination of the NSDF and Gaussian splats to leverage the structural and photometric hints for high-granularity surface reconstruction and geometrically consistent rendering, as shown in Fig.~\ref{fig:fast_livo_sysu}.
Our contributions are as follows:
\begin{enumerate}

  \item A unified LiDAR-visual system achieving geometrically consistent photorealistic rendering and high-granularity surface reconstruction.  
  \item A physically grounded Gaussian initialization based on the neural signed distance field to boost training convergence and reduce floating artifacts. 
  \item A thorough shape regularization gives bidirectional supervision between the neural signed distance field and Gaussian splats for geometrically consistent rendering and reconstruction. 
\end{enumerate}
Extensive experiments validate our approach across diverse scenarios, demonstrating superior reconstruction accuracy and rendering quality.

\section{Related Works}


\subsection{Geometrically Consistent Novel View Synthesis}

NeRFs \cite{mildenhall2021nerf} and 3DGS \cite{kerbl20233d} have been widely employed for novel view synthesis, owing to their remarkable photorealistic rendering capabilities.
However, neither of them can guarantee geometric consistency, frequently leading to the emergence of floating artifacts \cite{tao2024oxford}.
To tackle this problem, geometric regularizations and structural priors derived from depth sensors have been incorporated into both NeRF and 3DGS.
2DGS \cite{huang20242d} projects Gaussian ellipsoids into surfels and enforces normal-depth alignment geometric regularization.
PGSR \cite{chen2024pgsr} provides further guarantees for multi-view rendering consistency regularization.
RGBD cameras that provide aligned color and depth information are well practiced in both NeRFs \cite{sucar2021imap,jiang2023h} and 3DGS \cite{GSSLAM2024} to align rendered depth images with sensor-acquired ones.
However, the low measurement accuracy of depth cameras limited their application in indoor scene settings.
LiDAR sensors provide accurate depth information, but their sparse nature complicates integration with NeRFs \cite{tao2024silvr} or 3DGS \cite{hong2024liv} frameworks designed for dense, image-scale geometry.
M2Mapping \cite{liu2024neural} enforces geometric consistency to NeRF through the combination of NSDF and ray marching, albeit at a significant computational expense.
LIV-GaussMap \cite{hong2024liv} stabilizes splat orientations using planar priors derived from LiDAR-inertial odometry.
And LI-GS \cite{jiang2024li} utilizes Gaussian Mixture Models for constraining the Gaussian splats' distribution.

\subsection{Surface Reconstruction}
Surface reconstructions predominantly use point clouds with direct structural information.
The TSDF fusion \cite{newcombe2011kinectfusion} is well practiced for surface reconstructions\cite{vizzo2022vdbfusion}.
While the quest for continuous surface reconstruction motivates the exploration of implicit representations, like Poisson functions \cite{kazhdan2006poisson} or NSDFs \cite{ortiz2022isdf,zhong2022shine}.
In terms of the exploration of Gaussian splats for surface reconstruction, their discontinuous nature complicates surface reconstruction.
Recent approaches circumvent this limitation through intermediate volumetric representations.
SuGaR \cite{guedon2024sugar} computes an approximate density field based on the opacities of neighboring Gaussians, and GOF \cite{yu2024gaussian} defines per-point opacity as the minimal opacity across training views.
However, both density and opacity fields suffer from computational inefficiency and ambiguous surface-level sets for surface reconstruction.
In contrast, the SDF facilitates precise surface extraction through defined zero-level sets.
The TSDF fusion driven by the rendered depth images is favored by 3DGS-based surface reconstructions\cite{huang20242d,chen2024pgsr}, but yields insufficient accuracy.
Image-driven NSDFs combined with 3DGS \cite{lyu20243dgsr,yu2024gsdf} are still under exploration for achieving considerable accuracy in surface reconstructions.

\section{Methodology}

\begin{figure}[!h]
  \centering
  \includegraphics[width=0.47\textwidth]{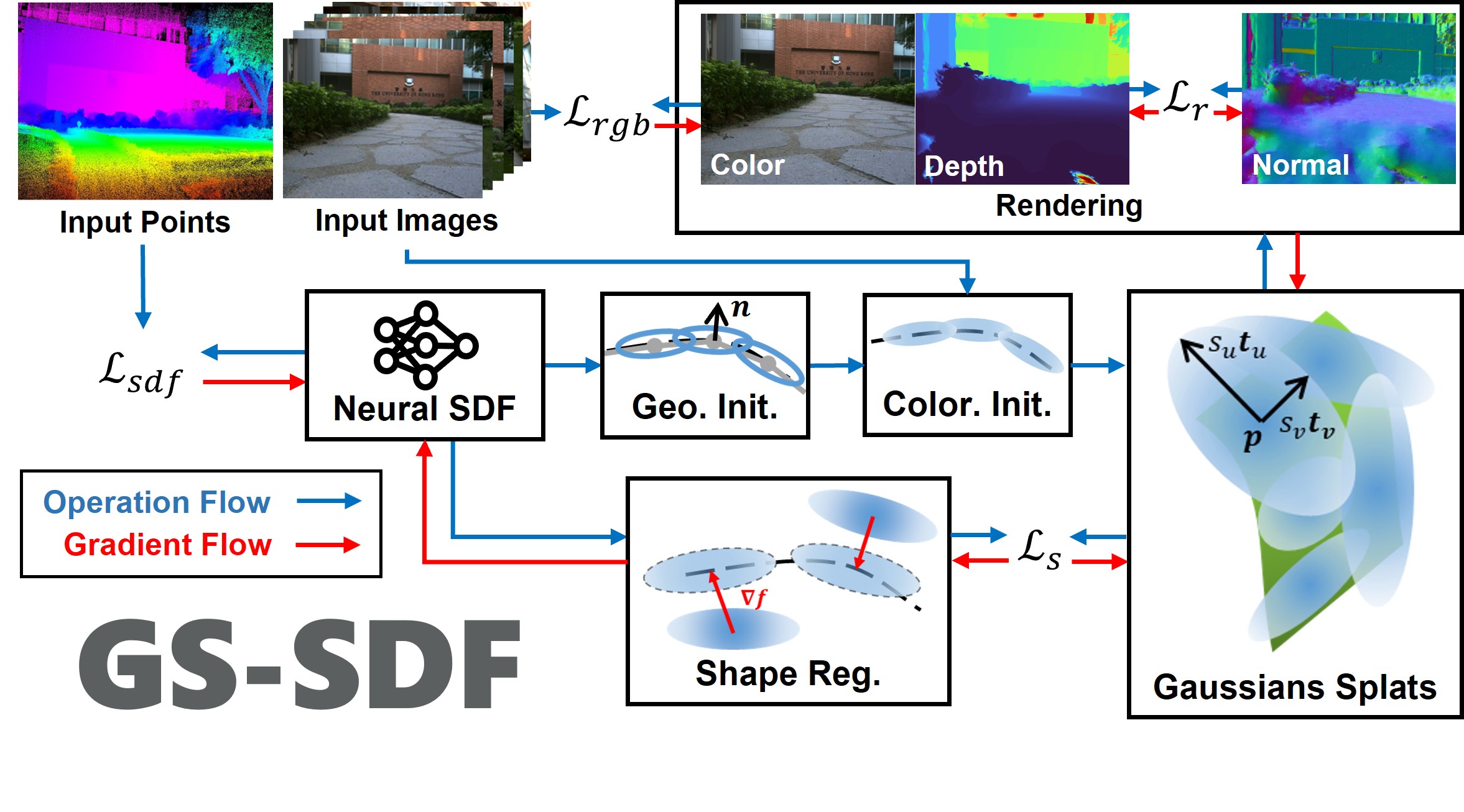}
  \vspace{-12pt}
  \caption{
    The overall pipeline of the proposed system, GS-SDF, for geometrically consistent rendering and reconstruction.
  }
  \vspace{-6pt}
  \label{fig:pipeline}
\end{figure}



Given posed images and point clouds from LiDAR-inertial-visual odometry (e.g., FAST-LIVO2 \cite{zheng2024fast}), we propose a framework to jointly reconstruct high-granularity scene geometry and photorealistic appearance.
As shown in Fig.~\ref{fig:pipeline}, our pipeline consists of three stages:
A neural signed distance field (NSDF) is first trained using point clouds to establish a manifold geometry field (Sec.~\ref{sec:nsdf}).
Gaussian splats are initialized from the NSDF (Sec.~\ref{sec:gs_init}), and a thorough SDF-aided shape regularization for both Gaussian splats and the NSDF (Sec.~\ref{sec:gs_reg}) achieves a geometric-consistent rendering and reconstruction.

\subsection{2D Gaussian Splatting}
\label{sec:preliminaries}

2DGS \cite{huang20242d} represents 3D scenes using planar Gaussian disks.
Each disk is characterized by a center $\boldsymbol{p} \in \mathbb{R}^3$, orthogonal tangent vectors $\{\boldsymbol{t}_u, \boldsymbol{t}_v\in \mathbb{R}^3\} $ that define its local plane, scale factors $\{s_u, s_v \in \mathbb{R}\}$, opacity $\alpha \in \mathbb{R}$, and view-dependent color $\boldsymbol{c} \in \mathbb{R}^3$ encoded using spherical harmonics.
The disk's normal vector is computed as the cross product of its tangent vectors: $\boldsymbol{n} = \boldsymbol{t}_u \times \boldsymbol{t}_v$.
A point $\boldsymbol{u} = [u, v]^\top$ in the tangent plane is mapped to 3D space via the transformation $\boldsymbol{p}(u,v) = \boldsymbol{p} + s_u u \boldsymbol{t}_u + s_v v \boldsymbol{t}_v$, with the Gaussian weight kernel defined as $\mathcal{G}(\boldsymbol{u}) = \exp\left(-\frac{u^2 + v^2}{2}\right)$.
For rendering, Gaussians are transformed into camera coordinates and depth-sorted.
The final pixel color $\boldsymbol{c}$ is computed through alpha blending:
\begin{equation}
  \boldsymbol{C} = \sum_{i=1} \boldsymbol{c}_i \alpha_i \mathcal{G}_i(\boldsymbol{u}) \prod_{j=1}^{i-1} \left(1 - \alpha_j \mathcal{G}_j(\boldsymbol{u})\right)
  =\sum_{i=1} \boldsymbol{c}_i w_i.
  \label{eq:blending}
\end{equation}
The Gaussian attributes are optimized using a combined rendering loss function of L1 loss and structural dissimilarity (DSSIM) metrics \cite{kerbl20233d}:
\begin{equation}
  \begin{aligned}
    \mathcal{L}_{c}=0.8 L_1({\boldsymbol{C}}, \bar{\boldsymbol{C}})+0.2 L_{\mathrm{DSSIM}}({\boldsymbol{C}}, \bar{\boldsymbol{C}}),
  \end{aligned}
\end{equation}
where $\bar{\boldsymbol{C}}$ denotes the ground truth pixel color.

\subsection{Neural Signed Distance Field}
\label{sec:nsdf}

We utilize hash encoding \cite{muller2022instant} and multi-layer perceptrons (MLPs) to construct a neural network for shaping a scalable signed distance field: $({s},{\beta}) = f(\boldsymbol{x})$.
A 3D point $\boldsymbol{x} \in \mathbb{R}^3$ is mapped to a signed distance value $s \in \mathbb{R}$ and a scale factor $\beta \in \mathbb{R}$\cite{liu2024neural}.
For a LiDAR measurement comprising an origin ${}^{L}\boldsymbol{o}$, a ray direction ${}^{L}\boldsymbol{d}$ and a distance $t$, we uniformly sample points along the ray path $\{t_i:{}^{L}\boldsymbol{x}_i = {}^{L}\boldsymbol{o} + t_i\ {}^{L}\boldsymbol{d}\}$.
The NSDF gives prediction at each sampled point: $(s_i, \beta_i) = f({}^{L}\boldsymbol{x}_i)$.
The ray distance $\bar{s}_i = t - t_i$ derived from the sampled distance is converted to an occupancy value $\bar{o}_i = \Phi(-\bar{s}_i, \beta_i)$ using the sigmoid function $\Phi(v, h) = (1 + \exp(-v/h))^{-1}$ and the predicted scale factor ${\beta}_i$.
The NSDF is trained using binary cross-entropy loss \cite{zhong2022shine}:
\begin{equation}
  \mathcal{L}_{sdf} = -\sum_{i=1} \big[{o}_{i} \log(\bar{o}_{i}) + (1 - {o}_{i}) \log(1 - \bar{o}_{i})\big],
  \label{eq:sdf_loss}
\end{equation}
where $o_i = \Phi(-{s}_i, {\beta}_i)$ is the predicted occupancy value.
The Eikonal regularization \cite{yang2023steik} is employed to ensure the gradient of the SDF maintains unit magnitude:
\begin{equation}
  \mathcal{L}_{e} = \sum_{i=1} \left(\|\nabla f(\boldsymbol{x}_i)\|_2 - 1\right)^2.
\end{equation}


\subsection{Initialization of Gaussian Splats}
\label{sec:gs_init}
In this section, we address the crucial role that Gaussian initialization plays in optimizations and propose a careful Gaussian initialization with the help of a coarse neural signed distance field trained using an input point cloud.
We provide a thorough initialization of Gaussians, which boosts the convergence efficiency and suppresses floating artifacts.

\subsubsection{SDF-aided Geometry Initialization}
The neural signed distance field, providing a manifold geometry field, can well stabilize the initial geometry of Gaussians.
We extract a surface mesh from the NSDF using marching cubes\cite{lorensen1987marching}, and utilize its vertices as initial Gaussian positions $\boldsymbol{p}$.
This approach provides spatially accurate and denser positional initialization compared to structure-from-motion points, and reduces the impact of sensor outliers compared to the sampling initialization from input point clouds. 
The marching cube resolution directly determines the initial scale $\{s_u, s_v\}$ of the Gaussian splats.
We leverage the SDF's geometric hints and Gram-Schmidt orthogonalization for Gaussian orientation initialization, using gradient direction $\boldsymbol{n} = \frac{\nabla f}{\|\nabla f\|}$ as Gaussian normals and principal curvature direction $\boldsymbol{b} = \frac{\nabla^2 f}{\|\nabla^2 f\|}$ to define the second rotation axis $\boldsymbol{t}_u = \frac{\boldsymbol{b} - (\boldsymbol{n}^T\boldsymbol{b})\boldsymbol{n}}{\|\boldsymbol{b} - (\boldsymbol{n}^T\boldsymbol{b})\boldsymbol{n}\|}$, and the first rotation axis $\boldsymbol{t}_v = \boldsymbol{n} \times \boldsymbol{t}_u$ for the final rotation: $\boldsymbol{R} = \left[\boldsymbol{t}_u, \boldsymbol{t}_v, \boldsymbol{n}\right]$.
The opacity is derived from the SDF value and Gaussian kernel: $\alpha = \exp(-\frac{s^2}{\beta})$.

\subsubsection{Sky Initialization}
For the infinite background, we uniformly tile the opaque Gaussian over a map-sized sphere.
It provides a canvas for infinite objects to avoid floating artifacts derived from the foreground Gaussians.
\subsubsection{Color Initialization}
Next, we maintain fixed Gaussians' structural attributes (position, rotation, scale, and opacity) while training the Gaussians on all training images in a single round for color initialization.
The color initialization helps to solidify the Gaussians' structure from being deviated in early training stages.

\subsection{Geometric Regularization for Rendering}
\label{sec:gs_reg}

The rasterization-based rendering of Gaussian splatting can lead to occlusion artifacts where foreground elements inappropriately obscure background content. 
Geometric regularization is designed to hold the rendering consistency of different views to avoid such artifacts.
Considering that the LiDAR structure priors are not always available everywhere, we incorporate both the rendering-based regularization in 2D space and the SDF-aided shape regularization in 3D space to guarantee a geometrically consistent rendering.

\subsubsection{Rendering Consistency Regularization}
Following the rendering regularization in 2DGS\cite{huang20242d}, we obtain the rendered normal images from splats' normals:
$\boldsymbol{N} =\sum_{i=1} \boldsymbol{n}_i w_i$.
And the finite differential normal images from the rendered depth images: $\hat{\boldsymbol{N}}(x,y) = \frac{\nabla_x \boldsymbol{p}_d \times \nabla_y \boldsymbol{p}_d}{\|\nabla_x \boldsymbol{p}_d \times \nabla_y \boldsymbol{p}_d\|}$, where $\boldsymbol{p}_d$ is the point derived from the rendered depth images at piexl $(x,y)$.
The rendering consistency regularization aligns the rendered normal with the finite differential normal as follows:
\begin{equation}
  \mathcal{L}_{r} = (1-\boldsymbol{N}^T\hat{\boldsymbol{N}}).
  \label{eq:render_reg}
\end{equation}

\subsubsection{SDF-aided Shape Regularization}
\label{sec:gs_shape_reg}

\begin{figure}[!h]
  \centering
  \setlength{\subfigcapskip}{-3pt} 

  \setcounter{subfigure}{0}
  \subfigure[Center Regularization]{
    \label{fig:geo_center_reg}
    \centering
    \includegraphics[width=0.23\textwidth]{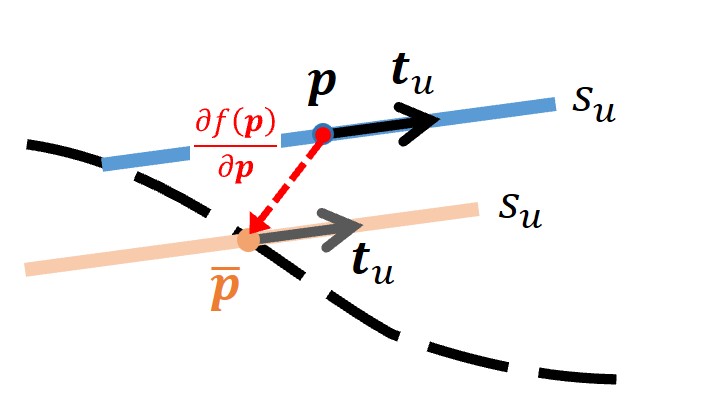}
  }
  \hspace{-16pt}
  \subfigure[Shape Regularization]{
    \label{fig:geo_shape_reg}
    \centering
    \includegraphics[width=0.23\textwidth]{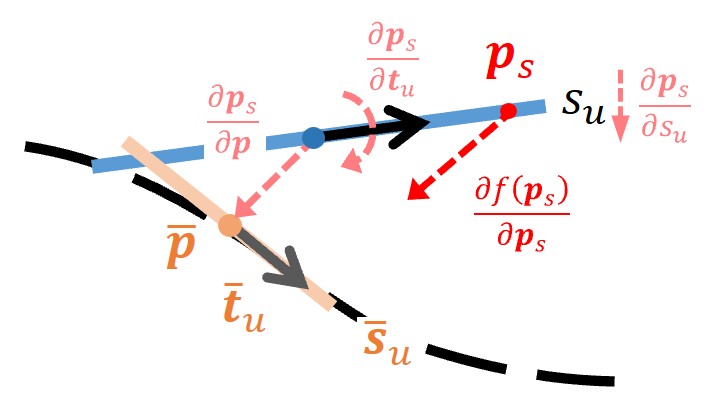}
  }

  \caption{
    Illustration of different geometric regularizations. The blue splats represent the initial splat, and the orange splats represent the optimized splat. The dashed lines represent the optimization directions.
  }
  \label{fig:geo_reg}
\end{figure}

With the NSDF providing a manifold geometry field, we can regularize the Gaussian splats' shape to align with the physical surface.
Consider that the splats represent a local plane of the surface, simply pulling the splat center $\boldsymbol{p}$ to the zero-level sets is insufficient\cite{yu2024gsdf}, as shown in Fig.~\ref{fig:geo_center_reg}.
We propose the shape regularization to enable comprehensive structural regularization across the entire splat surface.
For each splat, we sample a point $\boldsymbol{u}_{s}=[u_{s},v_{s}]^T \sim \mathcal{N}(0,1)$ on its disk surface:
\begin{equation}
  \begin{aligned}
    \boldsymbol{p}_{s} &= \boldsymbol{p} + \boldsymbol{t}_{u} s_{u} u_{s} + \boldsymbol{t}_{v} s_{v} v_{s}.
  \end{aligned}
\end{equation}
For $i$-th sampled point $\boldsymbol{p}_{s_i}$, we infer its signed distance value $f(\boldsymbol{p}_{s_i})$ to align splats with NSDF's zero-level sets:
\begin{equation}
  \label{eq:gs_sdf}
  \mathcal{L}_{s} = \sum_{i=1} \frac{1}{2} W_i \mathcal{G}(\boldsymbol{u}_{s_i}) f(\boldsymbol{p}_{s_i})^2,
\end{equation}
where $W_i = \sum_k w_k$ represents the accumulated weight of the $i$-th splat from Eq.~\ref{eq:blending} in a image.
Considering that the SDF gradient maintains unit magnitude $\|\frac{\partial f(\boldsymbol{p}_{s_i})}{\partial \boldsymbol{p}_{s_i}}\| = 1$, we adopt the $L_2$ loss function to enable adaptive refinement with proper gradients:
$\frac{\partial\mathcal{L}_{s}}{\partial \boldsymbol{p}_{s_i}} = W_i \mathcal{G}(\boldsymbol{u}_{s_i}) f(\boldsymbol{p}_{s_i}) \frac{\partial f(\boldsymbol{p}_{s_i})}{\partial \boldsymbol{p}_{s_i}}$.

As illustrated in Fig.~\ref{fig:geo_shape_reg}, we take one sampling for each Gaussian splat in the shape regularization.
This regularization provides a concise all-in-one shape alignment to enable gradient-based supervision of all structural attributes through the following partial derivatives: $\frac{\partial \boldsymbol{p}_{s}}{\partial \boldsymbol{p}} = \boldsymbol{I}$, $\frac{\partial \boldsymbol{p}_{s}}{\partial \boldsymbol{t}_{u}} = s_{u} u_{s}$, $\frac{\partial \boldsymbol{p}_{s}}{\partial \boldsymbol{t}_{v}} = s_{v} v_{s}$, $\frac{\partial \boldsymbol{p}_{s}}{s_{u}} = \boldsymbol{t}_u u_{s}$, $\frac{\partial \boldsymbol{p}_{s}}{s_{v}} = \boldsymbol{t}_{v} u_{s}$.

\subsection{Optimization}

The overall optimization function is a combination of the supervision from point clouds, images, and geometric regularizations:
\begin{equation}
  \begin{aligned}
    \mathcal{L} &= \mathcal{L}_{sdf} + \lambda_{e}\mathcal{L}_{e} + \mathcal{L}_{c} + \lambda_r\mathcal{L}_{r} + \lambda_s \mathcal{L}_{s},
  \end{aligned}
\end{equation}
where $\lambda_{e} = 0.1$, $\lambda_{r} = 0.01$, and $\lambda_{s} = 0.005$ are hyperparameters that balance the loss terms.





\begin{table*}[h]
  \centering
  \caption{Quantitative surface reconstruction, interpolation(I) and extrapolation(E) rendering results on the Replica dataset. }
  \label{tab:replica_quan_inter}
  \resizebox{1.0\textwidth}{!}{
    \begin{tabular}{ccccccccccc}
      \toprule
      \textbf{Metrics} & \textbf{Methods} & \textbf{Office-0} & \textbf{Office-1} & \textbf{Office-2} & \textbf{Office-3} & \textbf{Office-4} & \textbf{Room-0} & \textbf{Room-1} & \textbf{Room-2} & \textbf{Avg.} \\
      \midrule

      \multirow{6}*{C-L1[cm]$\downarrow$}

      & {VDBFusion}
      & 0.618 & 0.595 & 0.627 & 0.685 & 0.633 & 0.637 & 0.558 & 0.656 & 0.626
      \\
      & {iSDF}
      & 2.286 & 4.032 & 3.144 & 2.352 & 2.120 & 1.760 & 1.712 & 2.604 & 2.501
      \\
      & {SHINE-Mapping}
      & 0.753 & 0.663 & 0.851 & 0.965 & 0.770 & 0.650 & 0.844 & 0.826 & 0.790
      \\
      & {H2Mapping}
      & {0.557} & {0.529} & {0.585} & {0.644} & {0.616} & {0.568} & {0.523} & {0.616} & {0.580}
      \\
      & {M2Mapping} 
      & \textbf{0.494} & \textbf{0.476} & \textbf{0.501} & \textbf{0.517} & \textbf{0.531} & \textbf{0.486} & \textbf{0.455} & \textbf{0.532} & \textbf{0.499}
      \\

      & Ours
      & \underline{0.500} & \underline{0.487} & \underline{0.502} & \underline{0.525} & \underline{0.540} & \underline{0.489} & \underline{0.460} & \underline{0.542} & \underline{0.506}\\
      \midrule

      \multirow{6}*{F-Score[\%]$\uparrow$}

      & {VDBFusion}
      & 97.129 & 97.107 & 97.007 & 96.221 & 96.993 & 97.871 & 98.311 & 96.117 & 97.095
      \\
      & {iSDF}
      & 75.829 & 55.429 & 71.232 & 77.274 & 81.281 & 78.253 & 84.801 & 76.815 & 75.114
      \\
      & {SHINE-Mapping}
      & 94.397 & 95.606 & 92.150 & 87.855 & 94.427 & 97.010 & 91.307 & 92.601 & 93.169
      \\
      & {H2Mapping}
      & {98.430} & {98.279} & {97.935} & {97.077} & {97.391} & {98.850} & {99.003} & {96.904} & {97.983}
      \\
      & {M2Mapping}
      & \underline{98.970} & \textbf{98.771} & \underline{98.657} & \textbf{98.415} & \textbf{98.158} & \underline{99.238} & \textbf{99.496} & \textbf{97.677} & \textbf{98.673}
      \\

      & {Ours}
      & \textbf{98.986} & \underline{98.628} & \textbf{98.671} & \underline{98.290} & \underline{97.990} & \textbf{99.242} & \underline{99.468} & \underline{97.582} & \underline{98.607}\\
      \midrule
      \midrule

      \multirow{8}*{SSIM(I)$\uparrow$}

      & {InstantNGP}
      & 0.981 &  {0.980} & 0.963 & 0.960 & 0.966 & 0.964 & 0.964 & 0.967 & 0.968
      \\

      & {3DGS}
      & \textbf{0.987} & {0.980} & \textbf{0.980} & \textbf{0.976} & \textbf{0.980} & \underline{0.977} & \textbf{0.982} & \textbf{0.980} & \textbf{0.980}
      \\

      & {2DGS}
      & 0.978 & 0.958 & 0.967 & 0.964 & 0.969 & 0.966 & 0.971 & 0.969 & 0.968\\

      & {PGSR}
      & 0.981 & 0.967 &  0.966 & 0.963 & 0.966 & 0.898 & 0.973 & 0.972 & 0.961\\

      & {H2Mapping}
      & 0.963 & 0.960 & 0.931 & 0.929 & 0.941 & 0.914 & 0.929 & 0.935 & 0.938
      \\

      & {MonoGS}
      & {0.985} & \underline{{0.982}} & {0.973} & \underline{0.970} & {0.976} & {0.971} & {0.976} & {0.977}  & {0.975}
      \\

      & {M2Mapping}
      & {0.985} & \textbf{0.983} & {0.973} & \underline{0.970} & \underline{0.977} & 0.968 & 0.975 & {0.977} & {0.976}
      \\

      & {Ours}
      & \underline{0.986} & {0.971} & \underline{0.979} & \textbf{0.976} & \textbf{0.980} & \textbf{0.978} & \underline{0.980} & \underline{0.979} & \underline{0.979}\\

      \midrule

      \multirow{8}*{PSNR(I)$\uparrow$}

      & {InstantNGP}
      & 43.538 & \underline{{44.340}} & {38.273} & {37.603} & {39.772} & {37.926} & 38.859 & 39.568 & 39.984
      \\

      & {3DGS}
      & {{43.932}} & 43.237 & {{39.182}} & {{38.564}} & \underline{{41.366}} & \underline{{39.081}} & \underline{{41.288}} & \underline{{41.431}} & {{41.010}}
      \\

      & {2DGS}
      & 41.854 & 40.796 & 38.138 & 37.704 & 40.144 & 37.484 & 39.905 & 39.748 & 39.472\\

      & {PGSR}
      & 43.018 & 42.303 & 38.405 & 37.634 & 40.143 & 38.165 & 40.246 & 40.415 & 40.041\\

      & {H2Mapping}
      & 38.307 & 38.705 & 32.748 & 33.021 & 34.308 & 31.660 & 33.466 & 32.809 & 34.378
      \\

      & {MonoGS}
      & {43.648} & {43.690} & {37.695} & {37.539} & {40.224} & {37.779} & {39.563} & {40.134} & {40.034}
      \\

      & {M2Mapping}
      & \underline{44.369} & \textbf{44.935} & \underline{39.652} & \underline{{38.874}} & {41.318} & {38.541} & {40.775} & {40.705} & \underline{{41.146}}
      \\

      & Ours
      & \textbf{44.383} & 43.389 & \textbf{39.851} & \textbf{39.245} & \textbf{41.989} & \textbf{39.303} & \textbf{41.500} & \textbf{41.802} & \textbf{41.433}\\
      \midrule
      \midrule

      \multirow{8}*{SSIM(E)$\uparrow$}

      & {InstantNGP}
      & 0.972 & \underline{0.961} & 0.934 & 0.938 &  {0.952} & {0.918} & 0.936 & 0.941 & 0.944
      \\

      & {3DGS}
      & 0.936 & 0.897 & 0.924 & 0.917 & 0.925 & 0.881 & 0.915 & 0.919 & 0.914
      \\

      & {2DGS}
      & 0.944 & 0.829 & 0.922 & 0.920 & 0.940 & 0.901 & 0.924 & 0.922 & 0.913\\

      & {PGSR}
      & 0.949 &  0.870 & 0.932 & 0.931 & 0.938 &  0.970 & 0.924 &  0.930 & 0.930\\

      & {H2Mapping}
      & 0.957 & 0.955 & 0.932 & 0.925 & 0.937 & 0.866 & 0.916 & 0.917 & 0.926
      \\

      & {MonoGS}
      & \underline{0.974} & \underline{0.961} & {0.945} & {0.942} &  {0.950} & {0.912}& {0.942} & {0.946} & {0.947}
      \\

      & {M2Mapping}
      & \textbf{0.980} & \textbf{0.976} & \textbf{0.960} & \textbf{0.964} & \textbf{0.970} & \textbf{0.955} & \textbf{0.963} & \textbf{0.965} & \textbf{0.967}
      \\

      & {Ours}
      & 0.973 & 0.952 & \underline{0.956} & \underline{0.952} & \underline{0.958} & \underline{0.945} & \underline{0.949} & \underline{0.954} & \underline{0.955}\\
      \midrule

      \multirow{8}*{PSNR(E)$\uparrow$}

      & {InstantNGP}
      & \underline{{39.874}} & \underline{{39.120}} & 31.274 & \underline{32.135} & \underline{34.458} & \underline{{32.587}} & \underline{{33.024}} & {32.266} & \underline{{34.341}}
      \\

      & {3DGS}
      & 31.220 & 29.959 & 27.411 & 26.442 & 28.324 & 27.541 & 28.429 & 27.139 & 28.307
      \\
      & {2DGS}
      & 30.847 & 25.546 & 27.248 & 25.499 & 29.270 & 28.149 & 28.694 & 26.846 & 27.762\\
      & {PGSR}
      & 32.525 & 29.544 & 28.243 & 27.032 & 28.944 & 27.801 & 28.977 & 27.603 & 28.834\\

      & {H2Mapping}
      & 36.740 & 37.841 & \underline{31.427} & 31.144 & 31.988 & 28.815 & 31.192 & 30.603 & 32.468
      \\

      & {MonoGS}
      & 39.197 & 38.818 & 29.740 & 29.664 & 31.632 & 29.949 & 31.126 & 30.621 & 32.593
      \\

      & {M2Mapping}
      & \textbf{41.965} & \textbf{42.215} & \textbf{35.056} & \textbf{37.465} & \textbf{38.667} & \textbf{36.427} & \textbf{37.294} & \textbf{36.722} & \textbf{38.226}
      \\


      & {Ours}
      & 39.102 & 38.174 & 30.769 & 31.503 & 33.211 & 32.253 & 32.226 & \underline{32.566} & 33.725\\

      \bottomrule
    \end{tabular}
  }
  \vspace{-12pt}
\end{table*}

\section{Experiments}

\subsection{Implementation Details}

We represent our neural signed distance field following a similar architecture to InstantNGP \cite{muller2022instant}, utilizing a combination of multi-resolution hash encoding and a tiny MLP decoder.
Given any position $x$, the hash encoding concatenates each resolution's interpolation features to form a feature vector of size $32$.
A geometry MLP with 64-width and 3 hidden layers decodes the encoding feature to obtain the SDF value and scale.
We sample 4096 rays for NSDF training with 4 surface samples and 4 free space samples on each ray.
We first take 10000 iterations for NSDF training and leverage the trained NSDF for our initialization as described in Sec.~\ref{sec:gs_init}.
Then, we follow the setting of 2DGS\cite{huang20242d} to take 30000 iterations for the training.
The implementation is based on  LibTorch and CUDA, and all experiments are conducted on a platform equipped with an Intel i7-13700K CPU and an NVIDIA RTX 4090 GPU.

\begin{figure}[!h]
  \centering
  \setlength{\subfigcapskip}{-3pt} 

  \subfigure[VDB-Fusion]{
    \centering
    \includegraphics[width=0.15\textwidth]{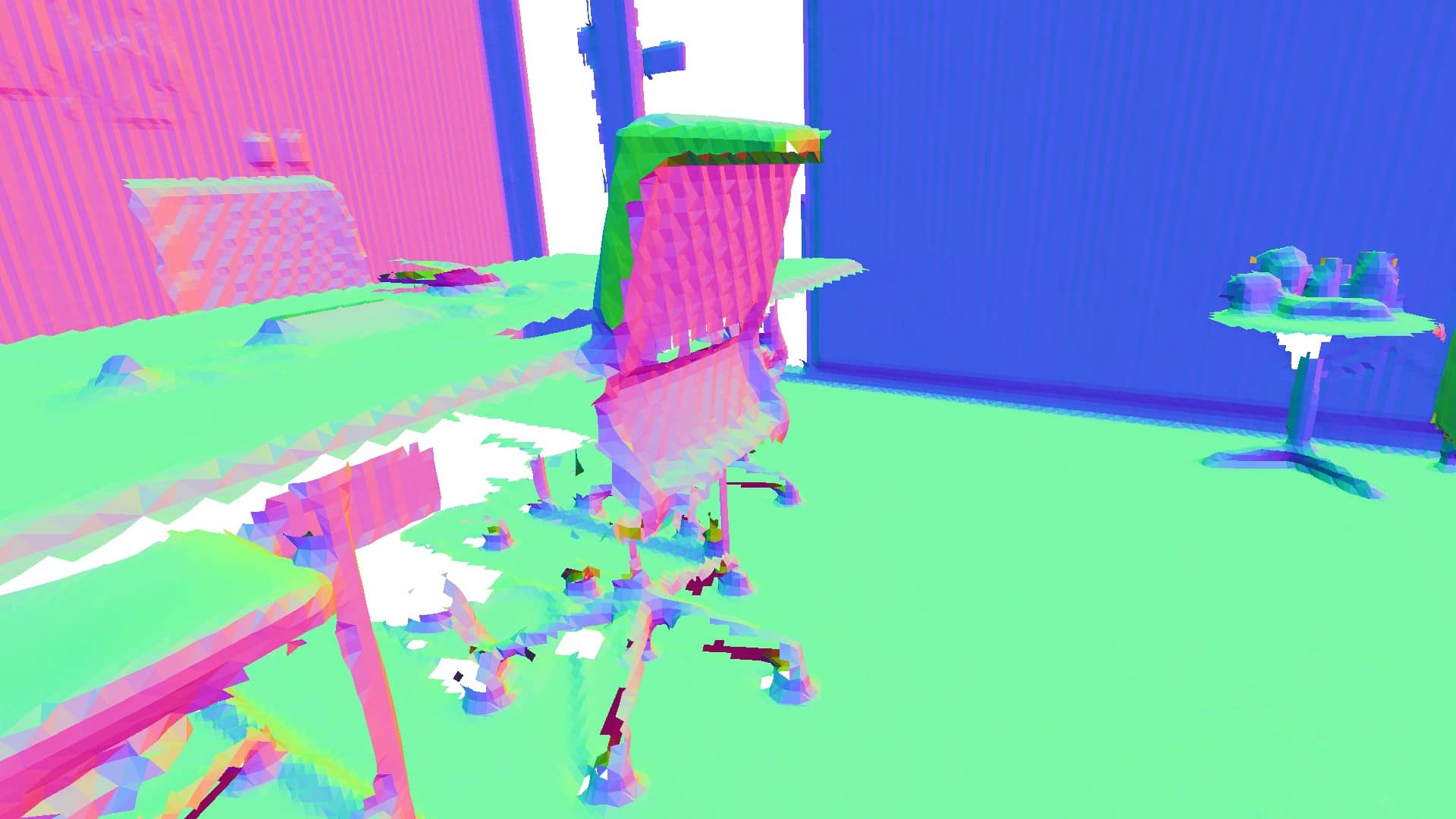}
  }
  \hspace{-11pt}
  \subfigure[iSDF]{
    \centering
    \includegraphics[width=0.15\textwidth]{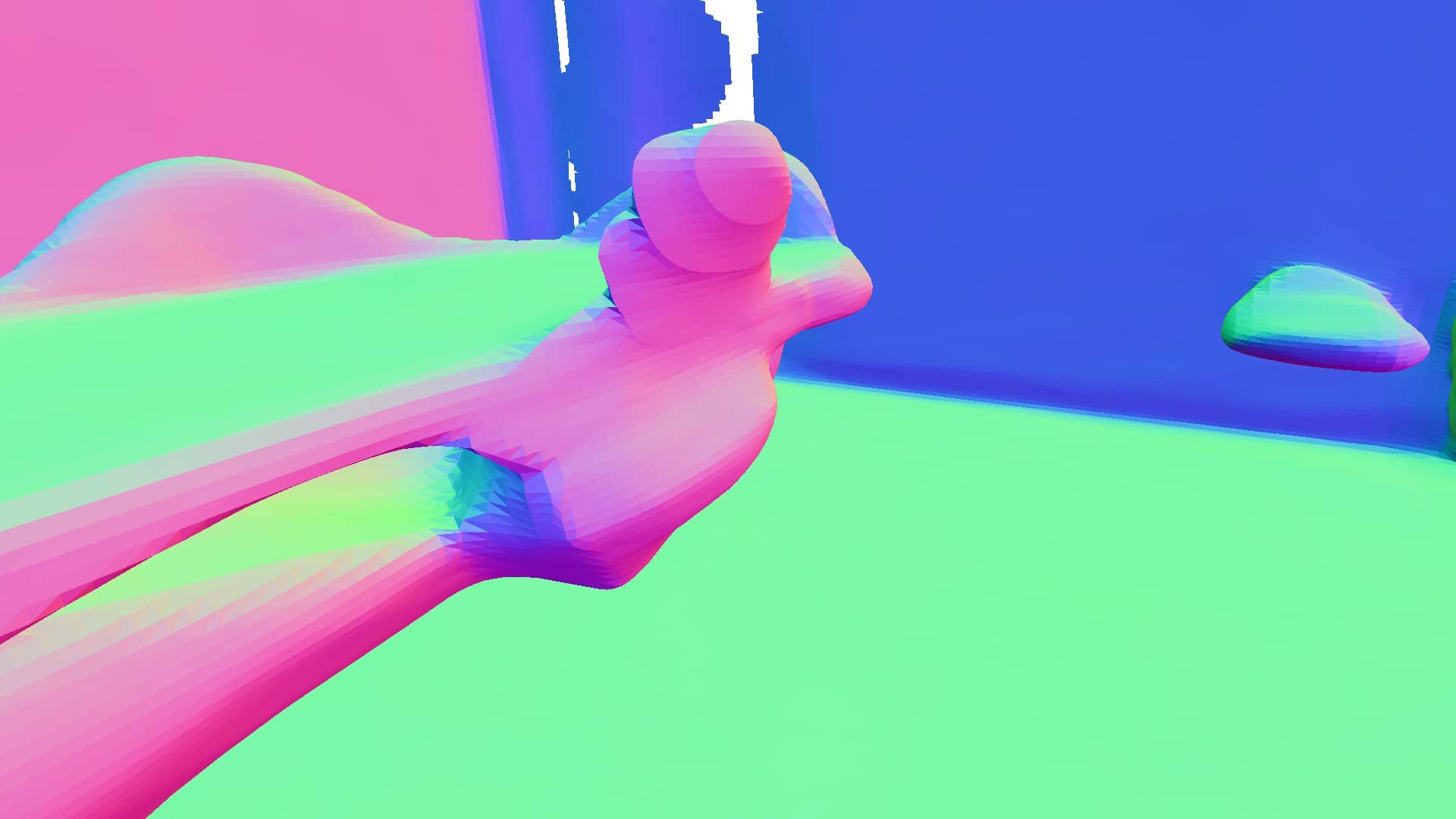}
  }
  \hspace{-11pt}
  \subfigure[SHINE-Mapping]{
    \centering
    \includegraphics[width=0.15\textwidth]{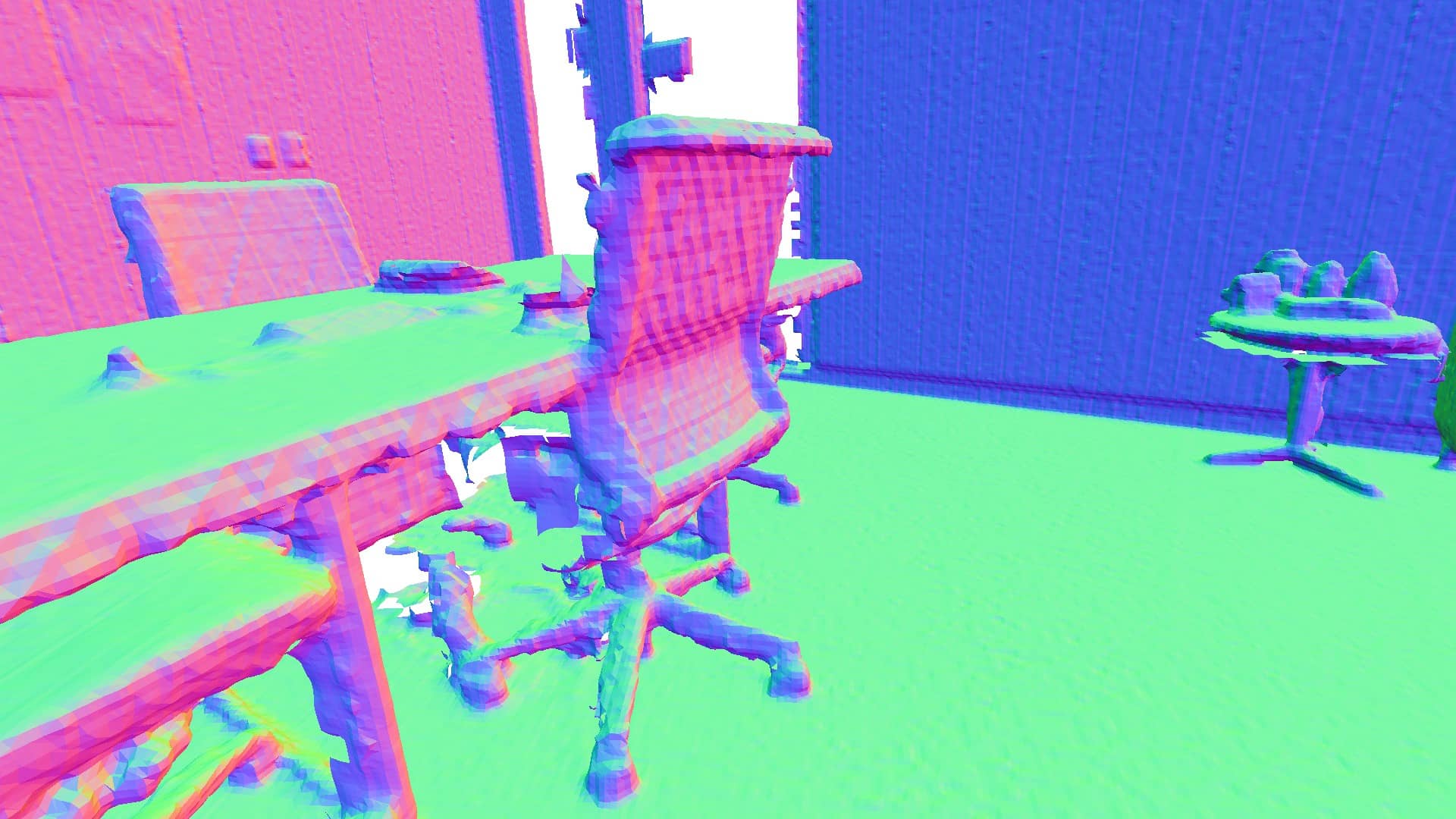}
  }

  \vspace{-6pt}

  \subfigure[2DGS]{
    \centering
    \includegraphics[width=0.15\textwidth]{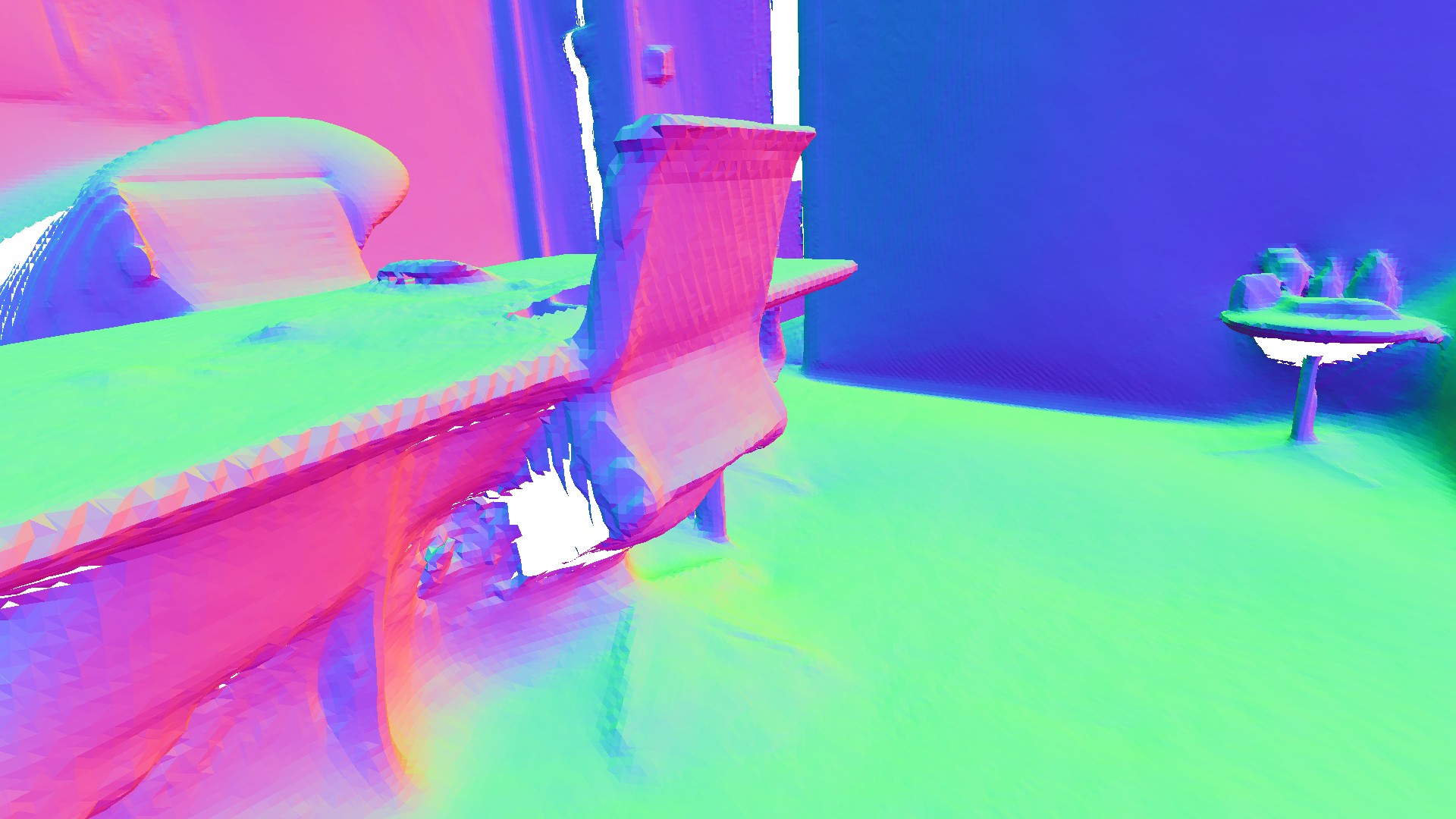}
  }
  \hspace{-11pt}
  \subfigure[PGSR]{
    \centering
    \includegraphics[width=0.15\textwidth]{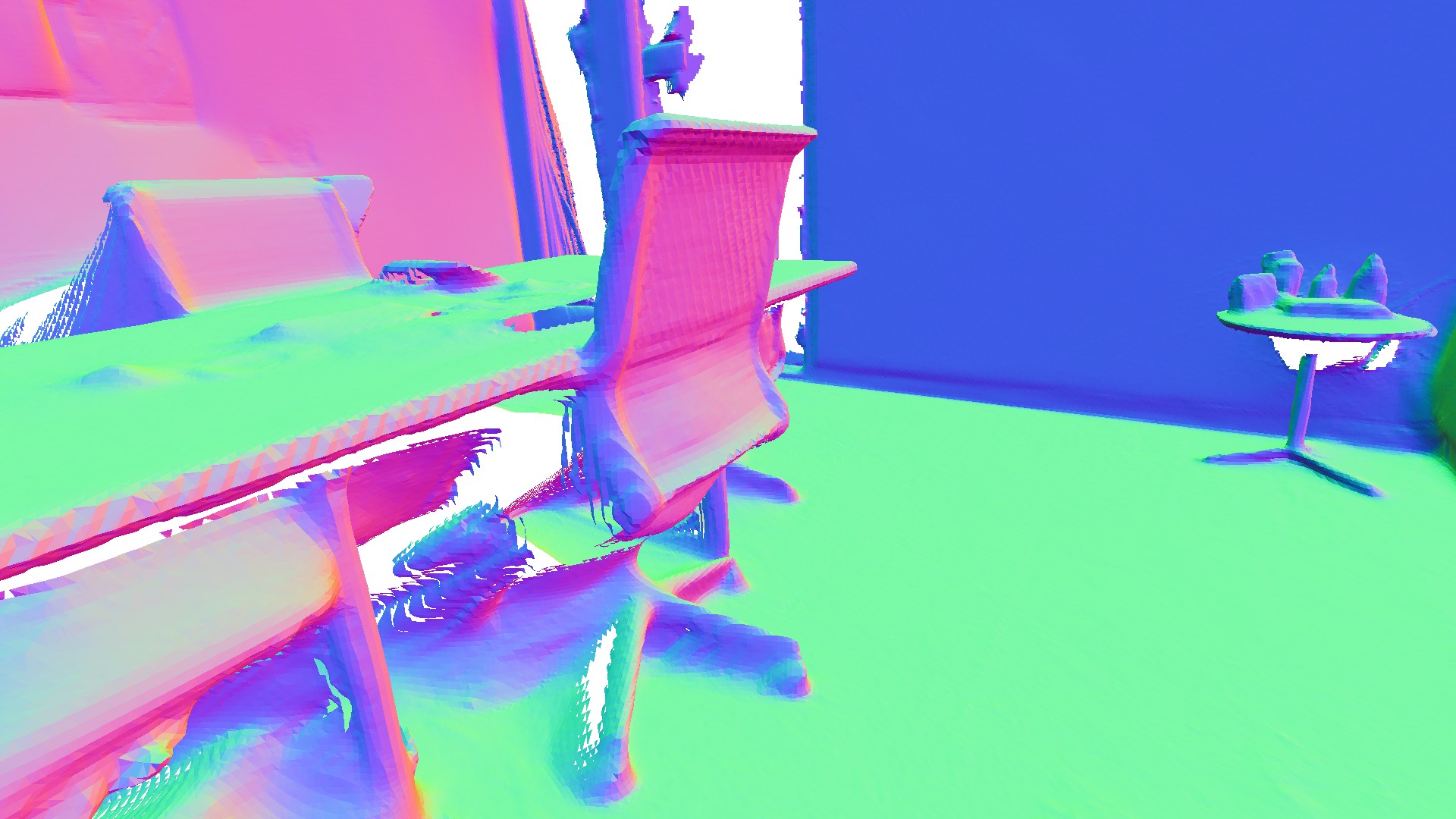}
  }
  \hspace{-11pt}
  \subfigure[H2Mapping]{
    \centering
    \includegraphics[width=0.15\textwidth]{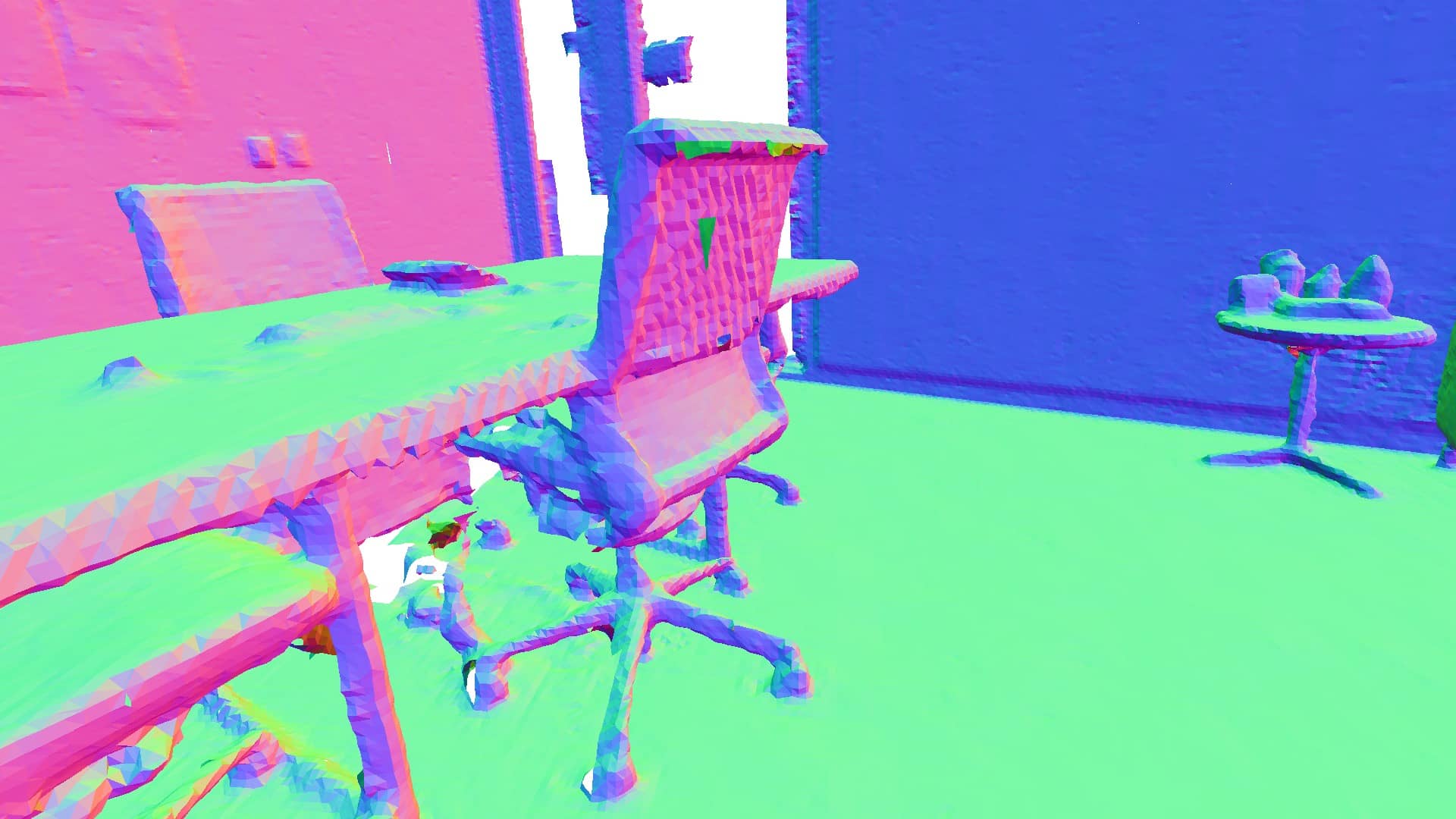}
  }

  \vspace{-6pt}
  \subfigure[M2Mapping]{
    \centering
    \includegraphics[width=0.15\textwidth]{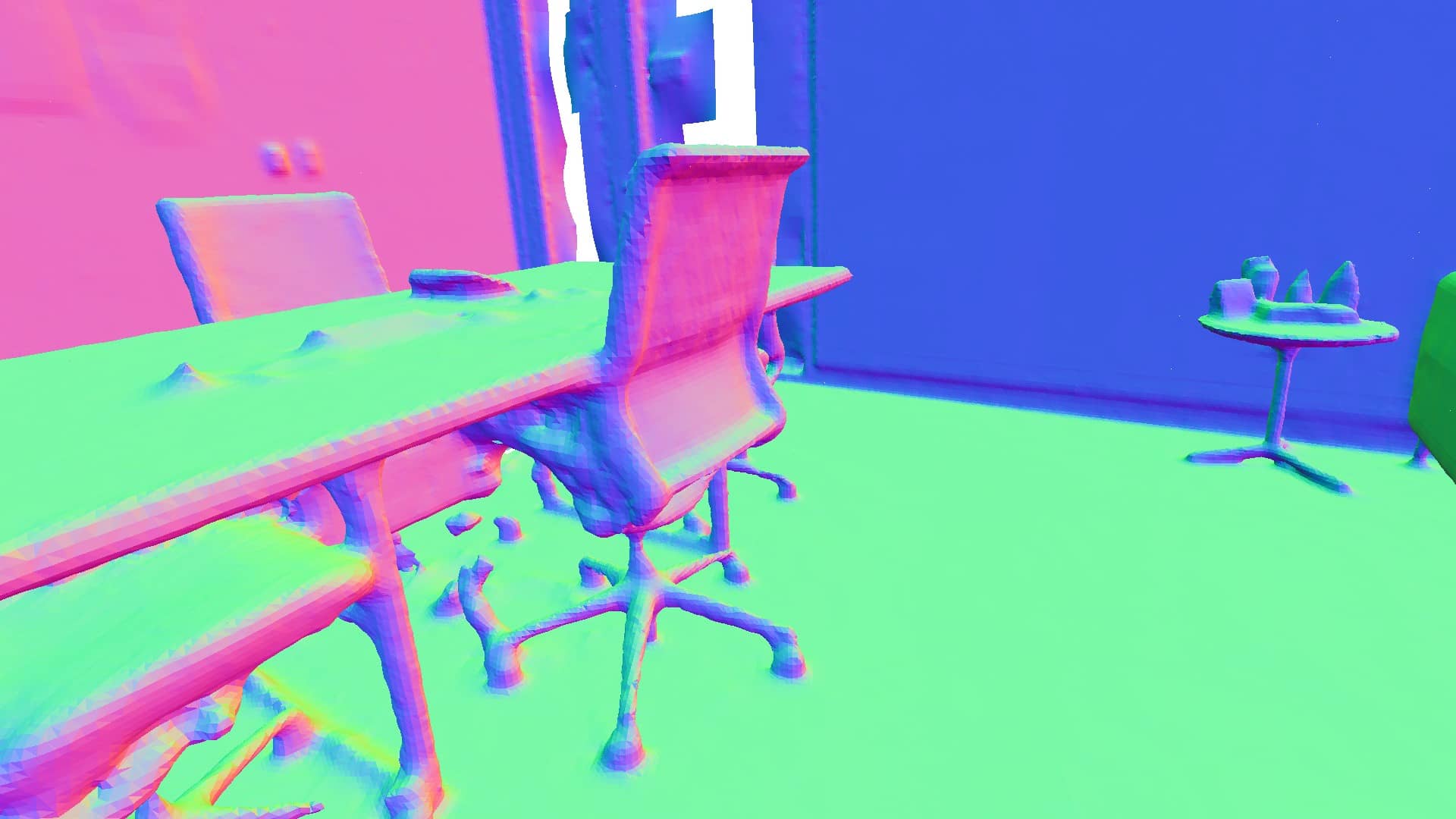}
  }
  \hspace{-11pt}
  \subfigure[Ours]{
    \centering
    \includegraphics[width=0.15\textwidth]{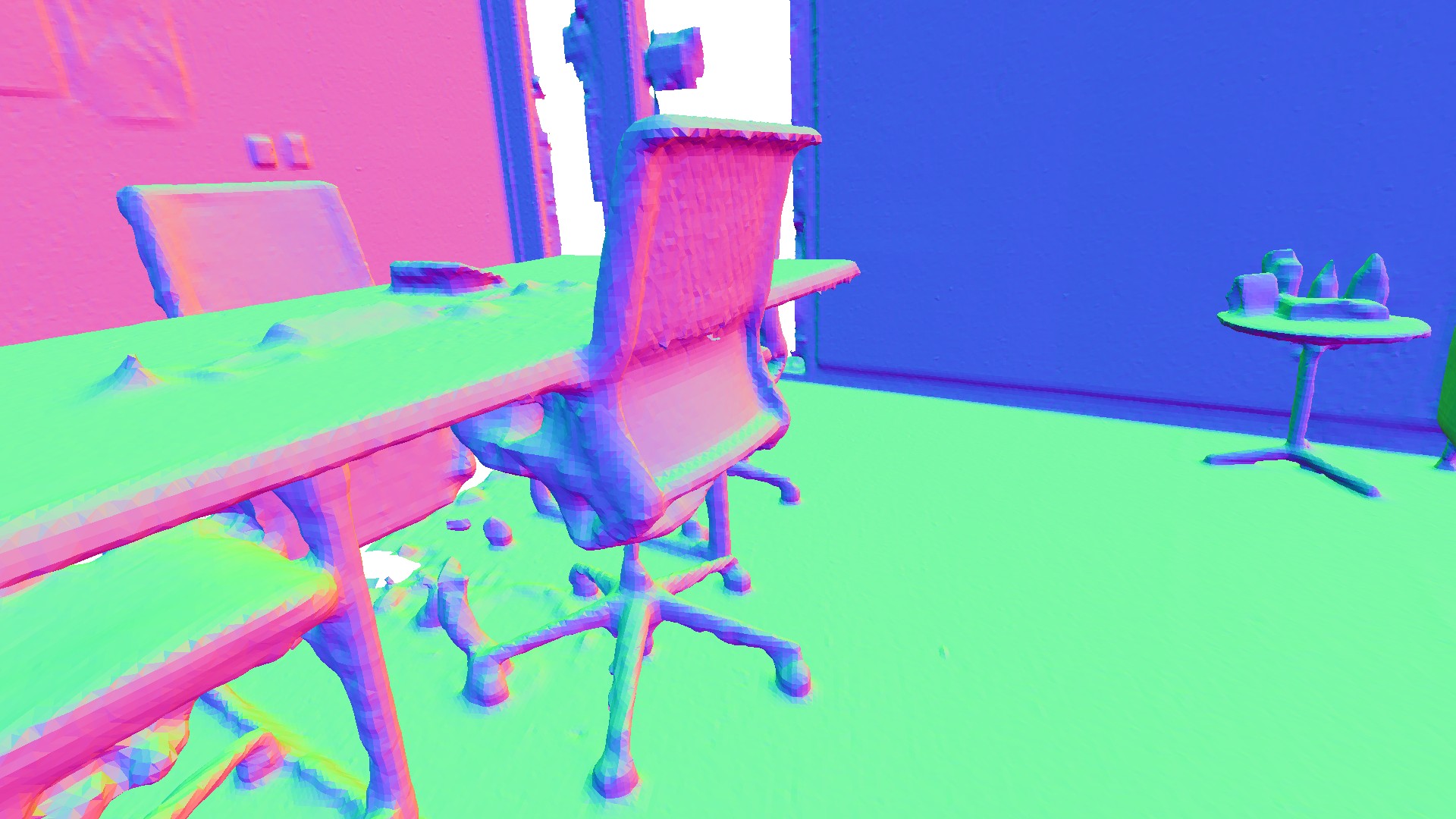}
  }
  \hspace{-11pt}
  \subfigure[Ground Truth]{
    \centering
    \includegraphics[width=0.15\textwidth]{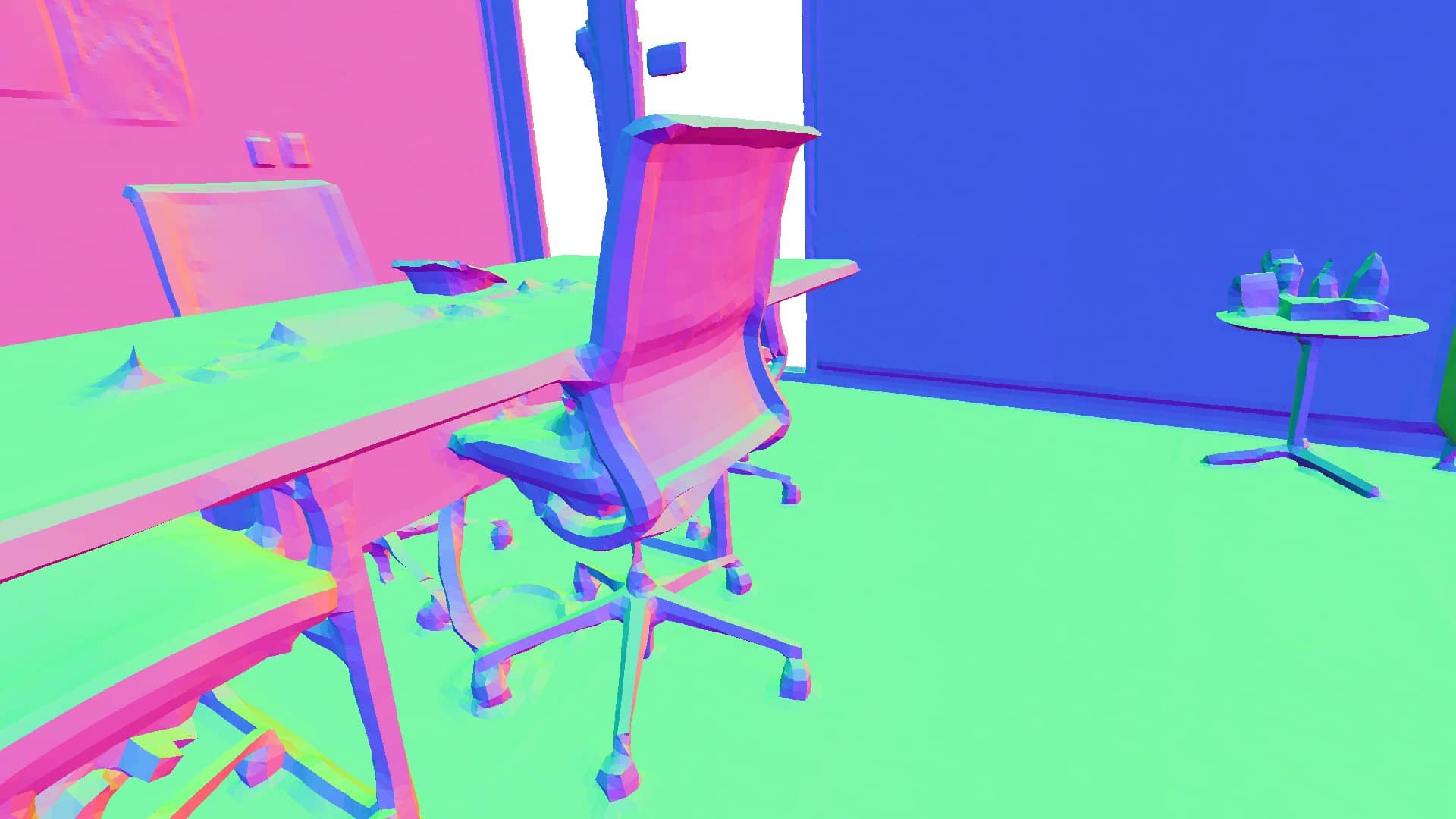}
  }

  \caption{
    Reconstructed mesh on the Replica dataset's Office-2 and colors indicate the direction of the surface normal.
  Our method and M2Mapping can capture precise geometric details in slim objects.}
  \label{fig:replica_qual_mesh}
  \vspace{-16pt}
\end{figure}

\begin{figure}[!h]
  \centering
  \setlength{\subfigcapskip}{-3pt} 
  \subfigure[InstantNGP]{
    \centering
    \includegraphics[width=0.15\textwidth]{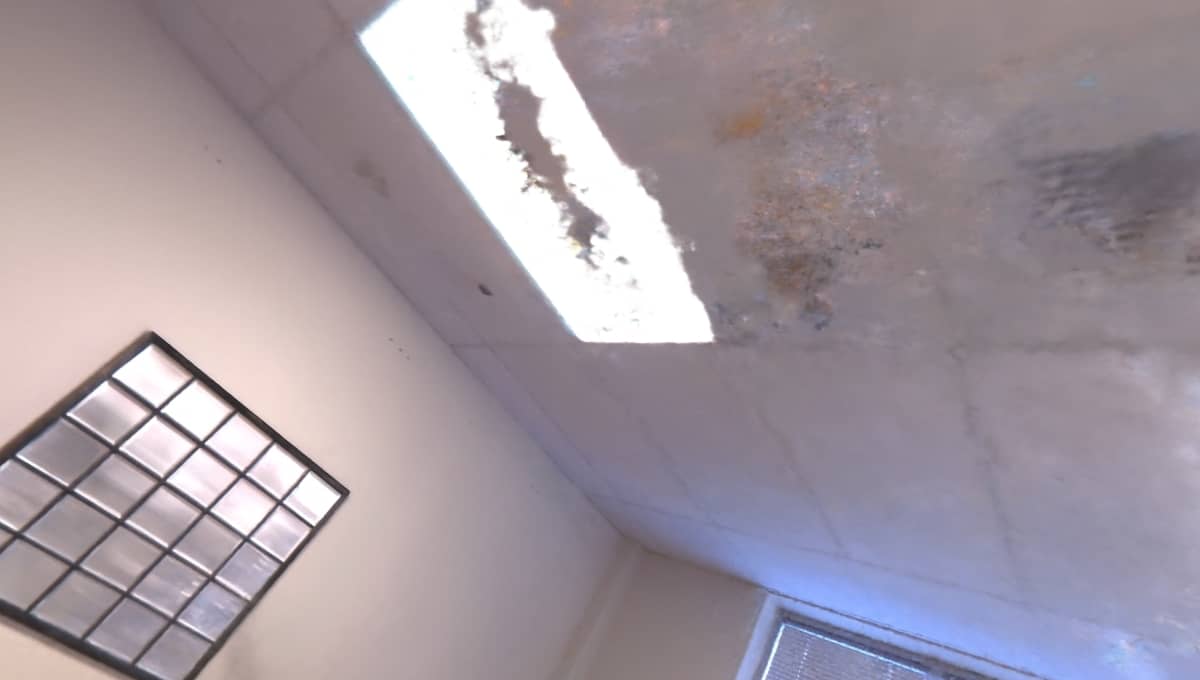}
  }
  \hspace{-11pt}
  \subfigure[3DGS]{
    \centering
    \includegraphics[width=0.15\textwidth]{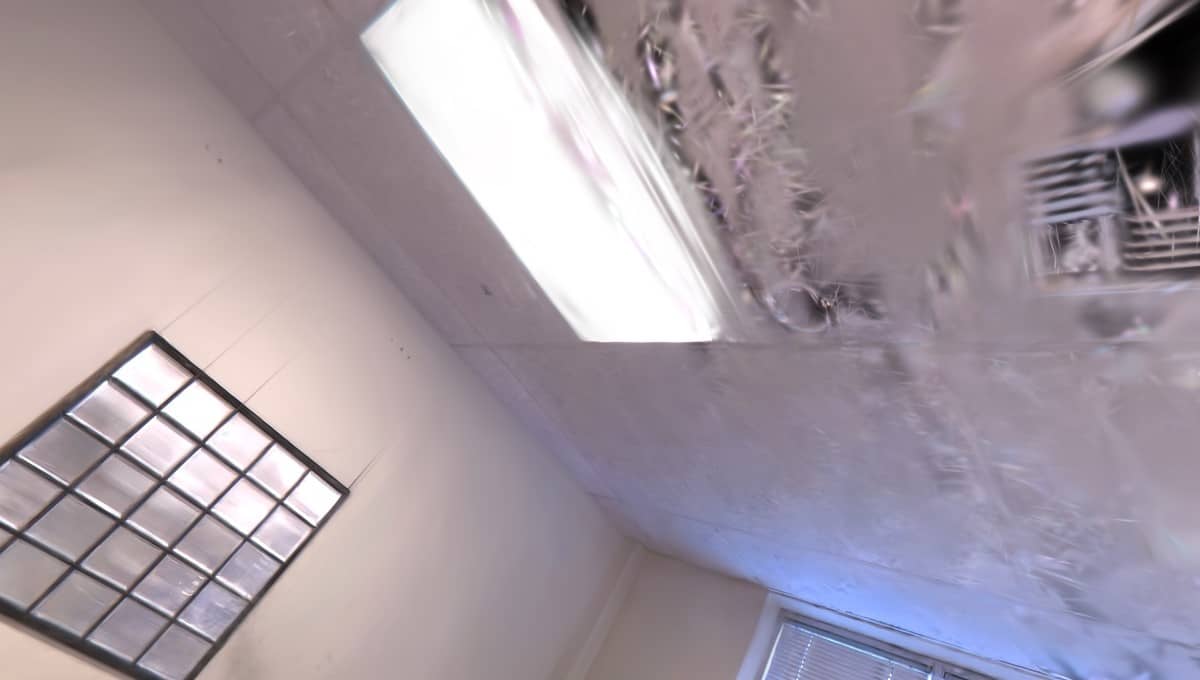}
  }
  \hspace{-11pt}
  \subfigure[2DGS]{
    \centering
    \includegraphics[width=0.15\textwidth]{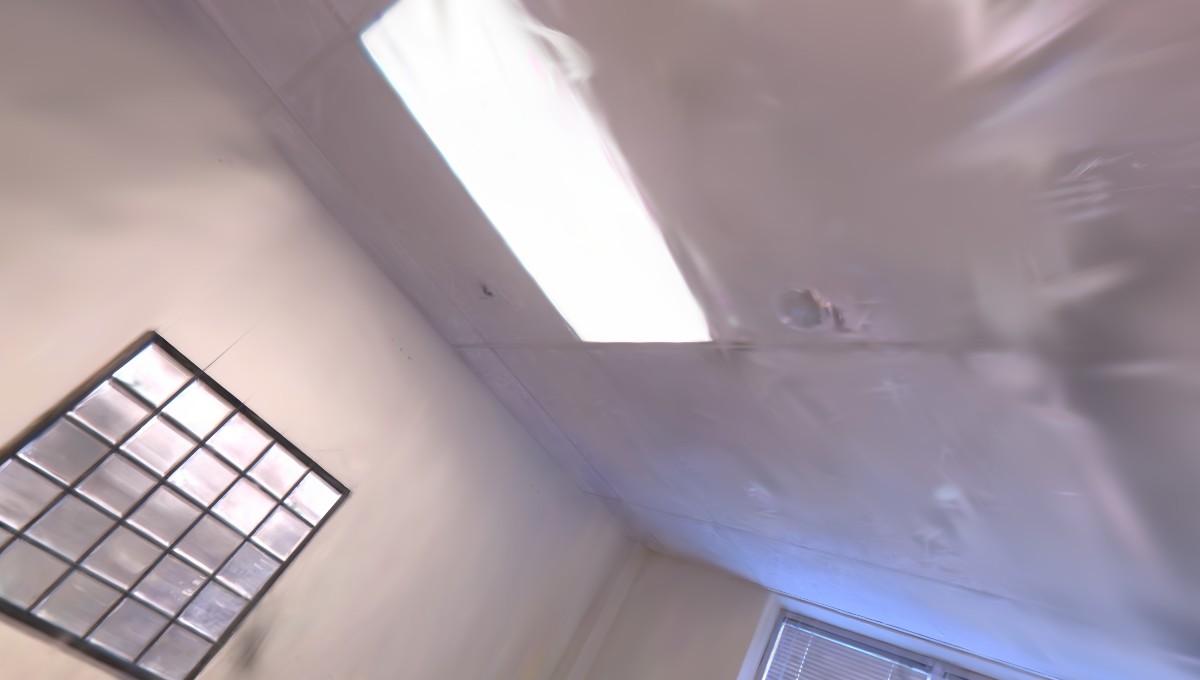}
  }

  \vspace{-6pt}

  \subfigure[PGSR]{
    \centering
    \includegraphics[width=0.15\textwidth]{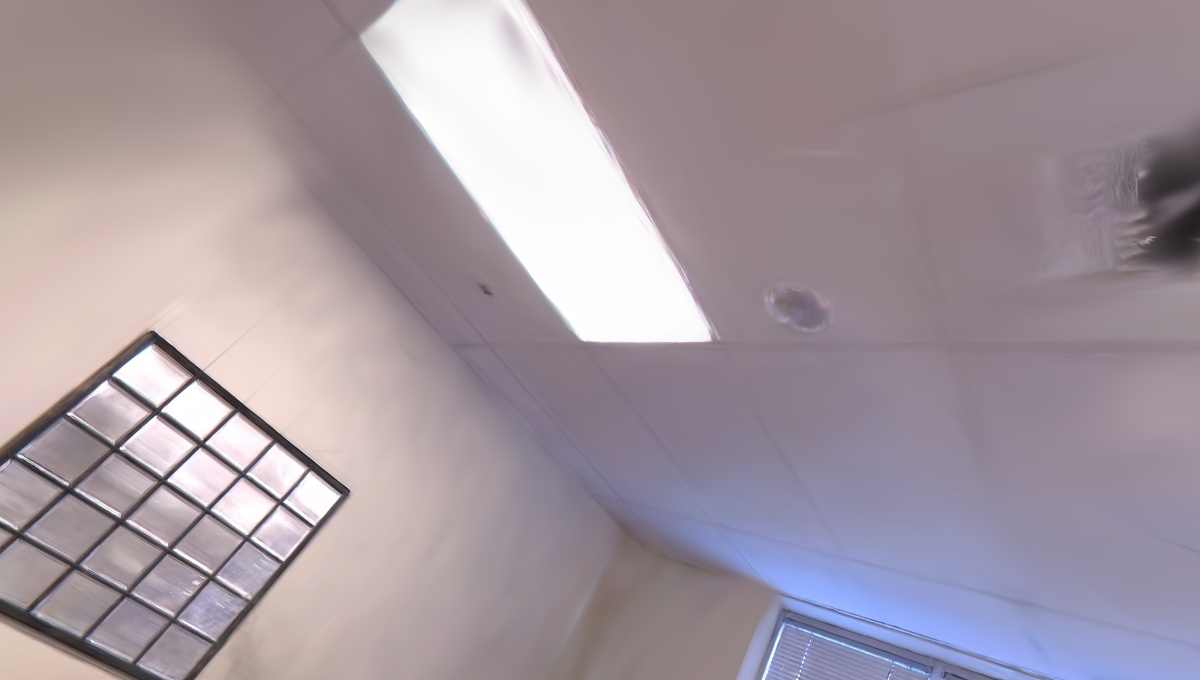}
  }
  \hspace{-11pt}
  \subfigure[MonoGS]{
    \centering
    \includegraphics[width=0.15\textwidth]{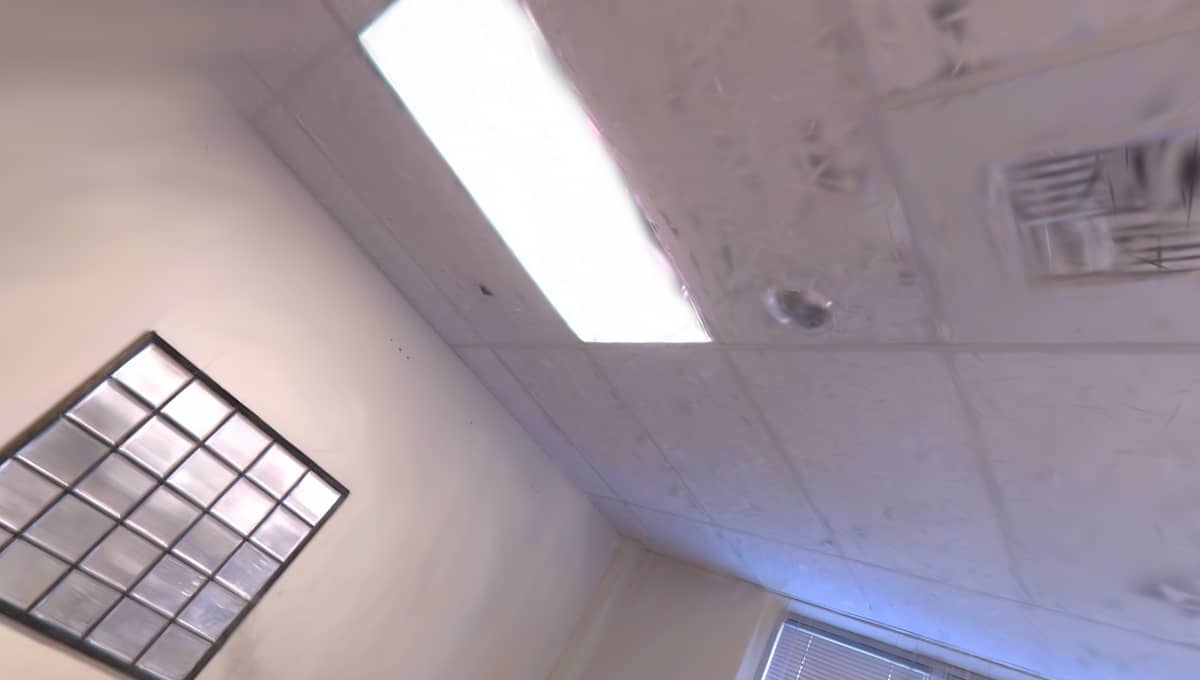}
  }
  \hspace{-11pt}
  \subfigure[H2Mapping]{
    \centering
    \includegraphics[width=0.15\textwidth]{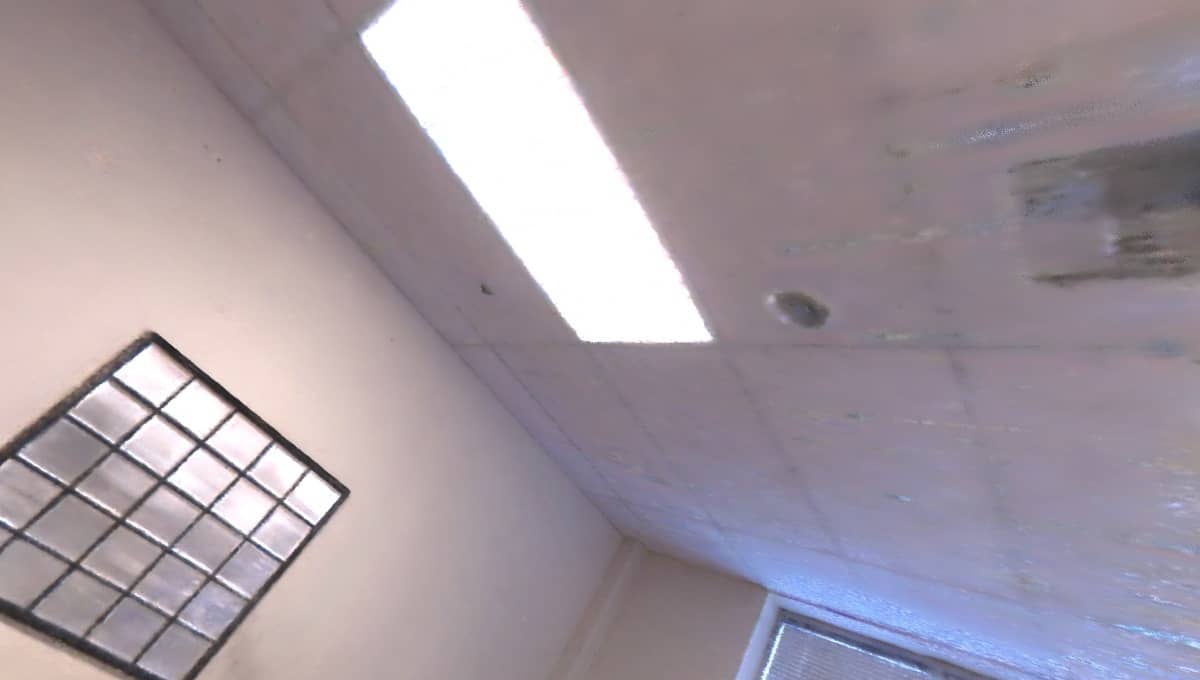}
  }

  \vspace{-6pt}

  \subfigure[M2Mapping]{
    \centering
    \includegraphics[width=0.15\textwidth]{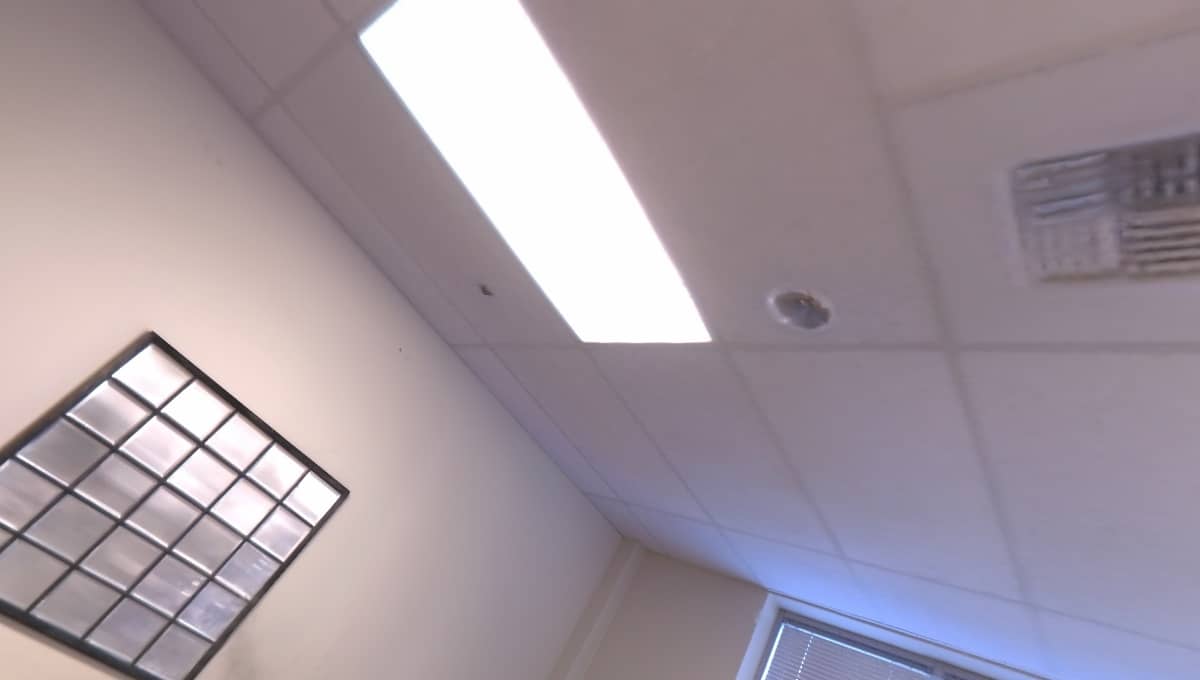}
  }
  \hspace{-11pt}
  \subfigure[Ours]{
    \centering
    \includegraphics[width=0.15\textwidth]{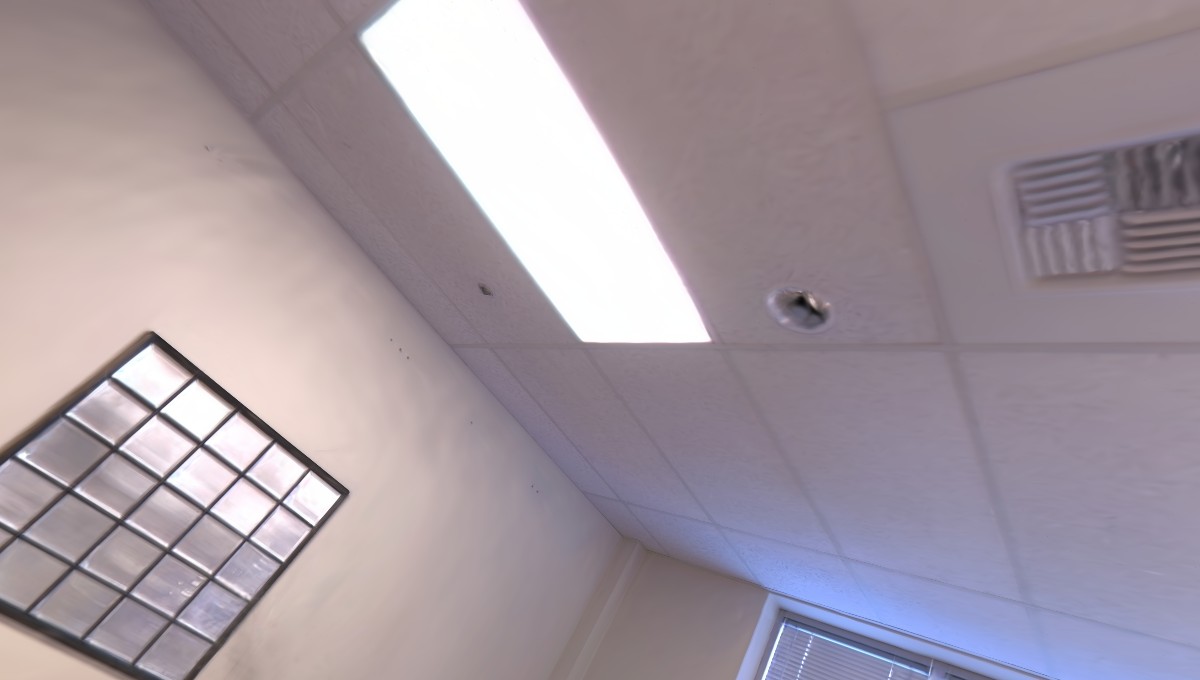}
  }
  \hspace{-11pt}
  \subfigure[Ground Truth]{
    \centering
    \includegraphics[width=0.15\textwidth]{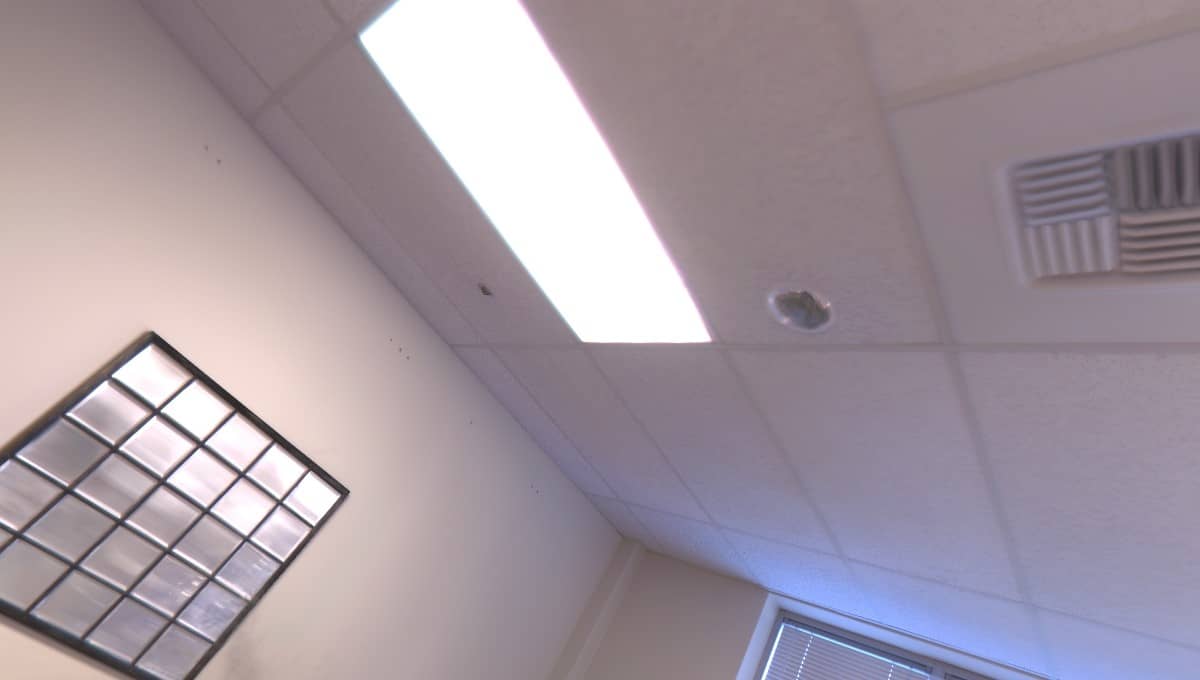}
  }

  \caption{
    Extrapolation rendering results on the Replica dataset's room-0.
  Our method can preserve great structure and texture details, while the NeRF-based method, M2Mapping, retains better consistency on the left wall.}
  \label{fig:replica_qual_render}
  \vspace{-12pt}
\end{figure}

\subsection{Experiments Settings}

\subsubsection{Baselines}
We compare our method with other prevailing approaches.
For pure surface reconstruction methods, we include the voxel-based method VDBFusion \cite{vizzo2022vdbfusion}, and the neural-network-based method SHINE-Mapping \cite{zhong2022shine}.
For novel view synthesis methods, we include the NeRF-based method: InstantNGP \cite{muller2022instant}, and the 3DGS-based methods initialized with the LiDAR point clouds: 3D Gaussian Splatting \cite{kerbl20233d}, 2D Gaussian Splatting \cite{huang20242d}, and PGSR\cite{chen2024pgsr}.
For depth-aided methods, we include RGBD-based methods: H2Mapping \cite{jiang2023h} and MonoGS \cite{GSSLAM2024}, and LiDAR-augmented NeRF-based method M2Mapping\cite{liu2024neural}.

\subsubsection{Metrics}
To evaluate the quality of the geometry, we recover the NSDF to a triangular mesh using marching cubes \cite{lorensen1987marching} and calculate the Chamfer Distance (C-L1, cm) and F-Score ($<2$ cm, \%) with the ground truth.
For rendering quality evaluation, we use the Structural Similarity Index (SSIM) and Peak Signal-to-Noise Ratio (PSNR) to compare the rendered image with the ground truth image.

\begin{figure*}[t]
  \centering
  \setlength{\subfigcapskip}{-3pt} 

  \subfigure{
    \centering
    \includegraphics[width=0.23\textwidth]{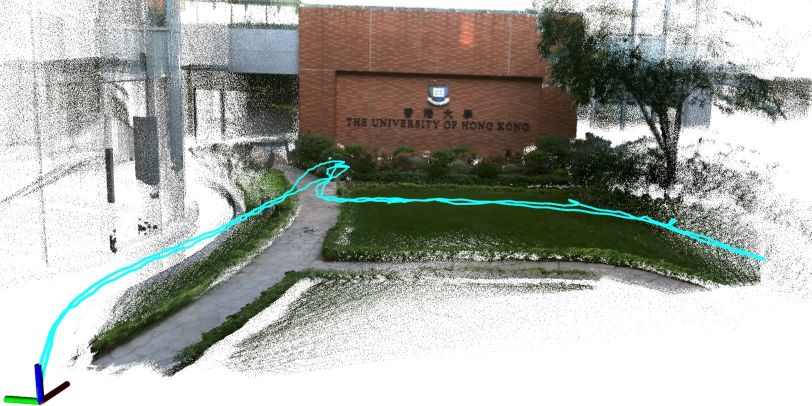}
  }
  \subfigure{
    \centering
    \includegraphics[width=0.23\textwidth]{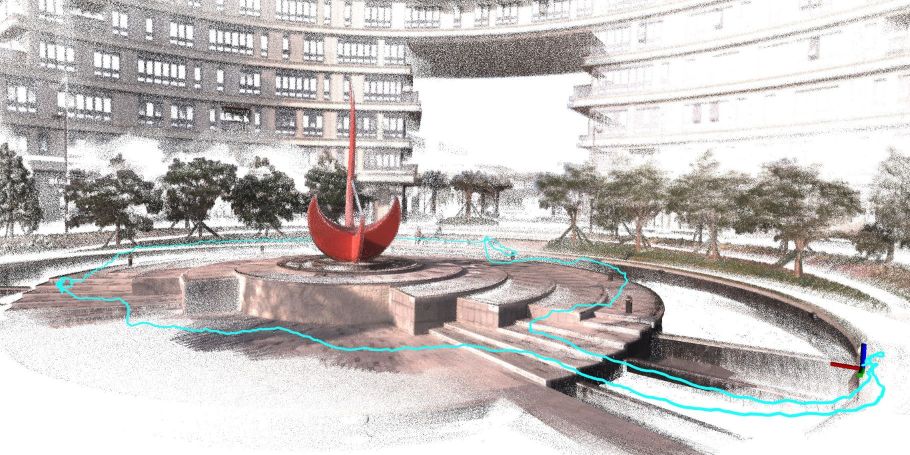}
  }
  \subfigure{
    \centering
    \includegraphics[width=0.23\textwidth]{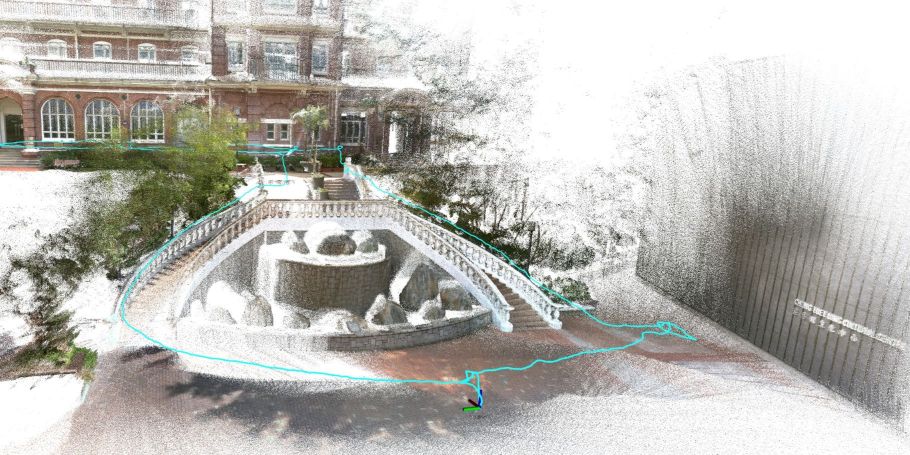}
  }
  \subfigure{
    \centering
    \includegraphics[width=0.23\textwidth]{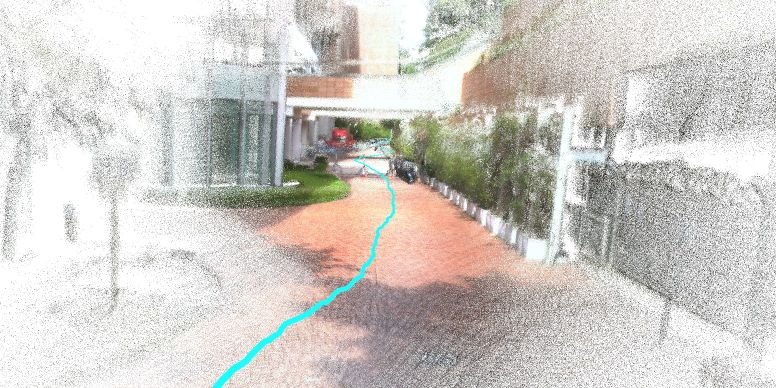}
  }

  \vspace{-10pt}

  \subfigure{
    \centering
    \includegraphics[width=0.23\textwidth]{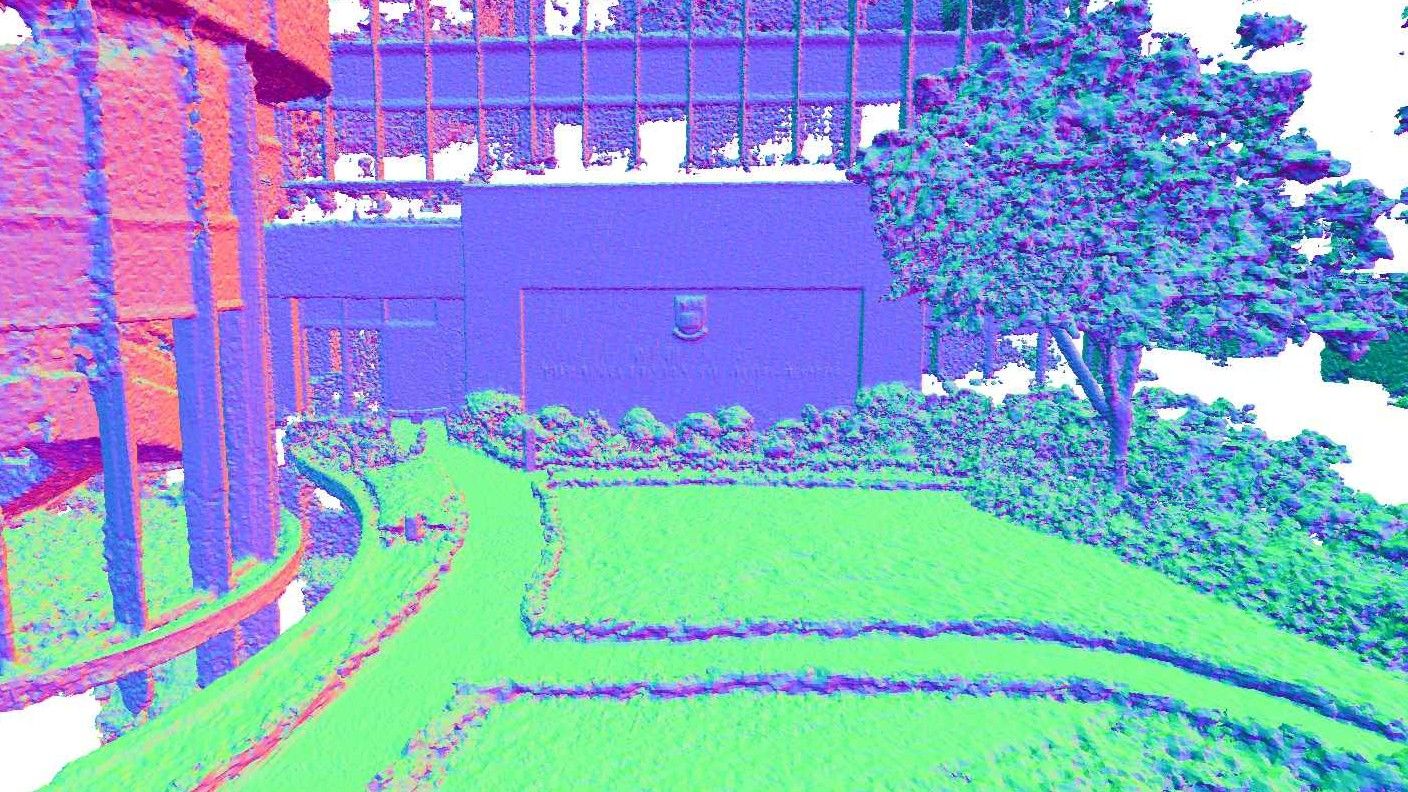}
  }
  \subfigure{
    \centering
    \includegraphics[width=0.23\textwidth]{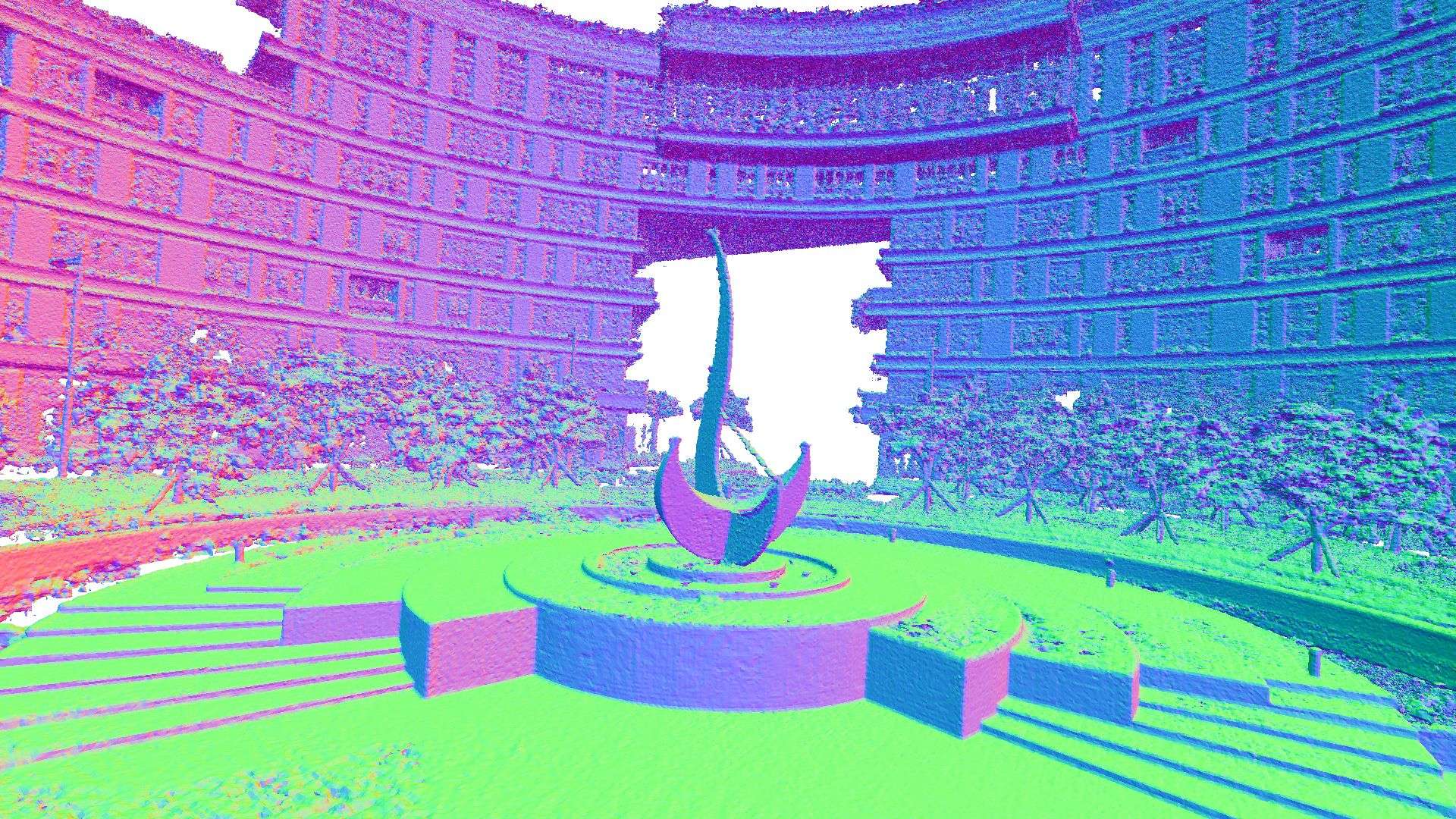}
  }
  \subfigure{
    \centering
    \includegraphics[width=0.23\textwidth]{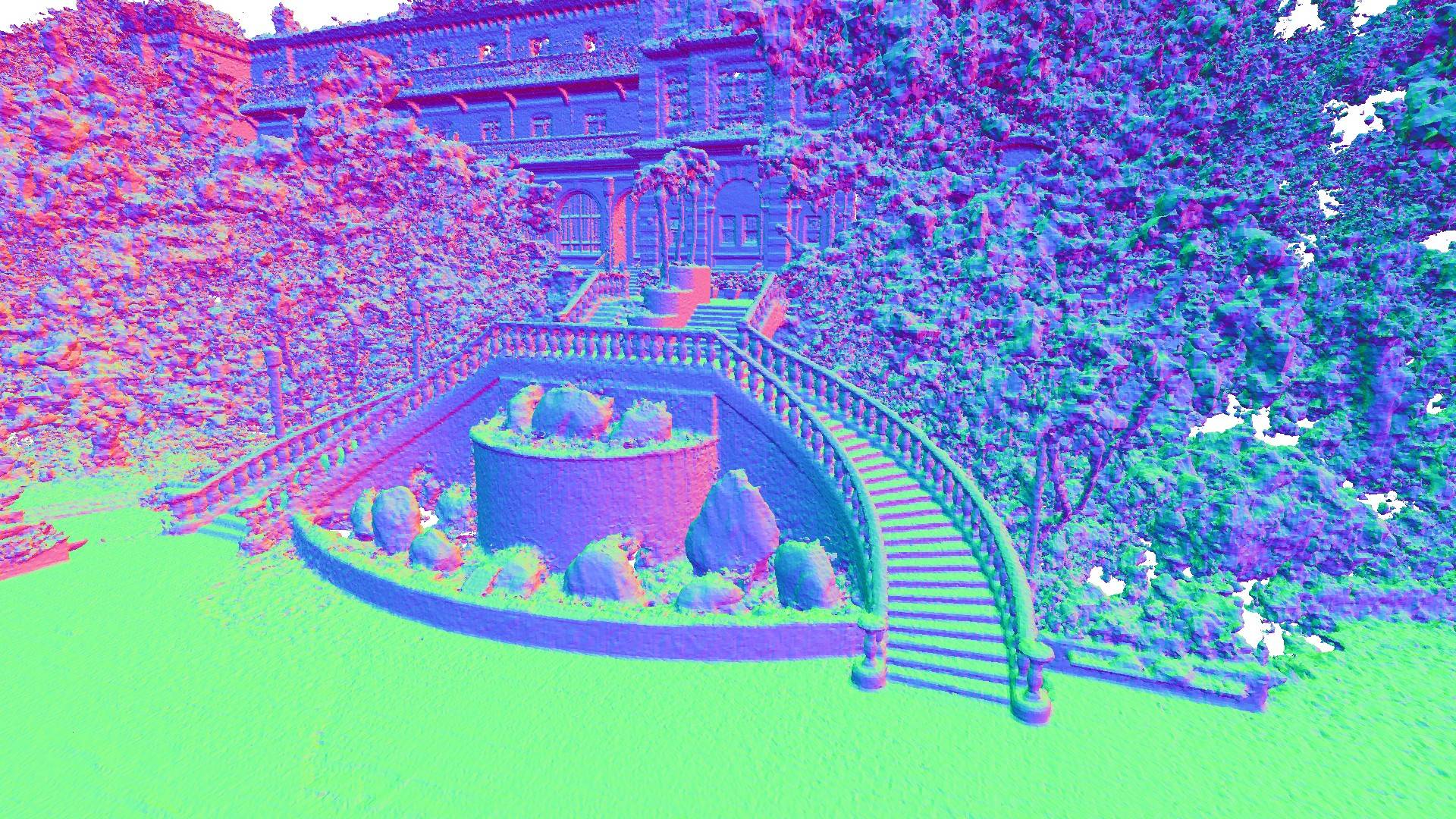}
  }
  \subfigure{
    \centering
    \includegraphics[width=0.23\textwidth]{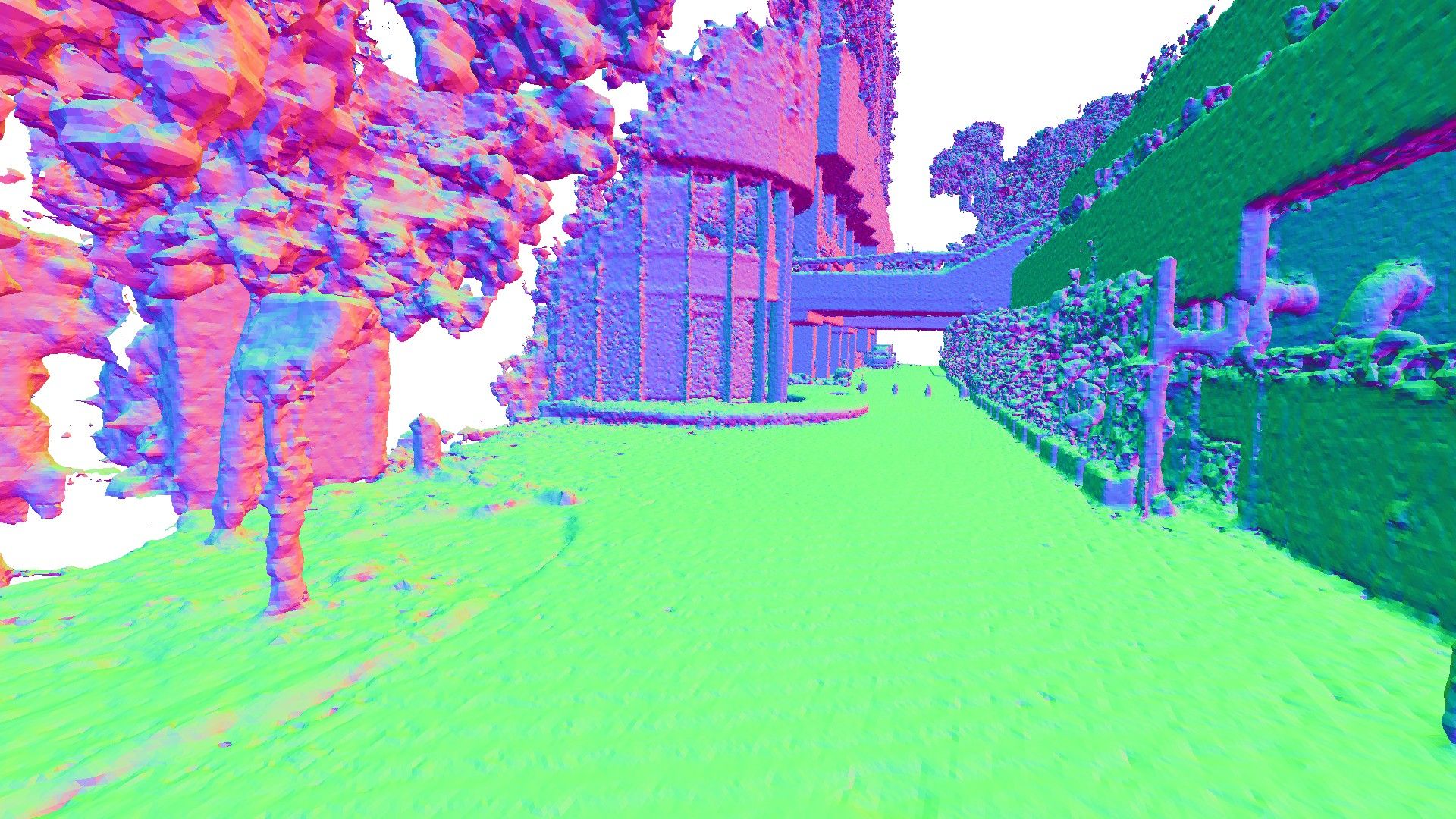}
  }

  \vspace{-10pt}

  \setcounter{subfigure}{0}
  \subfigure[Campus]{
    \centering
    \includegraphics[width=0.23\textwidth]{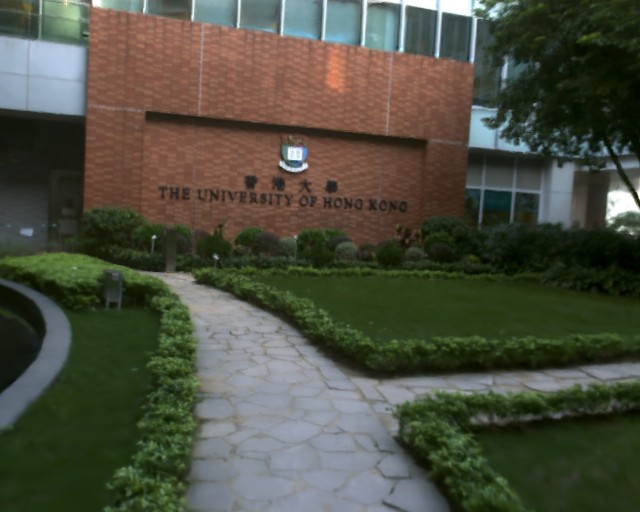}
  }
  \subfigure[Sculture]{
    \centering
    \includegraphics[width=0.23\textwidth]{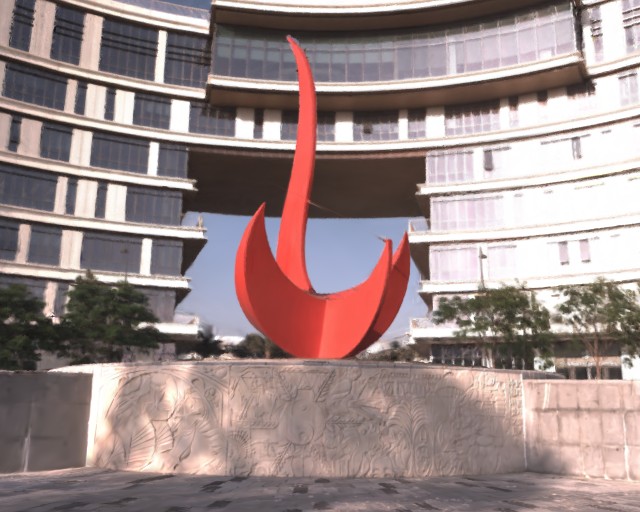}
  }
  \subfigure[Culture]{
    \centering
    \includegraphics[width=0.23\textwidth]{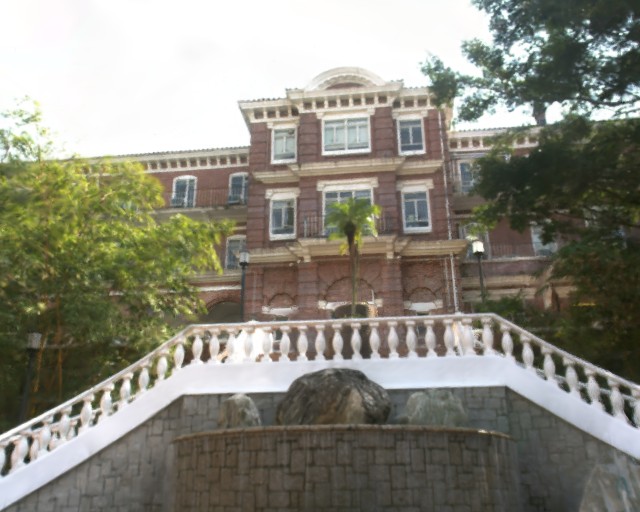}
  }
  \subfigure[Drive]{
    \centering
    \includegraphics[width=0.23\textwidth]{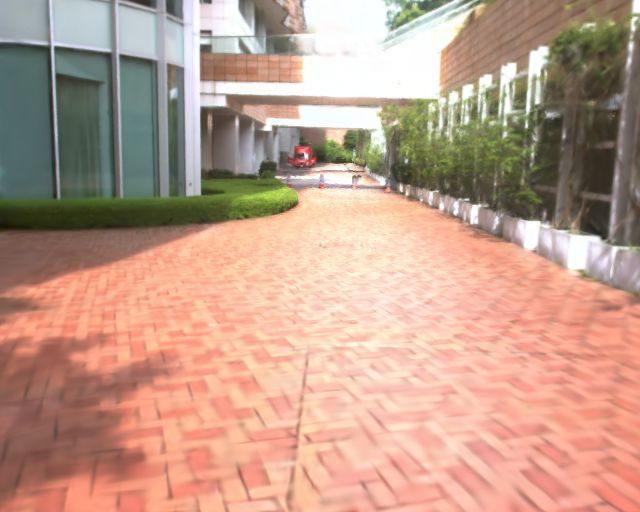}
  }

  \vspace{-6pt}

  \subfigure{
    \centering
    \includegraphics[width=0.48\textwidth]{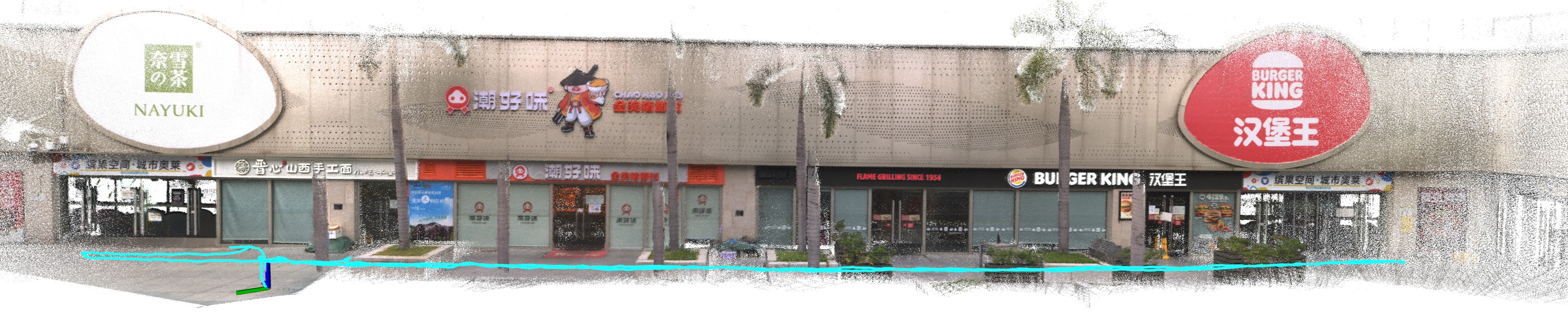}
  }
  \subfigure{
    \centering
    \includegraphics[width=0.48\textwidth]{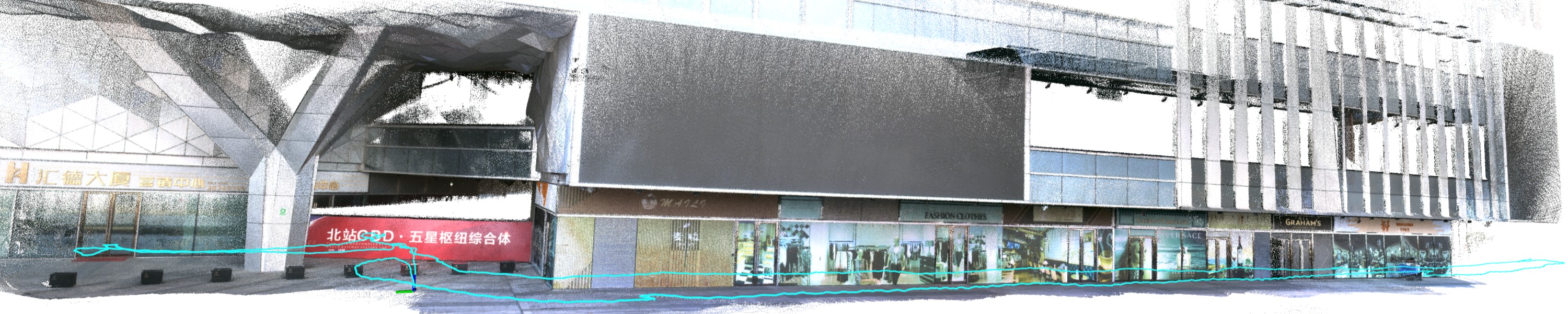}
  }

  \vspace{-10pt}

  \subfigure{
    \centering
    \includegraphics[width=0.48\textwidth]{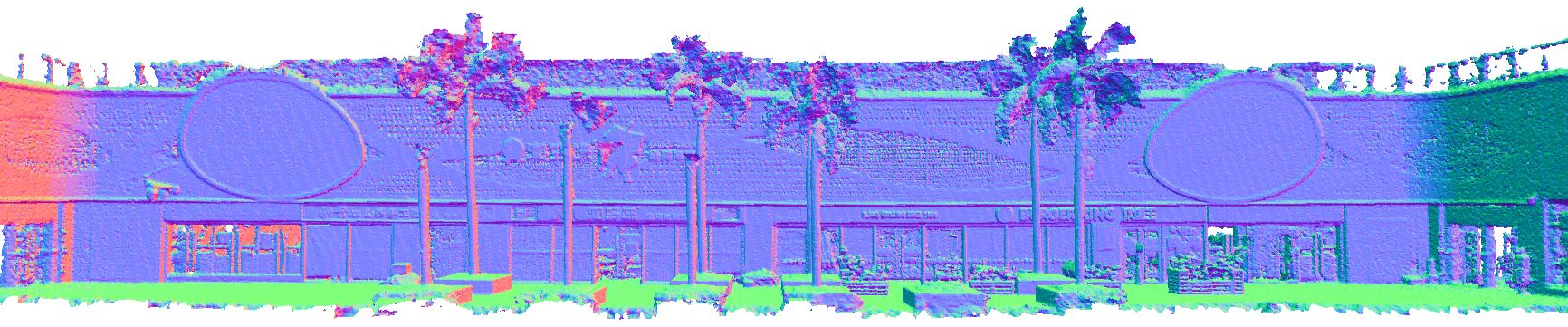}
  }
  \subfigure{
    \centering
    \includegraphics[width=0.48\textwidth]{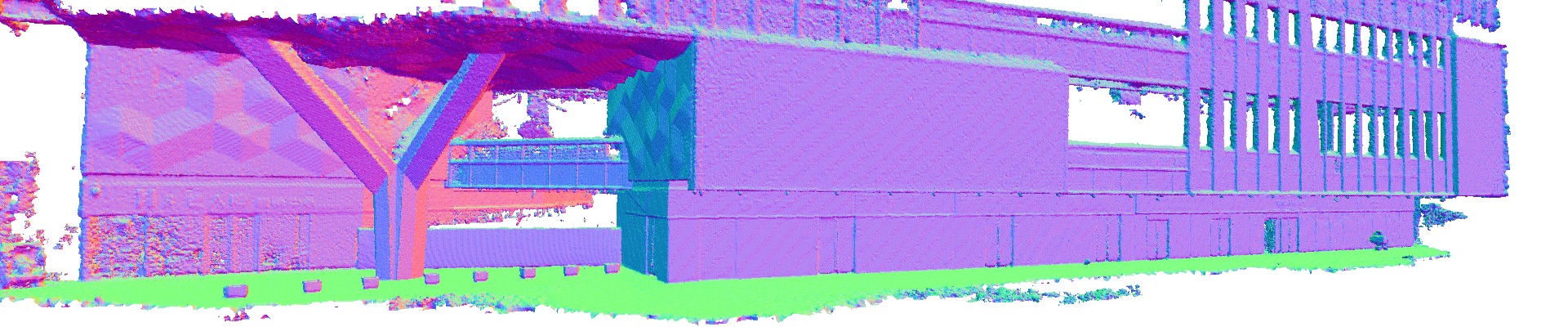}
  }

  \vspace{-10pt}

  \subfigure[Station]{
    \centering
    \includegraphics[width=0.48\textwidth]{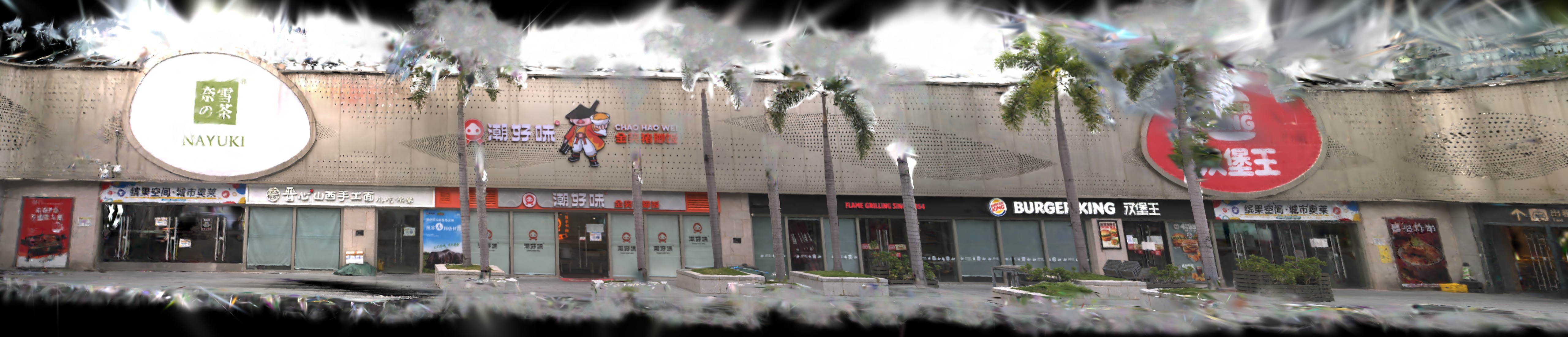}
  }
  \subfigure[CBD]{
    \centering
    \includegraphics[width=0.48\textwidth]{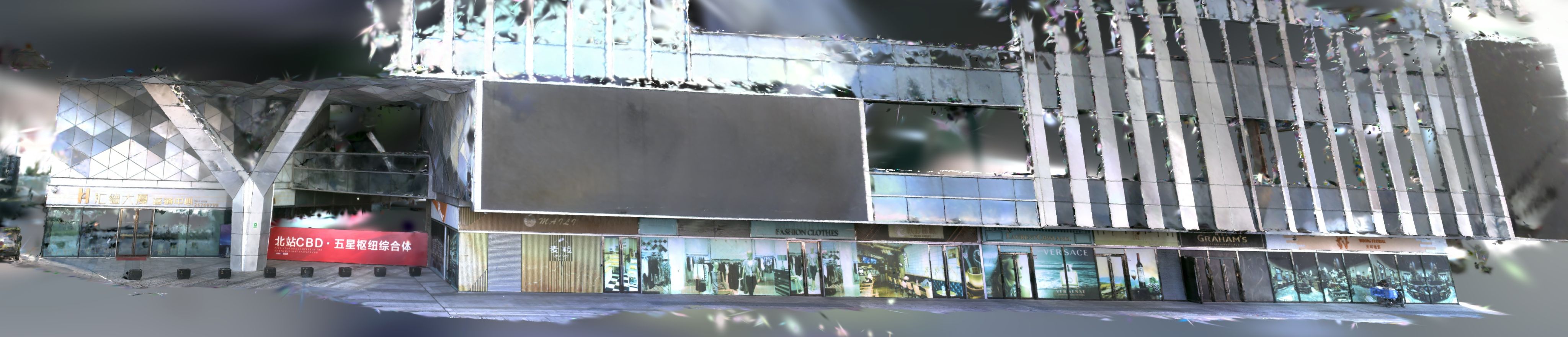}
  }

  \caption{We show our surface reconstruction (color indicates the normal direction) and rendering results on the FAST-LIVO2 Datasets collected by real-world LiDAR-visual sensor systems under different trajectories, where from top to bottom are collected point clouds, reconstructed meshes, and novel view synthesis by Gaussian splatting.}
  \label{fig:fast_livo}
  \vspace{-12pt}
\end{figure*}

\subsection{Comparative Study}

\subsubsection{Replica Datasets}
The Replica datasets \cite{straub2019replica} provide room-scale indoor simulation RGBD sensor data.
We follow the M2Mapping\cite{liu2024neural} to emphasize the issues of extrapolation rendering consistency with extrapolation evaluation datasets generated by uniformly sampling positions and orientations in each scene from Replica.
The quantitative results, comparing surface reconstruction, interpolation, and extrapolation rendering, are shown in Tab.~\ref{tab:replica_quan_inter}. Our method shows competitive surface reconstruction results with the state of the art, and the best performance in terms of interpolation rendering, but fails to compete with the extrapolation rendering results of the NeRF-based M2Mapping.
We validate that our method is capable of capturing more precise geometric details in slim objects, as shown in Fig.~\ref{fig:replica_qual_mesh}.
In the extrapolation views shown in Fig.~\ref{fig:replica_qual_render}, our method can retain more high-frequency textures (sharper spots) while NeRF-based methods show more natural transitions (smoother lighting).

\subsubsection{FAST-LIVO2 Datasets}

For real-world scenes, we evaluate generalized types of trajectory datasets collected with a camera and a LiDAR and use the localization results from FAST-LIVO2 \cite{zheng2024fast} as the input poses, as shown in the top row of Fig.~\ref{fig:fast_livo_sysu} and Fig.~\ref{fig:fast_livo}.
We adopt a train-test split for interpolation rendering evaluation, where every 8th photo is used for the test sets and the rest for the train sets \cite{kerbl20233d}, and the quantitative results are shown in Tab.~\ref{tab:fast_livo_quan}.
Our method yields the best rendering performance overall, and also returns high-granularity surface reconstructions and considerable extrapolation rendering results across generalized scenes, as shown in Fig.~\ref{fig:fast_livo_sysu} and Fig.~\ref{fig:fast_livo}.

\begin{table}[!h]
  \centering
  \caption{Quantitative results on the FAST-LIVO2 datasets.}
  \label{tab:fast_livo_quan}
  \resizebox{0.48\textwidth}{!}{
    \begin{tabular}{cccccccccc}
      \toprule
      \textbf{Metrics} & \textbf{Methods} & \textbf{Campus} & \textbf{Sculpture} & \textbf{Culture} & \textbf{Drive} & \textbf{Station} & \textbf{SYSU} & \textbf{CBD} & \textbf{Avg.} \\
      \midrule

      \multirow{6}*{SSIM$\uparrow$}
      & {InstantNGP}
      & 0.789 & 0.698 & 0.670 & 0.697 & 0.780 & 0.792 & 0.779 & 0.744
      \\

      & {3DGS}
      & \underline{0.849} & \underline{{0.769}} & {{0.726}} & \underline{0.780} & \underline{0.853} & 0.789 & 0.825 & \underline{0.799}
      \\

      & {2DGS}
      & 0.839 & 0.730 & 0.611 & 0.741 & 0.798 & \underline{0.808} & 0.760 & 0.755\\

      & {PGSR}
      & 0.836 & 0.745 & 0.595 & 0.762 & 0.828 & 0.806 & 0.735 & 0.758\\

      & {M2Mapping}
      & {0.834} & {0.729} & \underline{0.727} & {0.764} & 0.809 & 0.789 & \underline{0.850} & 0.786
      \\
      & {Ours}
      & \textbf{0.858} & \textbf{0.774} & \textbf{0.797} & \textbf{0.788} & \textbf{0.861} & \textbf{0.841} & \textbf{0.892} & \textbf{0.830}\\
      \midrule

      \multirow{6}*{PSNR$\uparrow$}
      & {InstantNGP}
      & 28.880 & 22.356 & 21.563 & 24.145 & 24.111 & 22.517 & 22.468 & 23.720
      \\

      & {3DGS}
      & \underline{{31.310}} & \underline{24.128} & {21.764} & {25.837} & \underline{26.859} & \textbf{28.798} & 22.078 & 25.825
      \\
      & {2DGS}
      & 30.611 & 22.654 & 19.218 & 24.847 & 24.756 & 27.517 & 22.022 & 24.518
      \\
      & {PGSR}
      & 30.648 & 23.302 & 18.438 & 25.382 & 25.504 & 27.563 & 20.662 & 24.500
      \\

      & {M2Mapping}
      & {30.681} & {23.453} & \underline{24.695} & \textbf{25.941} & 25.859 & 28.011 & \underline{26.655} & \underline{26.471}
      \\
      & Ours
      & \textbf{31.697} & \textbf{24.196} & \textbf{25.074} & \underline{25.657} & \textbf{27.065} & \underline{28.532} & \textbf{26.696} & \textbf{26.988}\\
      \midrule
      \bottomrule
    \end{tabular}
  }
  \vspace{-16pt}
\end{table}

\begin{figure}[t]
  \centering
  \setlength{\subfigcapskip}{-3pt} 
  \setcounter{subfigure}{0}
  \subfigure[VDBFusion]{
    \centering
    \includegraphics[width=0.15\textwidth]{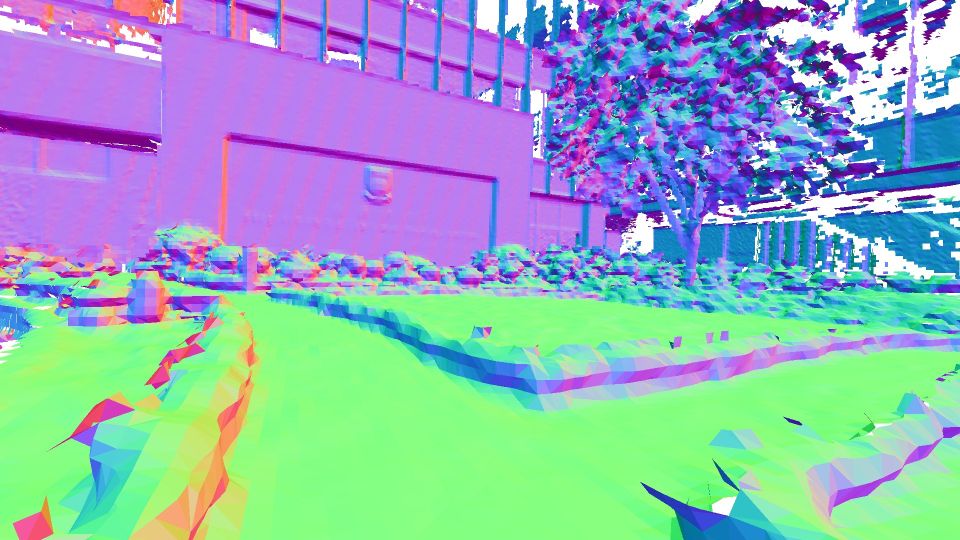}
  }
  \hspace{-12pt}
  \subfigure[SHINE-Mapping]{
    \centering
    \includegraphics[width=0.15\textwidth]{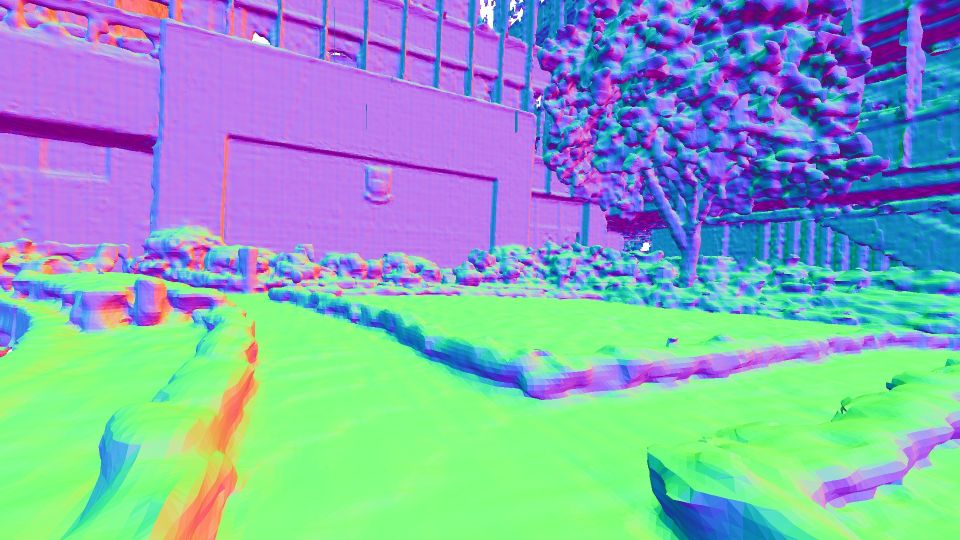}
  }
  \hspace{-12pt}
  \subfigure[2DGS]{
    \centering
    \includegraphics[width=0.15\textwidth]{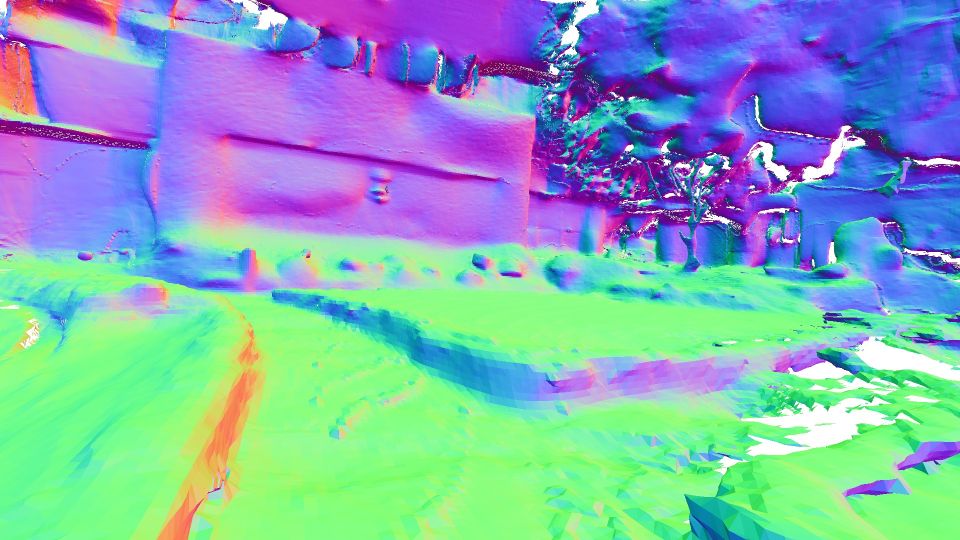}
  }

  \vspace{-6pt}
  \subfigure[PGSR]{
    \centering
    \includegraphics[width=0.15\textwidth]{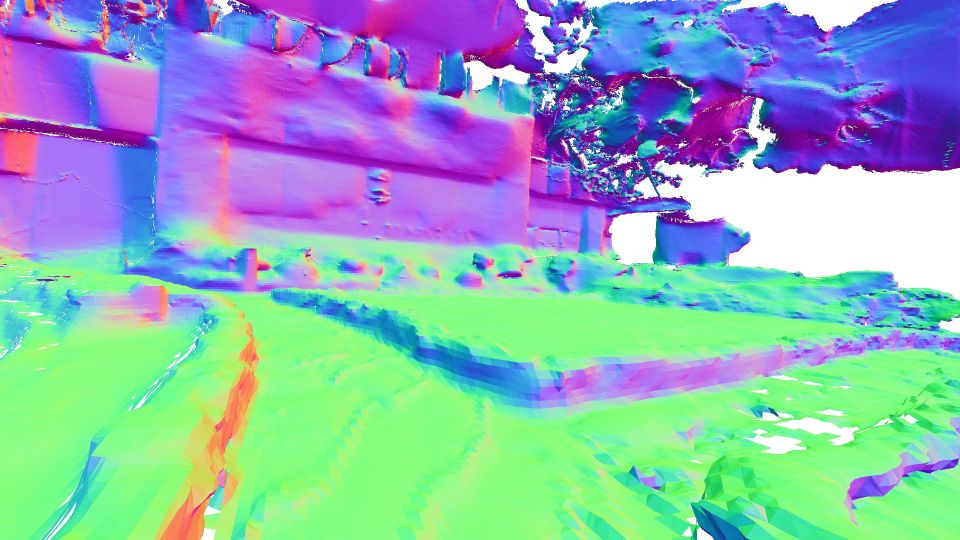}
  }
  \hspace{-12pt}
  \subfigure[M2Mapping]{
    \centering
    \includegraphics[width=0.15\textwidth]{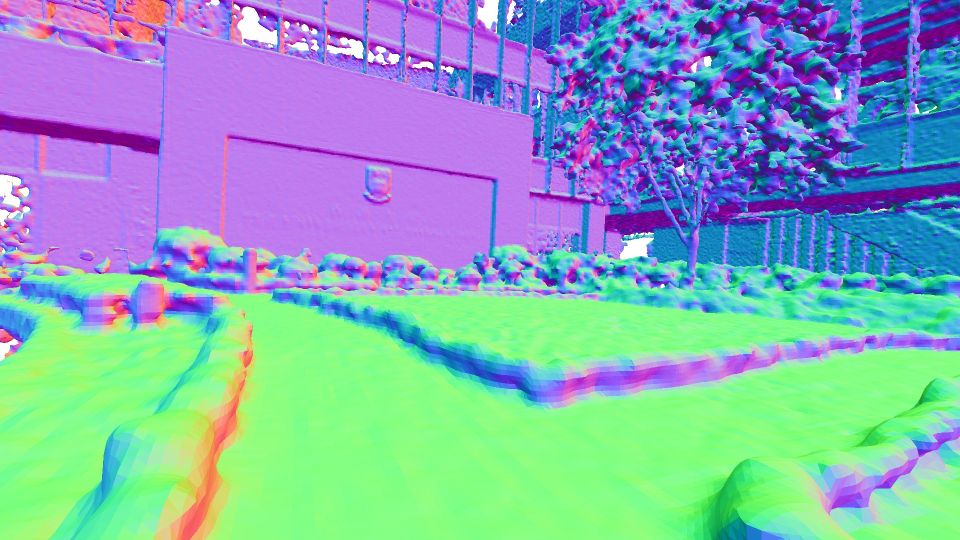}
  }
  \hspace{-12pt}
  \subfigure[Ours]{
    \centering
    \includegraphics[width=0.15\textwidth]{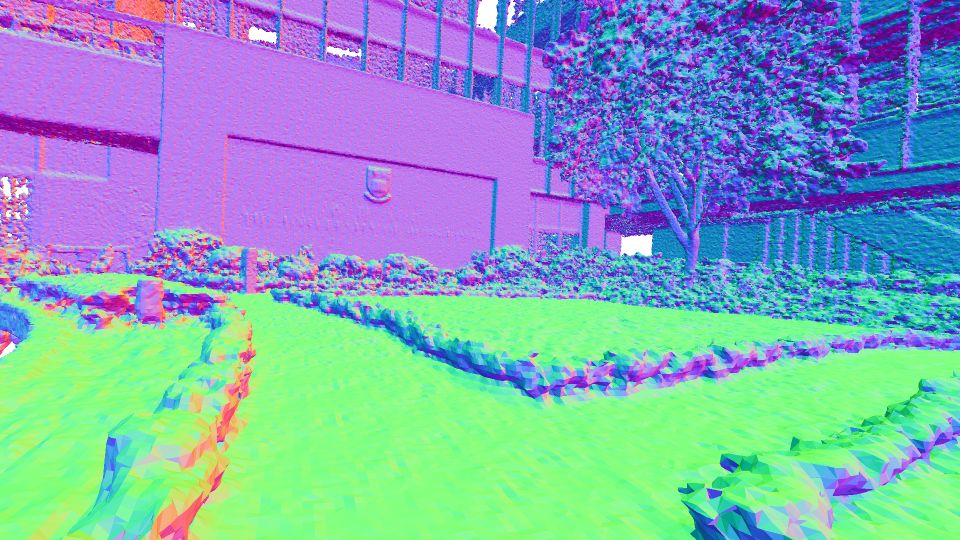}
  }

  \caption{
  The surface reconstructions in the FAST-LIVO2 Campus dataset.}
  \label{fig:fast_livo2_qual_mesh}
  \vspace{-8pt}
\end{figure}

\begin{figure}[t]
  \centering
  \setlength{\subfigcapskip}{-3pt} 
  \subfigure[InstantNGP]{
    \centering
    \includegraphics[width=0.15\textwidth]{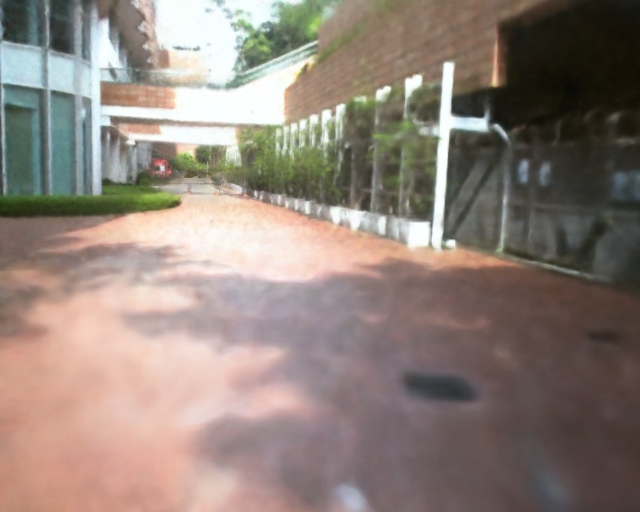}
  }
  \hspace{-12pt}
  \subfigure[3DGS]{
    \centering
    \includegraphics[width=0.15\textwidth]{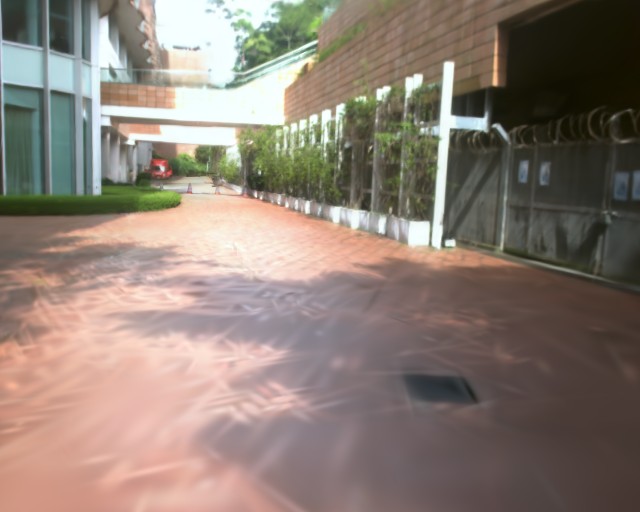}
  }
  \hspace{-12pt}
  \subfigure[2DGS]{
    \centering
    \includegraphics[width=0.15\textwidth]{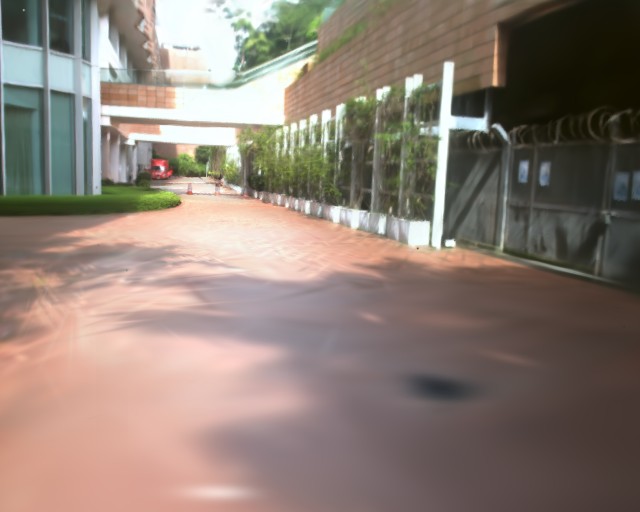}
  }

  \vspace{-6pt}

  \subfigure[PGSR]{
    \centering
    \includegraphics[width=0.15\textwidth]{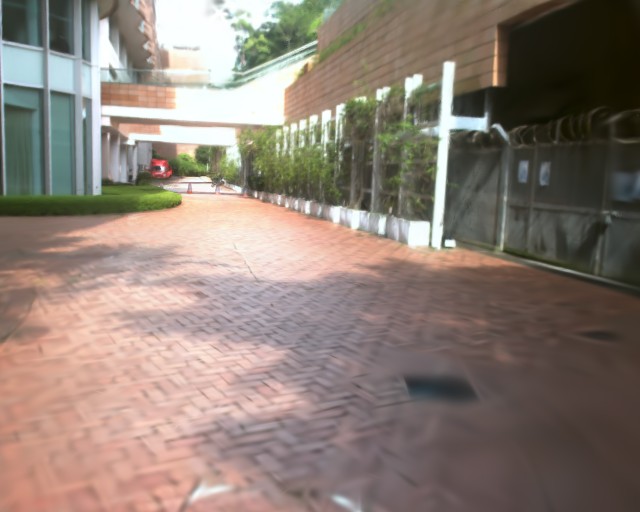}
  }
  \hspace{-12pt}
  \subfigure[M2Mapping]{
    \centering
    \includegraphics[width=0.15\textwidth]{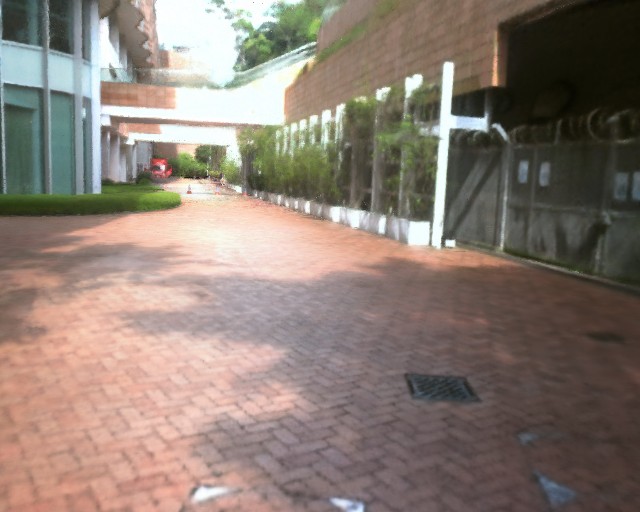}
  }
  \hspace{-12pt}
  \subfigure[Ground Truth]{
    \centering
    \includegraphics[width=0.15\textwidth]{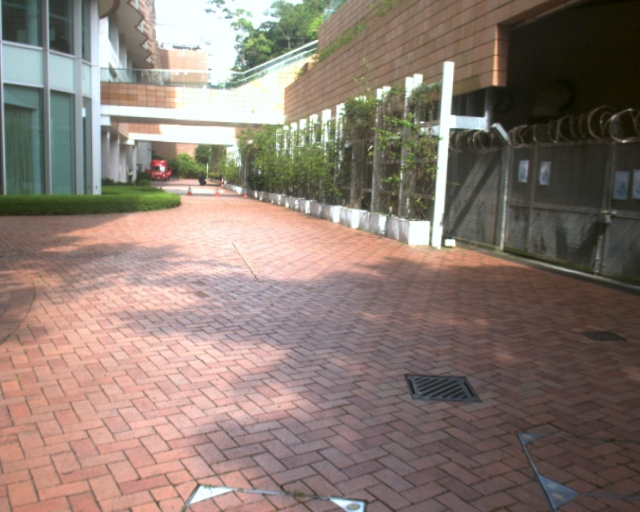}
  }

  \vspace{-6pt}

  \subfigure[RR]{
    \label{fig:fast_livo2_qual_compare:rr}
    \centering
    \includegraphics[width=0.15\textwidth]{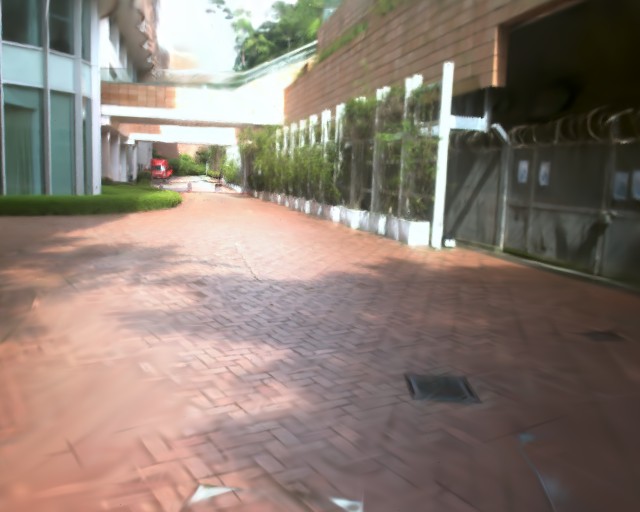}
  }
  \hspace{-12pt}
  \subfigure[RR+CR]{
    \label{fig:fast_livo2_qual_compare:cr}
    \centering
    \includegraphics[width=0.15\textwidth]{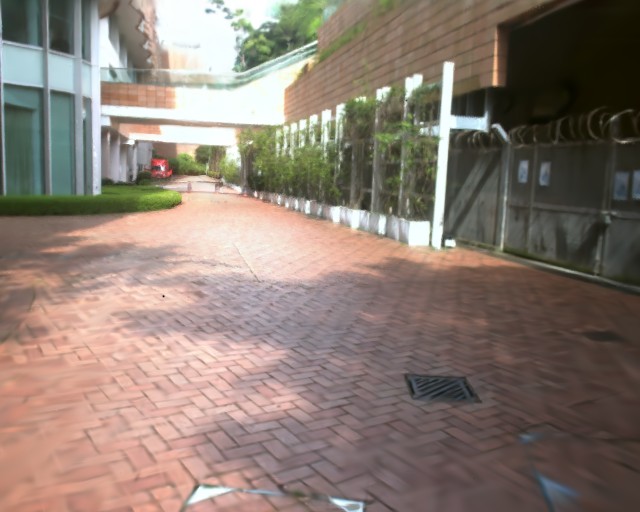}
  }
  \hspace{-12pt}
  \subfigure[Ours (RR+SR)]{
    \label{fig:fast_livo2_qual_compare:sr}
    \centering
    \includegraphics[width=0.15\textwidth]{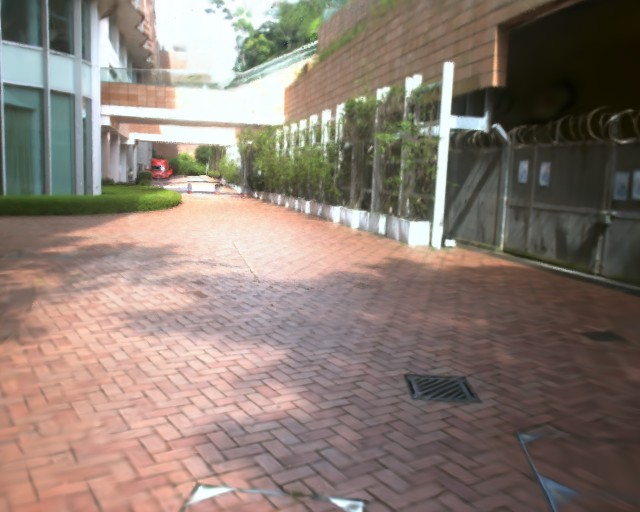}
  }


  \caption{
  The qualitative comparative results in the FAST-LIVO2 Drive dataset. We compare different geometric regularization strategies represented in Rendering Regularization (RR), Center Regularization (CR), and Shape Regularization (SR).}
  \vspace{-8pt}
  \label{fig:fast_livo2_qual_compare}
\end{figure}

We compare the qualitative rendering results in a free trajectory (Drive), as shown in Fig.~\ref{fig:fast_livo2_qual_compare}, where our method outperforms the others in retaining detailed textures of the floor with the help of geometric regularization.
It is noted that both the LiDAR-augmented NeRF-based method (M2Mapping) and 3DGS-based method (Ours) give physically grounded rendering results, while the NeRF-based method shows more complete rendering at views lacking observations, and the 3DGS-based methods show more detailed rendering overall.
We also compared the reconstruction results over the diverse types of methods, as shown in Fig.~\ref{fig:fast_livo2_qual_mesh}.
It is shown that LiDAR-based methods (VDBFusion, SHINE-Mapping, M2Mapping, Ours) offer far more detailed geometry than the vision-only methods (2DGS, PGSR).

\begin{figure}[!h]
  \centering

  \setlength{\subfigcapskip}{-3pt} 

  \subfigure[Random Initialization (0 iteration)]{
    \label{fig:ablation_init_exp:wo_init_init}
    \centering
    \begin{minipage}[b]{0.15\textwidth}
      \includegraphics[width=\linewidth]{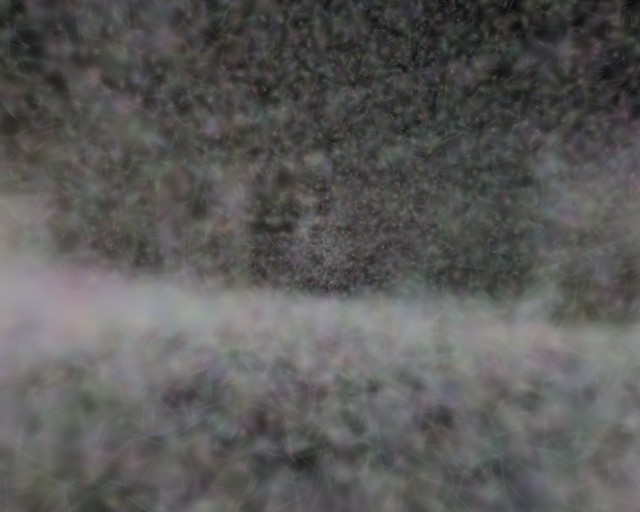}
    \end{minipage}
    \hfill
    \begin{minipage}[b]{0.15\textwidth}
      \includegraphics[width=\linewidth]{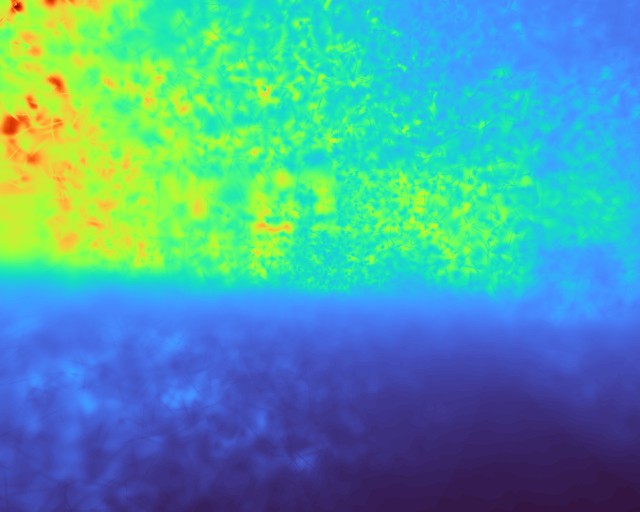}
    \end{minipage}
    \hfill
    \begin{minipage}[b]{0.15\textwidth}
      \includegraphics[width=\linewidth]{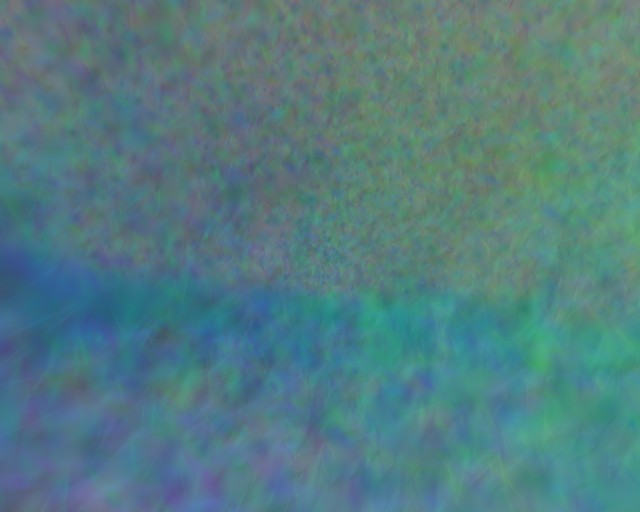}
    \end{minipage}
  }

  \vspace{-8pt}

  \subfigure[Random Initialization (30000 iteration)]{
    \label{fig:ablation_init_exp:wo_init_final}
    \centering
    \begin{minipage}[b]{0.15\textwidth}
      \includegraphics[width=\linewidth]{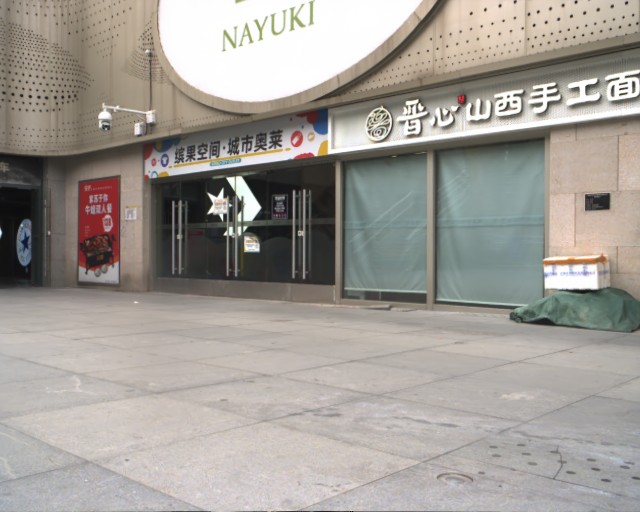}
    \end{minipage}
    \hfill
    \begin{minipage}[b]{0.15\textwidth}
      \includegraphics[width=\linewidth]{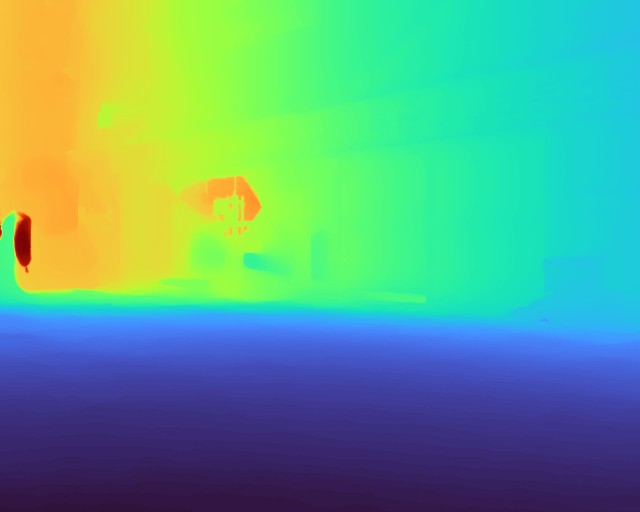}
    \end{minipage}
    \hfill
    \begin{minipage}[b]{0.15\textwidth}
      \includegraphics[width=\linewidth]{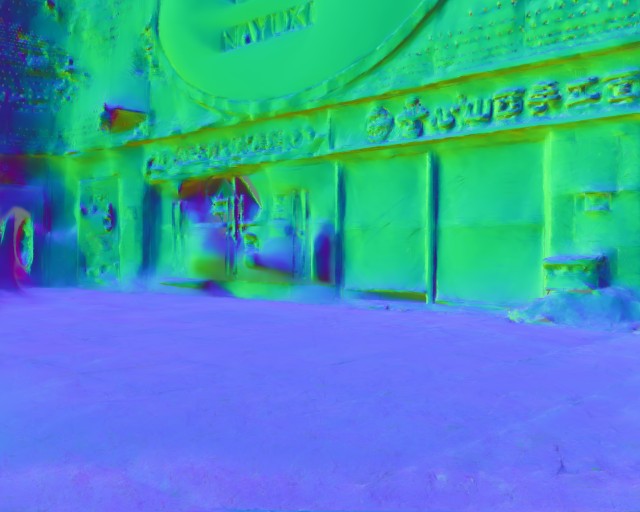}
    \end{minipage}
  }

  \vspace{-8pt}

  \subfigure[SDF-aided Initialization (0 iteration)]{
    \label{fig:ablation_init_exp:w_init_init}
    \centering
    \begin{minipage}[b]{0.15\textwidth}
      \includegraphics[width=\linewidth]{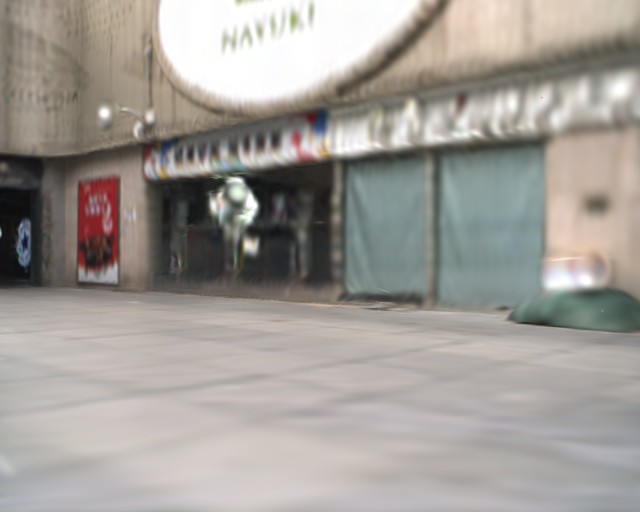}
    \end{minipage}
    \hfill
    \begin{minipage}[b]{0.15\textwidth}
      \includegraphics[width=\linewidth]{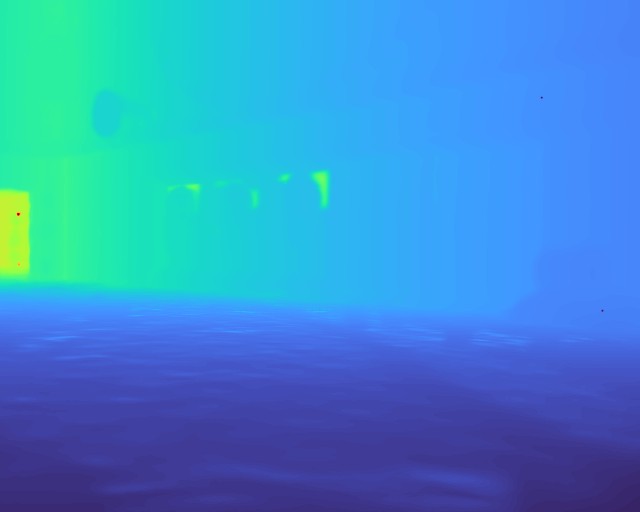}
    \end{minipage}
    \hfill
    \begin{minipage}[b]{0.15\textwidth}
      \includegraphics[width=\linewidth]{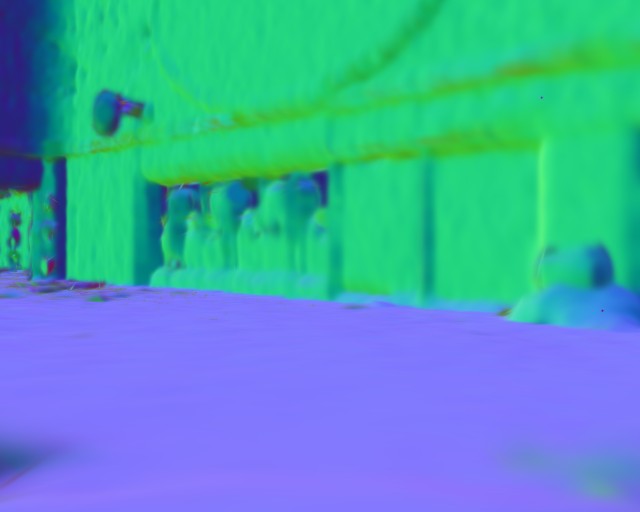}
    \end{minipage}
  }

  \vspace{-8pt}

  \subfigure[SDF-aided Initialization (30000 iteration)]{
    \label{fig:ablation_init_exp:w_init_final}
    \centering
    \begin{minipage}[b]{0.15\textwidth}
      \includegraphics[width=\linewidth]{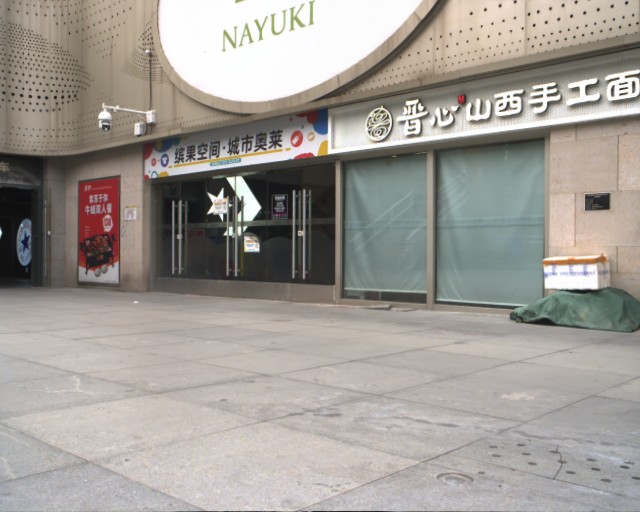}
    \end{minipage}
    \hfill
    \begin{minipage}[b]{0.15\textwidth}
      \includegraphics[width=\linewidth]{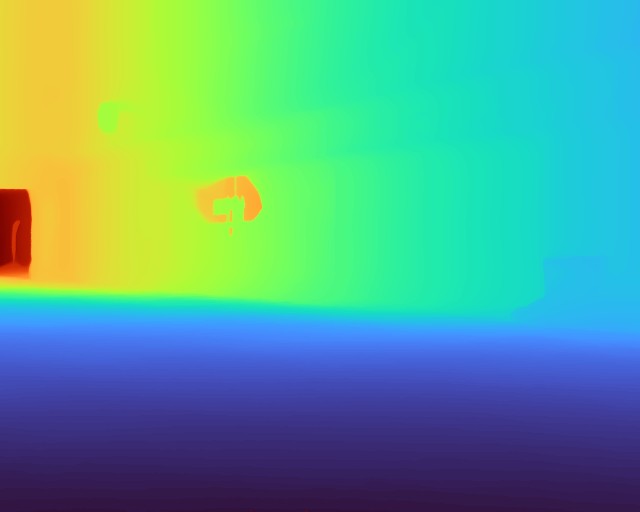}
    \end{minipage}
    \hfill
    \begin{minipage}[b]{0.15\textwidth}
      \includegraphics[width=\linewidth]{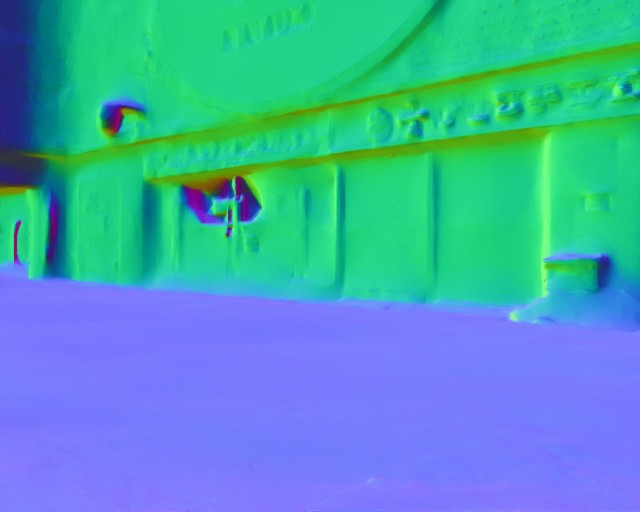}
    \end{minipage}
  }

  \vspace{-8pt}

  \subfigure[SDF-aided Initialization with Shape Reg. (30000 iteration)]{
    \label{fig:ablation_init_exp:w_init_final_sr}
    \centering
    \begin{minipage}[b]{0.15\textwidth}
      \includegraphics[width=\linewidth]{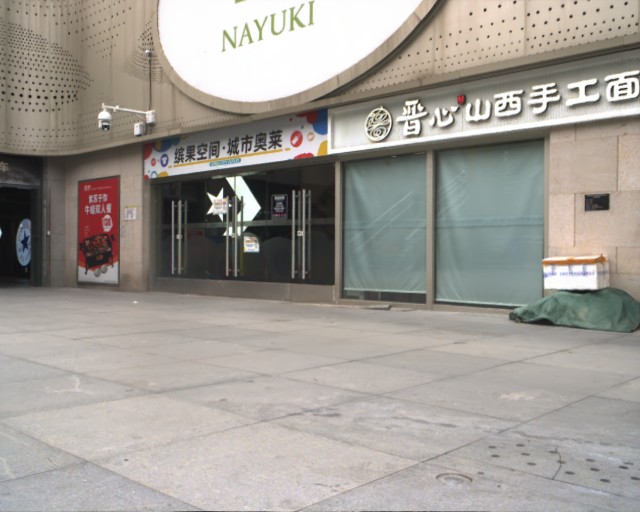}
    \end{minipage}
    \hfill
    \begin{minipage}[b]{0.15\textwidth}
      \includegraphics[width=\linewidth]{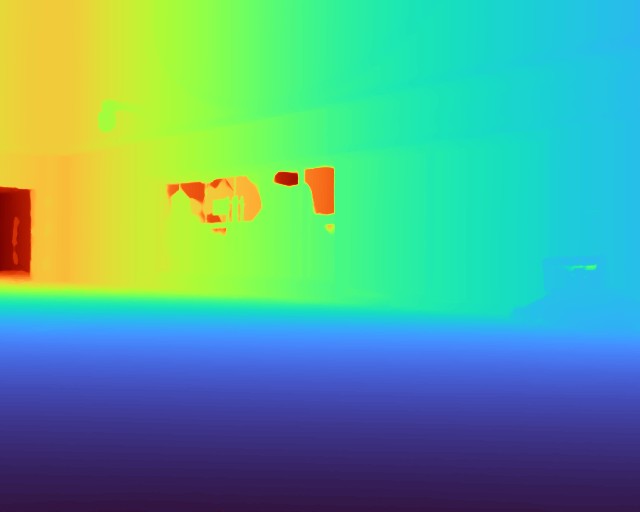}
    \end{minipage}
    \hfill
    \begin{minipage}[b]{0.15\textwidth}
      \includegraphics[width=\linewidth]{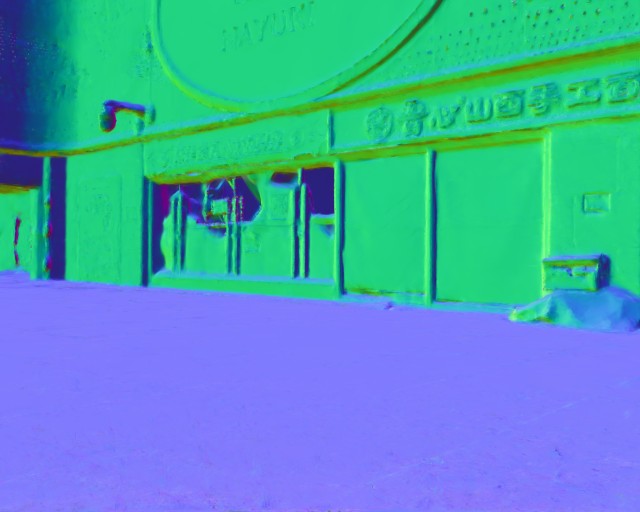}
    \end{minipage}
  }


  \caption{
    The ablation study of the SDF-aided Gaussian initialization and shape regularization in the FAST-LIVO2 Station scene.
    From left to right are the rendered color, depth, and normal images.
  }
  \vspace{-8pt}
  \label{fig:geo_init}
\end{figure}

\begin{figure}[!h]
  \centering
  \subfigure{
    \centering
    \includegraphics[width=0.15\textwidth]{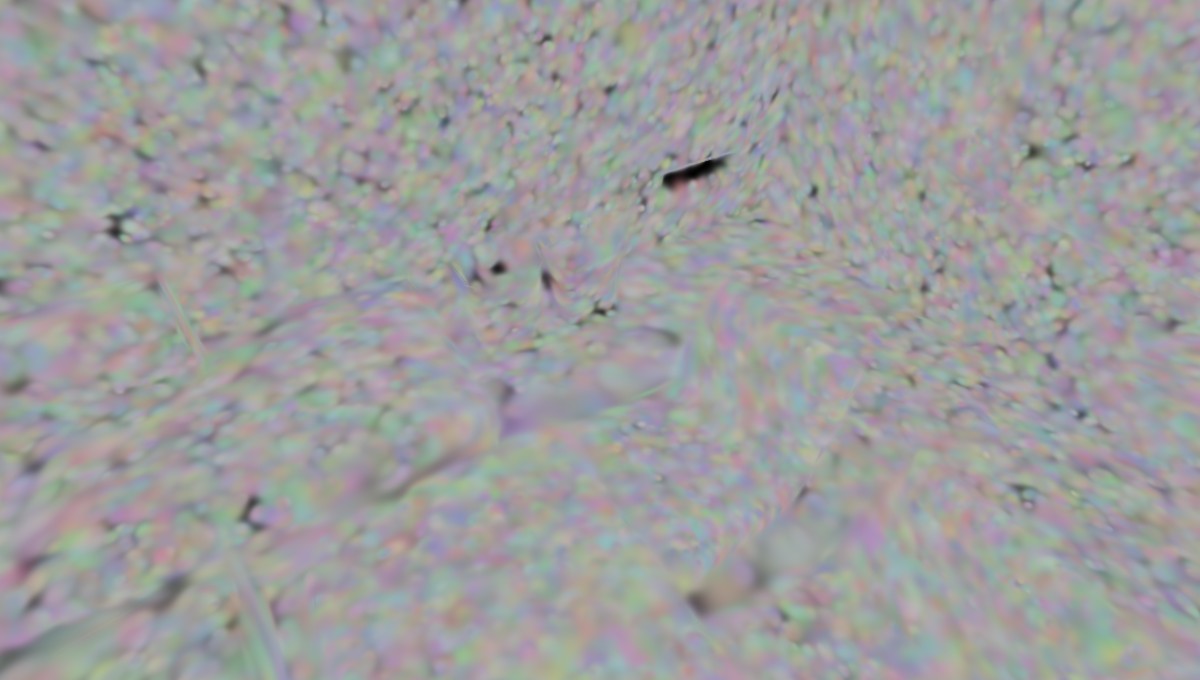}
  }
  \hspace{-11pt}
  \subfigure{
    \centering
    \includegraphics[width=0.15\textwidth]{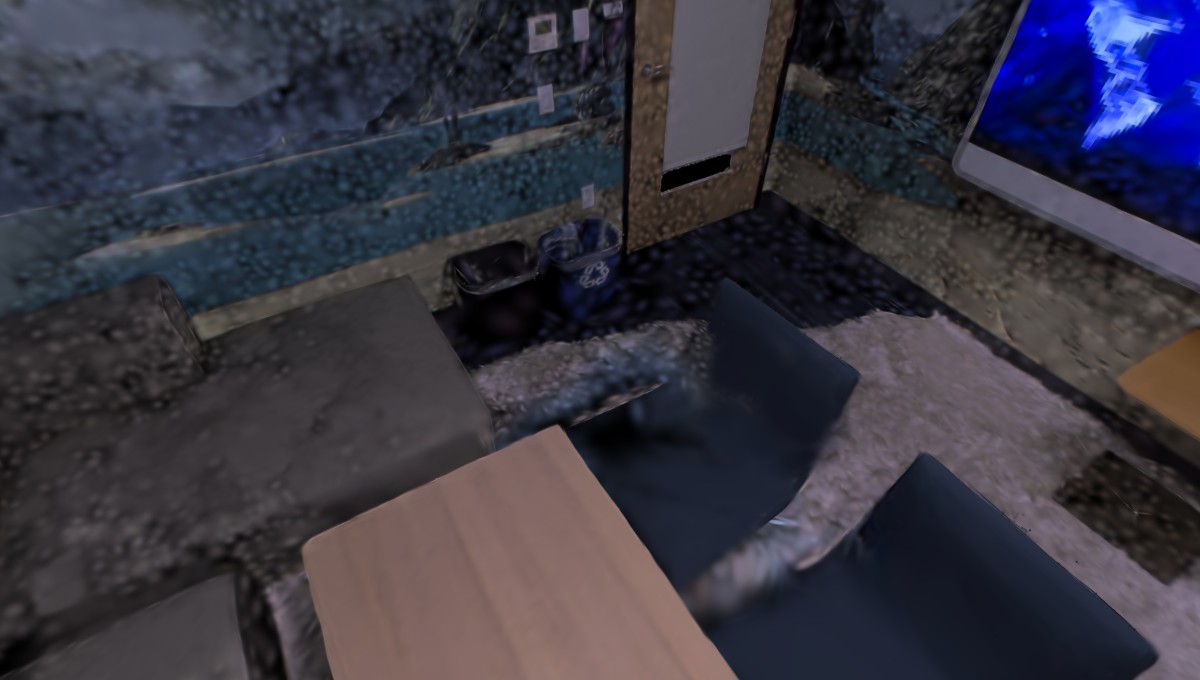}
  }
  \hspace{-11pt}
  \subfigure{
    \centering
    \includegraphics[width=0.15\textwidth]{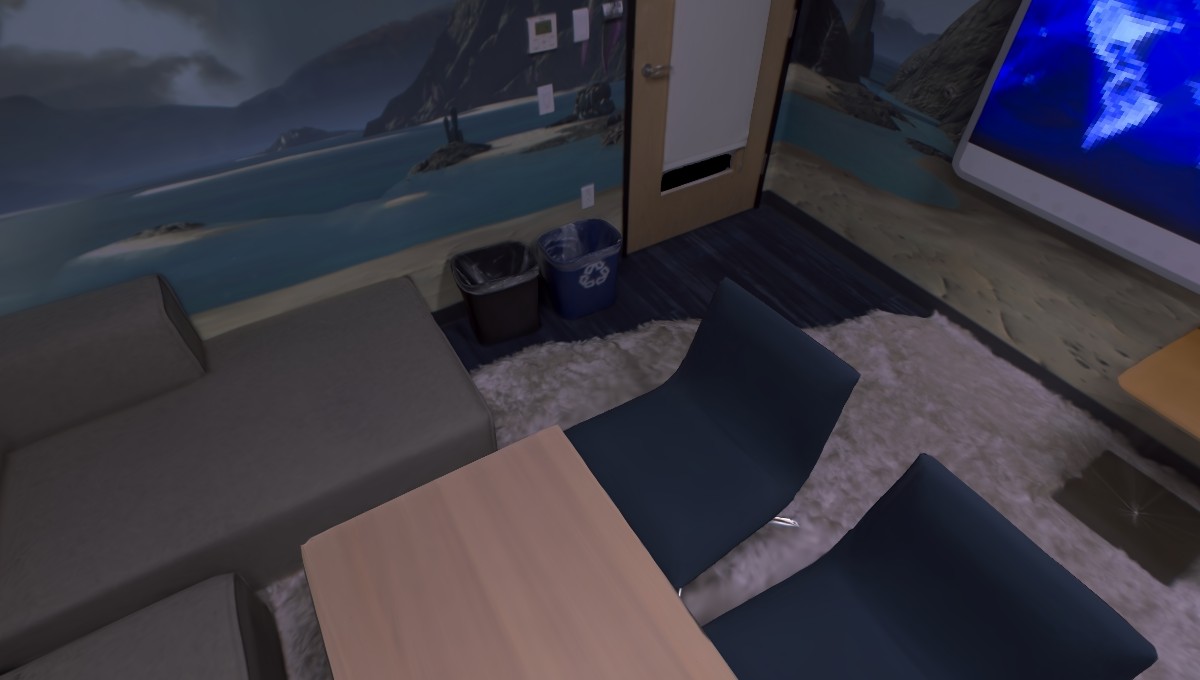}
  }

  \vspace{-8pt}

  \setcounter{subfigure}{0}
  \subfigure[0 iter.]{
    \centering
    \includegraphics[width=0.15\textwidth]{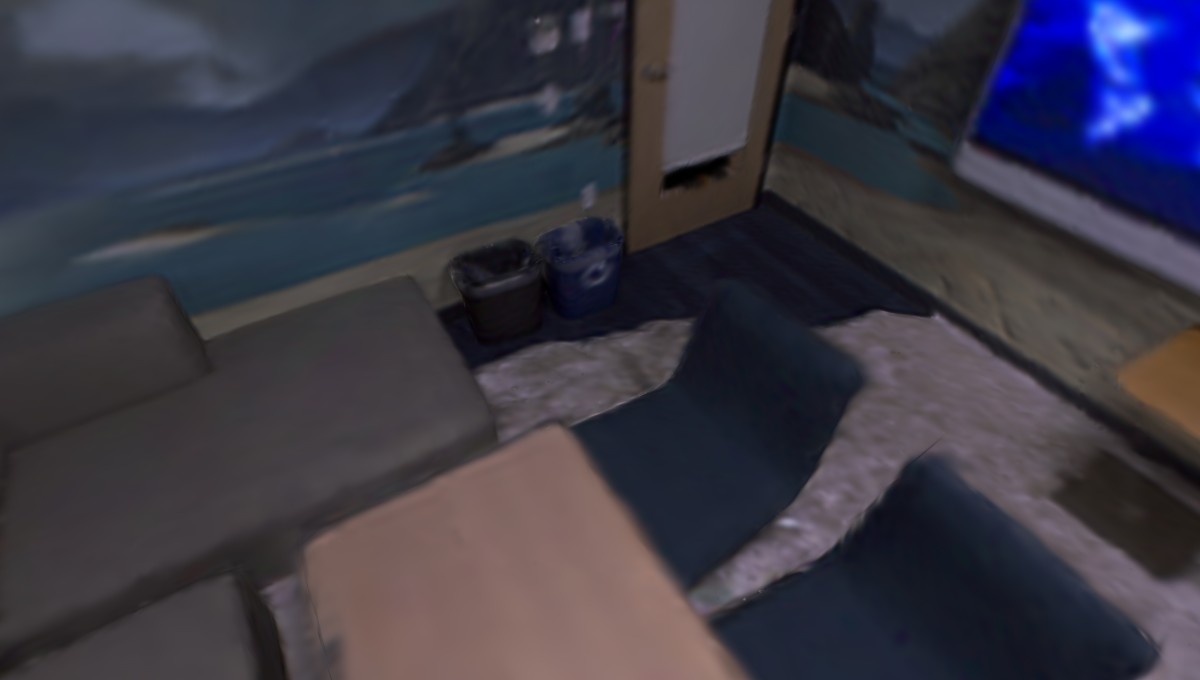}
  }
  \hspace{-11pt}
  \subfigure[2000 iter.]{
    \centering
    \includegraphics[width=0.15\textwidth]{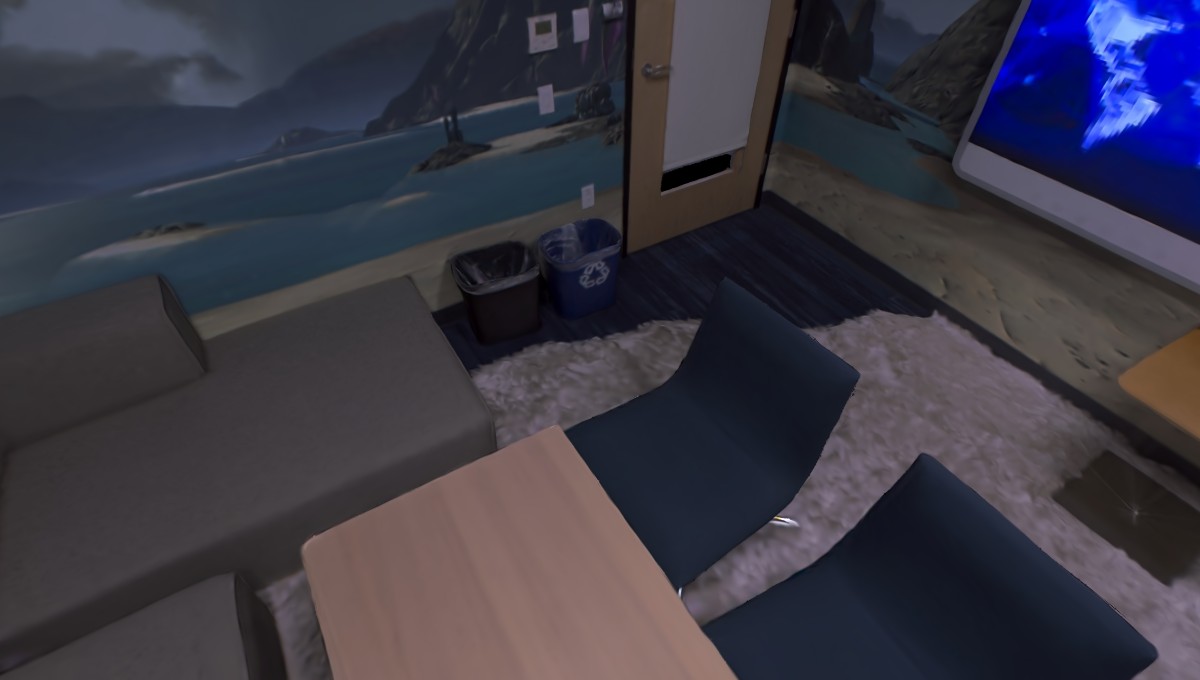}
  }
  \hspace{-11pt}
  \subfigure[30000 iter.]{
    \centering
    \includegraphics[width=0.15\textwidth]{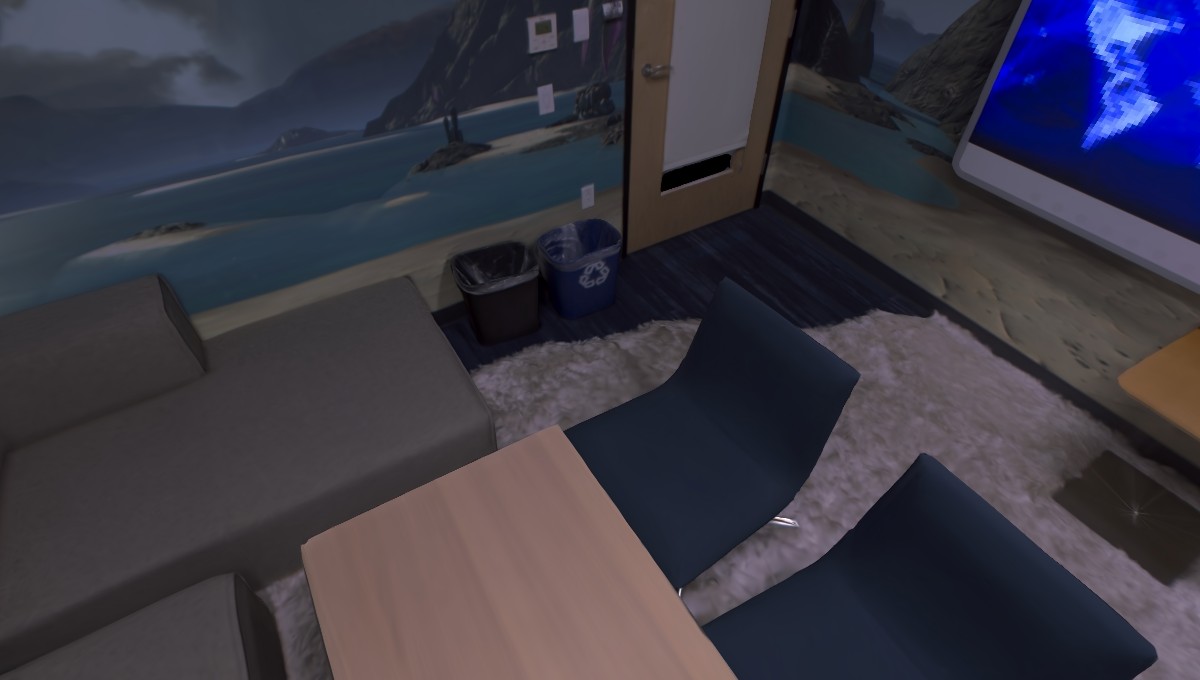}
  }

  \caption{The ablation study of the color initialization in the Replica Office0 dataset.
  The first row shows the results without color initialization, and the second row shows the results with color initialization.}
  \label{fig:ablation_geo_init_color}
  \vspace{-12pt}
\end{figure}

\subsection{Ablation Study}

\subsubsection{Initialization}


In this section, we only apply the render regularization (Eq.~\ref{eq:render_reg}) as a pure image supervision to address the importance of the proposed Gaussian initializations (Sec.~\ref{sec:gs_init}) in converging to a good structure of Gaussians.
As shown in Fig.~\ref{fig:geo_init}, the original initialization, with random sampling position from the input point clouds and random attributes initialization, falls into unreasonable structures (Fig.~\ref{fig:ablation_init_exp:wo_init_final}).
The proposed SDF-aided physically-grounded initialization gives a structure that is more in line with the real scenarios.
A color initialization assigns a proper appearance to the Gaussians, which is essential to stabilize the structure of the Gaussians.
As shown in Fig.~\ref{fig:ablation_geo_init_color}, the initialized Gaussian splats with proper structure but wrong color are deviated during early training.





\subsubsection{Geometric regularization}

We compare the geometric regularization methods between the render regularization \cite{huang20242d}, center regularization and shape regularization (Sec.~\ref{sec:gs_shape_reg}), as shown in Fig.~\ref{fig:fast_livo2_qual_compare:rr}-\ref{fig:fast_livo2_qual_compare:sr}.
The render regularization (PSNR: 27.24, SSIM: 0.805, Fig.~\ref{fig:fast_livo2_qual_compare:rr}) improves the geometric consistency to an extent, but still produces false structure on the flat surfaces under scenes with less multi-view constraints.
The center regularization (PSNR: 28.60, SSIM: 0.861, Fig.~\ref{fig:fast_livo2_qual_compare:cr}) shows its limited improvement in aligning Gaussians with the SDF field, still producing blur rendering on the ground.
And our proposed shape regularization (PSNR: 28.66, SSIM: 0.866, Fig.~\ref{fig:fast_livo2_qual_compare:sr}) shows better alignment with the surface to retain the details of the bricks.
As shown in Fig.~\ref{fig:ablation_init_exp:w_init_final_sr}'s rendered normal image, the shape regularization gives a more detailed and physically grounded structure of Gaussians for sharper rendering results, while the evaluation metrics do not show a significant improvement.

\subsection{Efficiency Analysis}

The training time for the Replica Room-2 scene takes 16.6 minutes, and the average rendering time for a 1200x680 image takes about 9.9 milliseconds and reaches 101.1 frames per second (FPS).
We evaluate the efficiency of the compared NeRF-based method, M2Mapping, which takes 53.9 milliseconds for rendering a 1200x680 image (18.4 FPS).
As our method trains both the NSDF and 3DGS, it takes more training time than the other 3DGS-based methods, like 2DGS (9.1 minutes / 103.2 FPS), but the inference time is comparable to the other methods and shows a significant improvement over the NeRF-based methods.




\section{Conclusion}

In this paper, we addressed the geometric inconsistency challenges of Gaussian splatting in robotics applications by proposing a unified LiDAR-visual system that synergizes Gaussian splatting with neural signed distance fields.
We leverage the NSDF to provide a physically grounded Gaussian initialization and effective shape regularization for geometrically consistent rendering and reconstruction.
Extensive experiments demonstrate our method's superior reconstruction accuracy and rendering quality across diverse trajectories.
However, comparative analysis reveals the limitations in extrapolative novel view synthesis capabilities contrasted with NeRF-based frameworks.
Therefore, we aim to tackle this limitation in future work by exploring advanced neural rendering techniques.



{
  \bibliographystyle{IEEEtran}
  \balance
  \bibliography{reference}
}

\end{document}